\documentclass[11pt]{article}
\usepackage{times}

\usepackage{fullpage}
\usepackage{hyperref}       
\usepackage{subcaption}
\usepackage{url}            
\usepackage{booktabs}       
\usepackage{nicefrac}       
\usepackage{microtype}      
\usepackage{amsmath,amsfonts,amssymb,amsthm}
\usepackage{graphicx}
\usepackage{ifthen}
\usepackage{color}

\def\[#1\]{\begin{align}#1\end{align}}
\def\(#1\){\begin{align*}#1\end{align*}}

\def\argmin{\operatornamewithlimits{arg\,min}}

\newcommand{\bprf}{\begin{proof}}
\newcommand{\eprf}{\end{proof}}
\newcommand{\blem}{\begin{lemma}}
\newcommand{\elem}{\end{lemma}}

\newcommand{\avg}{\mathrm{avg}}

\DeclareMathOperator{\poly}{poly}
\newcommand{\oo}{\mathcal{O}}
\newcommand{\oot}{\widetilde{\mathcal{O}}}
\newcommand{\p}[1]{\left(#1\right)}

\newcommand{\op}{\mathrm{op}}

\DeclareMathOperator{\diag}{diag}

\DeclareMathOperator{\Cov}{Cov}
\DeclareMathOperator{\sign}{sign}
\DeclareMathOperator{\tr}{tr}
\newcommand{\eqdef}{\stackrel{\mathrm{def}}{=}}

\newcommand{\bP}{\mathbb{P}}
\newcommand{\bI}{\mathbb{I}}
\newcommand{\bE}{\mathbb{E}}
\newcommand{\sH}{\mathcal{H}}
\newcommand{\sU}{\mathcal{U}}
\newcommand{\sW}{\mathcal{W}}
\newcommand{\sE}{\mathcal{E}}

\newcommand{\bR}{\mathbb{R}}
\newcommand{\sP}{\mathcal{P}}
\newcommand{\sN}{\mathcal{N}}

\newcommand{\KL}[2]{\operatorname{KL}\left(#1 \ \| \ #2\right)}

\newtheorem{theorem}{Theorem}[section]
\newtheorem{lemma}[theorem]{Lemma}
\newtheorem{proposition}[theorem]{Proposition}
\newtheorem{corollary}[theorem]{Corollary}

\newtheorem*{theorem*}{Theorem}
\newtheorem*{proposition*}{Proposition}
\theoremstyle{definition}
\newtheorem*{definition}{Definition}
\newtheorem{open-problem}[theorem]{Open Problem}

\newcommand{\func}{f}
\newcommand{\Efunc}{\bar{f}}
\newcommand{\sg}{\sigma}
\newcommand{\maxop}{S}
\newcommand{\opfrac}{\varepsilon}
\newcommand{\maxopi}{\maxop_{\opfrac}}

\newcommand{\goodfrac}{\alpha}
\newcommand{\radius}{r}
\newcommand{\param}{w}
\newcommand{\Param}{W}
\newcommand{\weight}{c}

\newcommand{\goodset}{\smash{I_{\mathrm{good}}}}
\newcommand{\goodsete}{I_{\mathrm{good}}}
\newcommand{\rf}{\radius_{\mathrm{final}}}
\newcommand{\bsserr}{\varepsilon}

\usepackage{natbib}
\definecolor{mydarkblue}{rgb}{0,0.08,0.45}                                         
\hypersetup{
  colorlinks=true,
  linkcolor=mydarkblue,
  citecolor=mydarkblue,
  filecolor=mydarkblue,
  urlcolor=mydarkblue,
}
\setcitestyle{authoryear,round,citesep={;},aysep={,},yysep={;}}
\usepackage{tikz}
\usetikzlibrary{patterns,positioning,fit,calc}
\usepackage{paralist}
\usepackage{enumitem}
\usepackage{mathtools}
\usepackage{algorithm,algorithmicx,algcompatible,algpseudocode}
\usepackage{bbm}

\newcommand{\theTitle}{Learning from Untrusted Data}
\author{
 \fontsize{11}{13}\selectfont {\bf Moses Charikar} \\
 \fontsize{11}{13}\selectfont Stanford University \\
 \fontsize{11}{13}\selectfont {\tt moses@cs.stanford.edu}
\and
 \fontsize{11}{13}\selectfont {\bf Jacob Steinhardt} \\
 \fontsize{11}{13}\selectfont Stanford University \\
 \fontsize{11}{13}\selectfont {\tt jsteinha@stanford.edu}
\and
 \fontsize{11}{13}\selectfont {\bf Gregory Valiant} \\
 \fontsize{11}{13}\selectfont Stanford University \\
 \fontsize{11}{13}\selectfont {\tt valiant@stanford.edu}
}

\title{\theTitle}
\date{}

\begin{document}
\algrenewcomment[1]{\hfill $\triangleright$ #1}

\clearpage
\maketitle
\begin{abstract}
The vast majority of theoretical results in machine learning and statistics assume that 
the available training data is a reasonably reliable reflection of the phenomena to be 
learned or estimated. Similarly, the majority of machine learning and statistical techniques 
used in practice are brittle to the presence of large amounts of biased or malicious data. 
In this work we consider two frameworks in which to study estimation, learning, and 
optimization in the presence of significant fractions of arbitrary data.

The first framework, 
\emph{list-decodable learning}, asks whether it is possible to return a list of 
answers, with the guarantee that at least one of them is accurate.  For example, given a 
dataset of $n$ points for which an unknown subset of $\alpha n$ points are drawn from a distribution 
of interest, and no assumptions are made about the remaining $(1-\alpha)n$ points, 
is it possible to 
return a list of $\poly(1/\alpha)$ answers, one of which is correct?  The second framework, which we term the 
\emph{semi-verified} learning model, considers the extent to which a small dataset of trusted 
data (drawn from the distribution in question) can be leveraged to enable the accurate 
extraction of information from a much larger but untrusted dataset (of which only an 
$\alpha$-fraction is drawn from the  distribution).

We show strong positive results in both settings, and provide an algorithm for robust learning 
in a very general stochastic optimization setting.  This general result has immediate 
implications for robust estimation in a number of settings, including for robustly estimating 
the mean of distributions with bounded second moments, robustly learning mixtures of such 
distributions, and robustly finding planted partitions in random graphs in which significant 
portions of the graph have been perturbed by an adversary.  
\end{abstract}

\thispagestyle{empty}
\newpage
\setcounter{page}{1}

\section{Introduction}
\label{sec:intro}

What can be learned from data that is only partially trusted?
In this paper, we study this question by considering the following setting: 
we observe $n$ data points, of which $\goodfrac n$ are drawn independently from 
a distribution of interest, $p^*$, and we make no assumptions about the remaining $(1-\goodfrac)n$ 
points---they could be very biased, arbitrary, or chosen by an adversary who is trying 
to obscure $p^*$. Our goal is to accurately recover a parameter of interest of $p^*$
(such as the mean), despite the presence of significant amounts of untrusted data.
Perhaps surprisingly, we will show that in high dimensions, accurate estimation and
learning is often possible, even when the fraction of real data is small (i.e., $\goodfrac \ll 1$). 
To do this, we consider two notions of successful learning--the \emph{list decodable} model 
and the \emph{semi-verified} model--and provide strong positive results for both notions. 
Our results have implications in a variety of domains, including building secure 
machine learning systems, performing robust statistics in the presence 
of outliers, and agnostically learning mixture models.

The goal of accurate robust estimation appears at first glance to be impossible 
if the fraction $\goodfrac$ of real data is less than one half.  Indeed, in the
$\goodfrac = \frac{1}{2}$ case, it is possible that the real and fake data 
are distributed identically, except that the mean of the fake data is shifted by 
some large amount; in such a case, it is clearly impossible to differentiate 
which of these two  distributions
is ``right''. Perhaps, however, such symmetries are the \emph{only} 
real problem that can occur. It might then be possible to output a short
list of possible parameter sets--if $\goodfrac = \frac{1}{2}$, perhaps
a list of two parameter sets--such that at least one is accurate. To this end, 
we consider a notion of successful 
learning called \emph{list decodable learning}, first introduced by 
\citet{balcan2008discriminative}. In analogy with list decodable coding 
theory, the goal is for the learning algorithm to output a short 
list of possible hypotheses.

\begin{definition}[List Decodable Learning]
We say that a learning, estimation, or optimization problem is $(m,\epsilon)$
\emph{list decodably solvable} if an efficient algorithm can output a
set of at most $m$ hypotheses/estimates/answers, with the guarantee that
at least one is accurate to within error $\epsilon$.
\end{definition}

\noindent A central question in this paper concerns which learning problems 
can be robustly solved in the above sense:
\begin{quotation}
\noindent \emph{To what extent are learning problems
robustly solvable in the {list decodable} sense?  If the dataset
consists of only an $\goodfrac$-fraction of real data, in what settings 
is it possible to efficiently output a list of at most $\frac{1}{\goodfrac}$ or 
$\poly(\frac{1}{\goodfrac})$ parameter sets or
estimates with the guarantee that at least one closely approximates
the solution that could be obtained if one were given only honest data?}
\end{quotation}
The intuition for why strong positive results are obtainable in
the list decodable setting is the following.  Given a dataset with an
$\goodfrac$ fraction of trusted data, the remaining data might do one of
two things: either it can be fairly similar to the good data, in which
case it can bias the overall answers by only a small amount, or the
adversarial data may be very different from the trusted data.
The key
is that if a portion of the untrusted data tries too hard to bias the final 
result, then it will end up looking quite different, and can be clustered out.
Given this viewpoint, the question of robust 
learning in the list decodable model becomes a question of the 
extent to which an adversary can \emph{completely} 
obscure the inherent structure possessed by a 
dataset drawn from a well-behaved distribution.

Our investigation of robust learning has three motivations.  First,
from a theoretical perspective, it is natural to ask 
what guarantees are possible in the setting in which a majority of
data is untrusted ($\goodfrac < \frac{1}{2}$).  Is it the case that learning
really becomes impossible (as is often stated), or can one at least
narrow down the possible answers to a small set?  Second, in
many practical settings, there is a trade-off between the amount of
data one can collect, and the quality of the data.  For a fixed price,
one might be able to collect either a small and accurate/trusted
dataset, or a large but less trusted dataset.  It is worth
understanding how the quality of models derived from such datasets
varies, across this entire range of dataset quality/quantity. 
Finally, robust learning with $\alpha \ll 1$ provides a new 
perspective on learning mixtures of distributions---by treating 
a single mixture component as the real data, and the remaining 
components as fake data, we can ask to what extent a mixture component 
can be learned, independently of the structure of the other 
components.
While this perspective may seem to give up too much, we will show, 
somewhat surprisingly, that it is possible to learn mixtures almost 
as well under these adversarial assumptions as under stochastic assumptions.

\paragraph{Semi-Verified Learning.}
When $\goodfrac \leq \frac{1}{2}$, the list decodable model 
handles symmetries by allowing the learner 
to output multiple possible answers; an alternative is to break 
these symmetries with a small amount of side information. 
In particular, in many practical settings it is possible to 
obtain a (sometimes extremely small) verified set of data that 
has been carefully checked, which could be used to determine 
which of multiple alternative answers is correct. This motivates 
us to introduce the following new notion of learnability:
%
%
\begin{definition}[The Semi-Verified Model]
In the \emph{semi-verified model}, we observe $n$ data points, 
of which an unknown $\goodfrac n$ are ``real'' data reflecting an 
underlying distribution $p^*$, and the remaining $(1-\goodfrac)n$ points are arbitrary.
Furthermore, we observe $k$ ``verified'' data points that are guaranteed to be drawn
from $p^*$.
\end{definition}

The definition of the semi-verified model is inspired
by the \emph{semi-supervised} model of learning (see
e.g. \cite{chapelle2006semisupervised}). In semi-supervised learning, one is
concerned with a prediction/labeling task, and has access to a large
amount of unlabeled data together with a small amount of labeled data; 
the central question is whether the presence of the unlabeled data can
reduce the amount of labeled data required to learn.
Analogously, in our robust learning setting, we are asking whether
the presence of a large amount of untrusted data can reduce the amount
of trusted 
data required for 
learning.  Clearly the answer is ``no'' if we make no
assumptions on the untrusted data.  Nevertheless,
the assumption that a significant fraction of that
data is drawn from $p^*$ seems plausible,  
and may be sufficient to achieve strong positive results. 
We therefore ask:
\begin{quotation}
\noindent \emph{To what extent can the availability of a modest amount of ``verified''
data facilitate (either computationally or information theoretically)
the extraction of the information contained in the larger but
untrusted dataset? What learning tasks can be performed in the above
semi-verified setting given $k \ll n$ verified data points?
How does the amount $k$ of verified data that is needed vary with the setting, the
fraction $\goodfrac$ of honest data, etc.?}
\end{quotation}

\noindent The above definition and associated questions reflect 
challenges faced in a number of practical settings, particularly
those involving large crowdsourced datasets, or datasets
obtained from unreliable sensors or devices.  In such settings, 
despite the unreliability of the data, it is often possible to obtain 
a small verified dataset that has been carefully checked. Given 
its pervasiveness, it is somewhat surprising that neither the
theory nor the machine learning communities have formalized this
model; we think it is important to develop an understanding 
of the algorithmic possibilities in this domain.
Obtaining theoretical guarantees seems especially important for 
designing \emph{provably secure} learning systems 
that are guaranteed to 
perform well even if an adversary obtains control 
over some of the training data used by the algorithm.


\paragraph{Relationships between the models.}
The semi-verified and list decodable models can be reduced to each other. 
Informally, given $m$ candidate outputs from a list decodable algorithm, we expect 
to be able to distinguish between them with $\oo\p{\log(m)}$ verified data points. 
Conversely, if a model is learnable with $k$ verified points then we can output 
$\oo\p{({\frac{1}{\goodfrac}})^k}$ 
candidate parameters in the list decodable setting (since if we sample that many $k$-tuples from 
the untrusted data, at least one is likely to contain only honest data). 
For simplicity we state most results in the list decodable model.

\paragraph{Our contributions.} 
We provide results on robust learnability in a general stochastic optimization 
setting, where we observe convex functions $\func_1, \ldots, \func_n$ of which $\goodfrac n$ 
are sampled from $p^*$, and want 
to minimize the population mean $\Efunc = \bE_{p^*}[\func]$.\footnote{
Typically, we actually observe data points $x_i$, and $\func_i$ is the loss function corresponding to $x_i$.}
Our results are given in terms of a spectral norm bound on the gradients $\nabla \func_i$.
Therefore, we obtain robustness in any setting where we can establish a matrix concentration 
inequality on the good data --- for instance, if the $\nabla \func_i$ are sub-Gaussian and Lipschitz, 
or sometimes even with only bounded second moments.

From our general results (discussed in detail in the next section), we immediately obtain corollaries in specific settings, 
starting with mean estimation:
\begin{itemize}
\item {\bf Robust mean estimation:} When $\goodfrac > \frac{1}{2}$ we can robustly 
      estimate the mean of a distribution $p^*$ to $\ell_2$ error $\oo\p{\sigma}$, where $\sigma^2$ 
      is a bound on the second moments of $p^*$. For $\goodfrac$ 
      bounded away from $1$, this improves upon existing work, which achieves 
      error either $\oo{(\sigma\sqrt{\log(d)})}$ under a $4$th moment bound on $p^*$ 
      \citep{lai2016agnostic} or matches our rate of $\oo\p{\sigma}$ but assumes 
      $p^*$ is sub-Gaussian \citep{diakonikolas2016robust}. For $\goodfrac \leq \frac{1}{2}$, 
      which was previously unexplored, we can estimate the mean to error 
      $\tilde{\oo}\p{\sigma/\sqrt{\goodfrac}}$.
\end{itemize}
Since our results hold for any stochastic optimization problem, we can also study 
density estimation, by taking $\func_i$ to be the negative log-likelihood:
\begin{itemize}
\item {\bf Robust density estimation:} Given an exponential family model 
      $p_{\theta}(x) \propto \exp\p{\theta^{\top}\phi(x)}$, we can find a 
      distribution $p_{{\theta}}$ with $\KL{p_{\theta^*}}{p_{{\theta}}} \leq \oo({\frac{\sigma\radius}{\sqrt{\goodfrac}}})$, 
      where $\Cov_{p_{\theta^*}}[\phi(x)] \preceq \sigma^2 I$ and $\radius = \|\theta^*\|_2$.
\end{itemize}
While density estimation could be reduced to mean estimation (via estimating the 
sufficient statistics), our analysis applies directly, to an algorithm that can be 
interpreted as approximately maximizing the log likelihood while removing outliers.

In the list decodable setting, our results also yield bounds for learning mixtures:
\begin{itemize}
\item {\bf Learning mixture models:} Given a mixture of $k$ distributions each with covariance 
      bounded by $\sigma^2$, and with minimum mixture weight $\goodfrac$, we can accurately 
      cluster the points if the means are separated by a distance 
      $\tilde{\Omega}\p{\sigma/\sqrt{\goodfrac}}$, even in the presence of 
      additional adversarial data. For comparison, even with few/no bad data points, 
      the best efficient clustering algorithms 
      require mean separation $\tilde{\Omega}{(\sigma\sqrt{k})}$ \citep{awasthi2012improved,achlioptas2005spectral},
      which our rate matches if $\goodfrac = \Omega(\tfrac{1}{k})$. 
\item {\bf Planted partition models:} In the planted partition model, we can 
      approximately recover the planted partition if the average degree is 
      $\tilde{\Omega}(1/\goodfrac^3)$, where $\goodfrac n$ is the size of the 
      smallest piece of the partition. The best computationally efficient 
      result (which assumes all the data is real) requires the degree to be 
      $\Omega(1/\goodfrac^2)$ \citep{abbe2015community,abbe2015detection}. 
\end{itemize}
It is fairly surprising that, despite making no assumptions on the 
structure of the data outside of 
a mixture component, we nearly match the best computationally efficient 
results that fully leverage this structure.
This suggests that there may be a connection between robustness and computation: 
perhaps the \emph{computational threshold} for recovering a planted structure in random data 
(such as a geometric cluster or a high-density subgraph) matches the 
\emph{robustness threshold} for recovering that structure in the presence of 
an adversary.

\paragraph{Technical highlights.}
Beyond our main results, we develop certain technical machinery 
that may be of broader interest. Perhaps the most relevant is a 
novel matrix concentration inequality, based on ideas from 
spectral graph sparsification \citep{batson2012twice}, 
which holds assuming only bounded second moments:\footnote{Since 
the writing of this paper, we were able to prove a stronger 
version of Proposition~\ref{prop:bss-intro} that requires 
only bounded first moments.}
\begin{proposition}
\label{prop:bss-intro}
Suppose that $p$ is a distribution on $\bR^d$ with $\bE_p[X] = \mu$ and 
$\Cov_p[X] \preceq \sigma^2 I$ for some $\sigma$.
Then, given $n \geq d$ samples from $p$, with probability $1 - \exp\p{-\frac{n}{64}}$ there is a subset 
$I \subseteq [n]$ of size at least $\frac{n}{2}$ such that 
$\lambda_{\max}\p{\frac{1}{|I|} \sum_{i \in I} (x_i-\mu)(x_i-\mu)^{\top}} \leq 24\sigma^2$, 
where $\lambda_{\max}$ denotes the maximum eigenvalue.
\end{proposition}
This result is strong in the following sense: if one instead uses all $n$ samples 
$x_i$, the classical result of \citet{rudelson1999random} only bounds the 
maximum eigenvalue by roughly ${\sigma^2\log(n)}$, and even then only in expectation. 
Even under stronger assumptions, one often needs either at least $d\log(d)$ samples, 
or incurs a $\log(d)$ factor in the bound on $\lambda_{\max}$. 
In the planted partition model, this log factor causes natural 
spectral approaches to fail on sparse graphs, 
and avoiding the log factor has been a topic of recent interest 
\citep{guedon2014community,le2015concentration,rebrova2015coverings,rebrova2016norms}.

Proposition~\ref{prop:bss-intro} says that the undesirable 
log factor only arises due to a manageable fraction of bad samples, 
which when removed give us sharper concentration. Our framework allows us to exploit 
this by defining the good data to be the (unknown) set $I$ 
for which the conclusion of Proposition~\ref{prop:bss-intro} holds.
One consequence is that we are able to recover planted partitions in 
sparse graphs essentially ``for free''.


Separately, we introduce a novel regularizer based on minimum trace 
ellipsoids. This regularizer allows us to control 
the spectral properties of the model parameters at multiple scales simultaneously, and yields 
tighter bounds than standard trace norm regularization. We define the regularizer in 
Section~\ref{sec:algorithm}, and prove a 
\emph{local H\"{o}lder's inequality} (Lemma~\ref{lem:local-holder}), 
which exploits this regularization to achieve 
concentration bounds solely from deterministic spectral information.

We also employ \emph{padded decompositions}, 
a space partitioning technique 
from the metric embedding literature \citep{fakcharoenphol2003tight}. 
Their use is the following: when the loss functions are strongly convex, 
we can improve our bounds by identifying clusters in the data, and 
re-running our main algorithm on each cluster. 
Padded decompositions help us because they can identify clusters 
even if the remaining data has arbitrary structure.
Our clustering scheme is described in Section~\ref{sec:strongly-convex}.

\paragraph{Related work.}
The work closest to ours is
\citet{lai2016agnostic} and \citet{diakonikolas2016robust}, who study high-dimensional 
estimation in the presence of adversarial corruptions. 
They focus on the regime $\goodfrac \approx 1$, while our work focuses on 
$\goodfrac \ll 1$. In the overlap of these regimes (e.g. for $\goodfrac = \frac{3}{4}$) our results 
improve upon these existing results. (The existing bounds are better as $\goodfrac \to 1$, but do 
not hold at all if $\goodfrac \leq \frac{1}{2}$.) 
The popular robust PCA algorithm
\citep{candes2011robust,chandrasekaran2011rank} allows for a constant fraction of the \emph{entries} 
to be arbitrarily corrupted, but assumes the locations of these entries are sufficiently 
evenly distributed.
However, \citet{xu2010principal} give a version of PCA that is robust to 
arbitrary adversaries if $\goodfrac > \frac{1}{2}$.
\citet{bhatia2015robust} study linear regression in the presence of adversaries, 
and obtain bounds for sufficiently large $\goodfrac$ 
(say $\goodfrac \geq \frac{64}{65}$) when the design matrix is 
\emph{subset strong convex}.
\citet{klivans2009learning} and \citet{awasthi2014power} provide strong bounds for robust classification 
in high dimensions for isotropic log-concave distributions.

The only works we are aware of that achieve general adversarial 
guarantees when $\goodfrac \leq \frac{1}{2}$ 
are \citet{hardt2013algorithms}, who study robust subspace recovery in the 
presence of a large fraction of outliers,
and \citet{steinhardt2016avoiding}, which is an early version of this work that focuses 
on community detection.


\citet{balcan2008discriminative} introduce the list-decodable learning model, 
which was later studied by others, e.g. \citet{balcan2009agnostic}
and \citet{kushagra2016finding}. That work provides bounds for clustering in 
the presence of some adversarial data, but has two limitations relative to 
our results (apart from being in a somewhat different setting): the fraction 
of adversaries tolerated is small ($\oo(\frac{1}{k})$), and the bounds are not 
meaningful in high dimensions (e.g. \citet{balcan2008discriminative} output 
a list of $k^{\oo(k/\gamma^2)}$ hypotheses, where $\gamma$ typically scales as 
$1/\sqrt{d}$).

\citet{kumar2010clustering} and the follow-up work of 
\citet{awasthi2012improved} find deterministic conditions 
under which efficient $k$-means clustering is possible, even in high dimensions. While the goal is different from 
ours, their condition is similar to the quantities appearing in our error bounds, and there is 
some overlap in techniques. They also obtain bounds in the presence of adversaries, but only 
if the fraction of adversaries is smaller than $\frac{1}{k}$ as in \citet{balcan2008discriminative}. 
Our corollaries for learning mixtures can be thought of as 
extending this line of work, by providing deterministic conditions under which 
clustering is possible even in the presence of a large fraction of adversarial data. 

Separately, there has been 
considerable interest in
\emph{semi-random graph models}  \citep{blum1995coloring,feige2000finding,
feige2001heuristics,coja2004coloring,krivelevich2006semirandom,
coja2007solving,makarychev2012approximation,chen2014improved,guedon2014community,
moitra2015robust,agarwal2015multisection}
and \emph{robust community detection} \citep{kumar2010clustering,moitra2015robust,makarychev2015learning,cai2015robust}.
In these models, a random graph is generated with 
a planted structure (such as a planted clique or partition) 
and adversaries are then allowed to modify some parts of this 
structure. 
Typically, 
the adversary is constrained to only modify $o(n)$ nodes or to 
only modify the graph in restricted ways, though some of the above 
work considers substantially stronger adversaries as well.

Robust learning is interesting from not just an information-theoretic 
but also a computational perspective. 
\citet{guruswami2009hardness} and \citet{feldman2009agnostic} show that 
learning half-spaces is NP-hard for any $\goodfrac < 1$,
while \citet{hardt2013algorithms} show that learning 
$k$-dimensional subspaces in $\bR^d$ is hard if $\goodfrac < \frac{k}{d}$.
More generally, algorithms for list decodable learning imply algorithms 
for learning mixture models, e.g. planted partitions or mixtures 
of sub-Gaussian distributions, which is thought to be 
computationally hard in at least some regimes.

Finally, there is a large literature on learning in the 
presence of errors, spanning multiple communities including learning theory 
\citep{kearns1993learning} and statistics \citep{tukey1960survey}.
We refer the reader to 
\citet{huber2009robust} and \citet{hampel2011robust} for some recent surveys.

\paragraph{Comparison of techniques.} We pause to explain how our techniques 
relate to those of other recent robust learning papers by \citet{diakonikolas2016robust} and 
\citet{lai2016agnostic}. At a high level, our algorithm works by solving a convex 
optimization problem whose objective value will be low if all of the data come 
from $p^*$; then, if the objective is high, by looking at the dual we can identify 
which points are responsible for the high objective value and remove them as outliers.

In contrast, \citet{diakonikolas2016robust} solve a convex \emph{feasibility} problem, 
where the feasible set depends on the true distribution $p^*$ and hence is not observable. 
Nevertheless, they show that given a point that is far from feasible, it is possible 
to provide a separating hyperplane demonstrating infeasibility. Roughly speaking, then, 
we solve a ``tainted'' optimization problem and clean up errors after the fact, while 
they solve a ``clean'' (but unobserved) optimization problem and show that it is 
possible to make progress if one is far from the optimum. The construction of the 
separation oracle in \citet{diakonikolas2016robust}
is similar to the outlier removal step we present here, and it would be interesting to 
further understand the relationship between these approaches.

\citet{diakonikolas2016robust} also propose another algorithm based on \emph{filtering}. 
In the case of mean estimation, the basic idea is to compute the maximum eigenvector
of the empirical covariance of the data --- if this eigenvector is too large, 
then we can find a collection of points that are responsible for it being large, and remove 
them as outliers. Though it is not phrased this way, it can be thought 
of--similarly to our approach--as solving a tainted optimization problem 
(top eigenvalue on the noisy data) and then cleaning up outliers afterwards. 
Their outlier removal step seems tighter than ours, and it would be interesting 
to find an approach that obtains such tight bounds for a general class of optimization 
problems.

Finally, \citet{lai2016agnostic} pursue an approach based on iteratively 
finding the top $n/2$ eigenvectors (rather than just the top) and projecting 
out the remaining directions of variation, as well as removing outliers if the 
eigenvalues are too large. This seems similar in spirit to the filtering approach 
described above.

\paragraph{Outline.}
Our paper is organized as follows. In Section~\ref{sec:results} we present our main 
results and some of their implications in specific settings. In Section~\ref{sec:algorithm} 
we explain our algorithm and provide some intuition for why it should work. 
In Section~\ref{sec:proof-outline} we provide a proof outline for our main results. 
In Sections~\ref{sec:local-holder} and \ref{sec:strongly-convex}, we sharpen our results, 
first showing how to obtain concentration inequalities on the errors, and then showing 
how to obtain tighter bounds and stronger guarantees for strongly convex losses. 
In Section~\ref{sec:lower-bounds} we present lower bounds showing that our results are optimal 
in some settings. In Section~\ref{sec:intuition} we present some intuition for our bounds, and 
in Section~\ref{sec:applications} we prove our main corollaries. 
The remaining sections are dedicated to deferred proofs.

\paragraph{Acknowledgments.}
We thank the anonymous reviewers who made many helpful comments that improved 
this paper. 
MC was supported by NSF grants CCF-1565581, CCF-1617577, CCF-1302518 
and a Simons Investigator Award.
JS was supported by a Fannie \& John Hertz Foundation Fellowship, a 
NSF Graduate Research Fellowship, and a Future of Life Institute grant.
GV was supported by NSF CAREER award CCF-1351108, a Sloan Foundation
Research Fellowship, and a research grant from the Okawa Foundation.

\section{Main Results and Implications}
\label{sec:results}

We consider a general setting of stochastic optimization with adversaries. 
We observe convex functions $\func_1, \ldots, \func_n : \sH \to \bR$, where 
$\sH \subseteq \bR^d$ is a convex parameter space. For a subset $\goodset \subseteq [n]$ 
of size $\goodfrac n$, $\func_i \stackrel{\scriptstyle i.i.d.}{\sim} p^*$ for $i \in \goodset$, and the remaining $\func_i$ are chosen by 
an adversary whose strategy can depend on the $f_i$ for $i \in \goodset$. 

Let $\Efunc$ denote the mean of $\func$ under $p^*$, i.e. $\Efunc(\param) \eqdef \bE_{f \sim p^*}[f(\param)]$ for 
$\param \in \sH$; our goal is to find a parameter $\hat{\param}$ such that 
$\Efunc(\hat{\param}) - \Efunc(\param^*)$ is small, where $\param^*$ is the minimizer 
of $\Efunc$. We use $\radius$ to denote 
the $\ell_2$-radius of $\sH$, i.e. $\radius \eqdef \max_{\param \in \sH} \|\param\|_2$.

This stochastic optimization setting captures most concrete settings of interest --- 
for instance, mean estimation corresponds to $\func_i(\param) = \|\param - x_i\|_2^2$, 
linear regression to $\func_i(\param) = (y_i - \langle \param, x_i \rangle)^2$, 
and logistic regression to $\func_i(\param) = \log(1 + \exp(-y_i\langle \param, x_i \rangle))$.

\paragraph{A key player: spectral norm of gradients.}
To state our main results, we need to define the following key quantity, where 
$\|\cdot\|_{\op}$ denotes the spectral or operator norm:
\begin{equation}
\label{eq:maxop-def}
\maxop \eqdef \max_{\param \in \sH} \frac{1}{\sqrt{|\goodset|}} \left\|\left[ \nabla \func_i(\param) - \nabla \Efunc(\param) \right]_{i \in \goodset}\right\|_{\op}.
\end{equation}
In words, if we form the matrix of gradients $\left[ \nabla \func_{i_1}(\param) \ \cdots \ \nabla \func_{i_{\goodfrac n}}(\param) \right]$, 
where $\{i_1,\ldots,i_{\goodfrac n}\} = \goodset$, then $\maxop$ measures the difference between this matrix and its expectation in 
operator norm, maximized over all $\param \in \sH$. This will turn out to be a key quantity 
for understanding learnability in the adversarial setting. It acts as an analog of uniform 
convergence in classical learning theory, where one would instead study the quantity
$\max_{\param \in \sH} \|\frac{1}{|\goodset|} \sum_{i \in \goodset} (\nabla \func_i(\param) - \nabla \Efunc(\param))\|_2$. Note that this latter quantity is always 
bounded above by $\maxop$. 

The fact that $\func_i \sim p^*$ is irrelevant 
to our results---all that matters is the quantity $\maxop$, which exists even for 
a deterministic set of functions $\func_1, \ldots, \func_n$. Furthermore, $\maxop$ 
only depends on the good data and is independent of the adversary.

\paragraph{Scaling of $\maxop$: examples.}
The definition \eqref{eq:maxop-def} is a bit complex, so we go over some 
examples for intuition. We will see later that for the first two examples below
(estimating means and product distributions), our implied error bounds are ``good'', 
while for the final example (linear classification), our bounds are ``bad''.

\emph{Mean estimation:} Suppose that $f_i(w) = \frac{1}{2}\|w - x_i\|_2^2$, where $x_i \sim \sN(\mu, \sigma^2 I)$. 
Then $\nabla \func_i(w) - \nabla \Efunc(w) = x_i - \mu$, and so $\maxop$ is simply the maximum singular 
value of $\frac{1}{\sqrt{|\goodset|}} [x_i - \mu]_{i \in \goodset}$. This is the square root of the maximum 
eigenvalue of $\frac{1}{|\goodset|} \sum_{i \in \goodset} (x_i-\mu)(x_i-\mu)^{\top}$, 
which converges to $\sigma$ for large $n$.

\emph{Product distributions:} Suppose that $x_i$ is drawn from a product distribution 
on $\{0,1\}^d$, where the $j$th coordinate is $1$ with probability $p_j$. Let 
$f_i(w) = \sum_{j=1}^d x_{ij}\log(w_j) + (1-x_{ij})\log(1-w_j)$. In this case 
$\Efunc(w) = \sum_{j=1}^d p_j\log(w_j) + (1-p_j)\log(1-w_j)$, and $w_j^* = p_j$, 
so that $\Efunc(w) - \Efunc(w^*)$ is the KL divergence between $p$ and $w$.

The $j$th coordinate of $\nabla f_i(w) - \nabla \Efunc(w)$ is $(x_{ij} - p_j)(1/w_j + 1/(1-w_j))$. 
In particular, the matrix in the definition of $\maxop$ can be written as 
$D(w) \cdot [x_i - p]_{i \in \goodset} / \sqrt{|\goodset|}$, where $D(w)$ is a 
diagonal matrix with entries $1/w_j + 1/(1-w_j)$. Suppose that $p$ is \emph{balanced}, 
meaning that $p_j \in [1/4,3/4]$, and that we restrict $w_j \in [1/4,3/4]$ as well. 
Then $\|D(w)\|_{\op} \leq 16/3$, while the matrix $[x_i - p] / \sqrt{|\goodset|}$ 
has maximum singular value converging to $\max_{j=1}^d p_j(1-p_j) \leq \frac{1}{4}$ 
for large enough $n$. Thus $\maxop = \oo(1)$ in this setting.

\emph{Linear classification:} Suppose that $x_i \sim \sN(0,I)$ and that 
$y_i = \sign(u^{\top}x_i)$ for some unknown vector $u$. Our loss function 
is the logistic loss $f_i(w) = \log(1 + \exp(-y_i\langle w, x_i \rangle))$.
In this case $\nabla f_i(w) = -\frac{y_i}{1 + \exp(y_i \langle w, x_i \rangle)} x_i$. 
It is less obvious how to compute $\maxop$, but 
Lemma~\ref{lem:sg-1} below implies that it is $\oo(1)$.

\emph{Sub-gaussian gradients:} 
A useful general bound on $\maxop$ can be obtained assuming that the $\func_i$ have 
sub-Gaussian gradients. Recall that a random variable $X$ is 
\emph{$\sg$-sub-Gaussian} if $\bE[\exp(u^{\top}(X-\mu))] \leq \exp(\frac{\sg^2}{2}\|u\|_2^2)$, 
where $\mu = \bE[X]$. If $\nabla f_i$ is sub-Gaussian, then $\maxop = \oo(\sg)$ if 
$\goodfrac n \geq \tilde{\Omega}(d)$:
\begin{lemma}
\label{lem:sg-1}
If $\nabla \func_i(\param) - \nabla \Efunc(\param)$ is $\sg$-sub-Gaussian and $L$-Lipschitz for 
$\func_i \sim p^*$, 
and $\goodfrac n$ is at least 
$d\max(1, \log(\tfrac{\radius L}{\sg})) + \log(1/\delta)$, 
then $\maxop = \oo\p{\sg}$ with probability $1-\delta$.
\end{lemma}
In most of our concrete settings 
(such as mean estimation), sub-Gaussianity of $\nabla f_i$ corresponds to sub-Gaussianity of 
the data points $x_i \in \bR^d$. In some settings we will be able to obtain bounds on 
$\maxop$ under weaker (second-moment) conditions, as in Proposition~\ref{prop:bss-intro}.

\subsection{Main Results}
We can now state our main results.
Our first result is that, just using the untrusted data, 
we can output a small ellipse which contains a parameter attaining small error under $\Efunc$.
This meta-result leads directly to our more specific results in the list decoding and semi-verified settings.
\begin{theorem}
\label{thm:main-init}
Given $n$ data points containing a set $\goodset$ of $\goodfrac n$ data points with 
spectral norm bound $\maxop$, we can obtain an ellipse $\sE_Y = \{ \param \mid \param\param^{\top}\preceq Y \}$ such that $\tr(Y) \leq \oo\p{\frac{\radius^2}{\goodfrac}}$ and 
\vskip -0.08in
\begin{equation}
\label{eq:main-init}
\min_{{\param} \in \sE_Y} \Efunc({\param}) - \Efunc(\param^*) \leq \oo\p{\frac{\maxop\radius}{\sqrt{\goodfrac}}}.
\end{equation}
\end{theorem}
Recall here that $\radius$ is the $\ell_2$-radius of the parameter space $\sH$.
Also note that when $Y$ is invertible, $\param\param^{\top} \preceq Y$ is equivalent to 
$\param^{\top}Y^{-1}\param \leq 1$, so $Y$ really does define an ellipse. 
Theorem~\ref{thm:main-init} shows that the unverified 
data is indeed helpful, by narrowing the space of possible parameters from all of 
$\sH$ down to the small ellipse $\sE_Y$. 

To interpret the bound \eqref{eq:main-init}, consider the mean 
estimation example above, where $f_i(w) = \frac{1}{2}\|x_i - w\|_2^2$ 
with $x_i \sim \sN(\mu, \sigma^2 I)$.
Note that $\Efunc(w) - \Efunc(w^*) = \frac{1}{2}\|w - \mu\|_2^2$. 
Assuming that $\|\mu\|_2$ is known to within a constant factor, 
we can take $\sH$ to be the $\ell_2$-ball of radius $\radius = \|\mu\|_2$. This leads to the bound 
$\|w - \mu\|_2^2 = \oo\p{\sigma \|\mu\|_2 / \sqrt{\goodfrac}}$, for some $w$ in an ellipse of 
trace $\frac{\|\mu\|_2^2}{\alpha}$. Note that the $\ell_2$-ball itself has trace $d\|\mu\|_2^2$, 
so the ellipse $\sE_Y$ is much smaller than $\sH$ if $d$ is large. Moreover, a random $x_i \sim p^*$ 
will have $\|x_i - \mu\|_2^2 \approx d\sigma^2$, so $w$ is much closer to $\mu$ than a random draw 
from $p^*$, assuming that $\|\mu\|_2 \ll d\sigma$.

By applying our algorithm multiple times we can improve the bound 
$\|w - \mu\|_2^2 = \oo\p{\sigma \|\mu\|_2 / \sqrt{\goodfrac}}$ to $\|w- \mu\|_2^2 = \oo\p{\sigma^2 / \goodfrac}$, so that our bounds are meaningful for large $d$ independent of $\|\mu\|_2$. 
We discuss this in more detail in Section~\ref{sec:strongly-convex}.

For an example where Theorem~\ref{thm:main-init} is less meaningful, consider 
the linear classification example from above. In that case $\maxop = \oo(1)$, 
and $\radius$ is likely also $\oo(1)$, so we obtain the bound 
$\Efunc(w) - \Efunc(w^*) = \oo(1/\sqrt{\goodfrac})$. However, 
$\Efunc(0) = \log(2)$ while $\Efunc(w^*) \geq 0$, so this bound is essentially vacuous.

\paragraph{List decodable learning.}
Using Theorem~\ref{thm:main-init} as a starting point we can derive bounds 
for both models defined in Section~\ref{sec:intro}, starting with the 
list decodable model. Here, we must make the further assumption that the 
$\func_i$ are $\kappa$-strongly convex, meaning that 
$\func_i(\param') - \func_i(\param) \geq (\param'-\param)^{\top}\nabla \func_i(\param) + \frac{\kappa}{2}\|\param'-\param\|_2^2$.
The strong convexity allows us to show that for the good $\func_i$, 
the parameters $\hat{\param}_i = \argmin_{\param \in \sE_Y} \func_i(\param)$ 
concentrate around $\param^*$, with radius $\radius' \ll \radius$. 
By clustering the $\hat{\param}_i$ and iteratively re-running our 
algorithm on each cluster, we can obtain bounds that do not depend on $\radius$,
and output a single candidate parameter $\hat{\param}_j$ for each cluster.
We can thereby show: 
\begin{theorem}
\label{thm:main-init-convex}
Suppose the functions $\func_i$ are $\kappa$-strongly convex, and suppose there 
is a set $\goodset$ of size $\goodfrac n$ with spectral norm bound $\maxop$. 
Then, for any $\opfrac \leq \frac{1}{2}$, it is possible to obtain a set 
of $m \leq \lfloor \frac{1}{(1-\opfrac)\goodfrac} \rfloor$ candidate parameters 
$\hat{\param}_1, \ldots, \hat{\param}_m$, such that 
$\min_{j=1}^m \|\hat{\param}_j - \param^*\|_2 \leq \oo\p{\frac{\maxop}{\kappa}\sqrt{\frac{\log(\frac{2}{\goodfrac})}{\goodfrac\opfrac}}}$.
\end{theorem}
\noindent In Section~\ref{sec:strongly-convex} we state and prove a stronger version of 
this result. 
A key tool in establishing Theorem~\ref{thm:main-init-convex} is 
\emph{padded decompositions} \citep{fakcharoenphol2003tight}, 
which identify clusters in data while 
making minimal assumptions on the geometry of points outside of a cluster, and 
are thus useful in our adversarial setting.

\paragraph{Semi-verified learning.}
If the $\func_i$ are not strongly convex then we cannot employ 
the clustering ideas above. However, because we have reduced 
$\sH$ to the much smaller set $\sE_Y$, we 
can nevertheless often approximate $\param^*$ with only a small 
amount of verified data. In fact, in some settings we only need 
a single verified data point:
\begin{lemma}
\label{lem:semi-verified}
Suppose that $\func(\param) = \phi(\param^{\top}x)$, where 
$\phi$ is $1$-Lipschitz, and suppose that $x$ has bounded $q$th 
moments in the sense that $\bE_{p^*}[|\langle x - \bE[x], u \rangle|^q]^{1/q} \leq \sigma_q$ 
for all unit vectors $u$ and some $q \geq 2$. 
Then given $Y$ from Theorem~\ref{thm:main-init} and a single verified $x \sim p^*$, we can obtain a $\hat{\param}$ such that
$\bP_{x \sim p^*}\left[\Efunc(\hat{\param}) \geq \Efunc(\param^*) + C \cdot \frac{(\maxop + t \sigma_q)\radius}{\sqrt{\goodfrac}}\right] \leq t^{-q}$, 
for a universal constant $C$.
\end{lemma}
The particular functional form for $\func_i$ was needed to obtain a concrete 
bound, but analogs of Lemma~\ref{lem:semi-verified} should be possible in 
any setting where we can leverage the low complexity of $\sE_Y$
into a bound on $\func - \Efunc$.
Note that if we replace $\sE_Y$ with $\sH$ in Lemma~\ref{lem:semi-verified}, then 
the $\frac{\radius}{\sqrt{\goodfrac}}$ dependence becomes $\radius\sqrt{d}$, which 
is usually vacuous.


\paragraph{Optimality?} The dependence on $\maxop$, $\radius$ and $\kappa$ in the 
results above seems essentially 
necessary, though the optimal dependence on $\goodfrac$ is less clear. In Section~\ref{sec:lower-bounds} 
we show lower bounds for robust mean estimation even if $p^*$ is known to be Gaussian. 
These bounds roughly translate to a lower bound of $\Omega\big({\frac{\maxop}{\kappa}{\sqrt{\log{({1}/{\goodfrac})}}}}\big)$ 
for strongly convex $\func_i$, and $\Omega\big({\maxop\radius\sqrt{\log(\smash{{1}/{\goodfrac}})}}\big)$ for 
linear $\func_i$, and hold in both the list decodable and semi-verified settings. 
It is unclear whether the optimal dependence on $\goodfrac$ is $\sqrt{{1}/{\goodfrac}}$ or 
$\sqrt{\log({1}/{\goodfrac})}$ or somewhere between. We do note that any dependence better than 
$\sqrt{{1}/{\goodfrac}}$ would improve the best known results for 
efficiently solving $k$-means 
for well-separated clusters, which may suggest at least a computational barrier to 
achieving $\sqrt{\log({1}/{\goodfrac})}$.

\subsection{Implications}
We now go over some implications of our general results in some 
more specific settings. All of the results below follow as corollaries 
of our main theorems, and are proved in Section~\ref{sec:applications}.

\paragraph{Robust mean estimation.}
Suppose we observe points $x_1, \ldots, x_n \in \bR^d$, 
of which $\goodfrac n$ are drawn from a distribution $p^*$ with bounded covariance, 
and our goal is to recover the mean $\mu = \bE_{x \sim p^*}[x]$. 
If we take $\func_i(\param) = \|\param - x_i\|_2^2$, then 
Theorem~\ref{thm:main-init-convex}, together with the matrix concentration bound 
Proposition~\ref{prop:bss-intro}, implies the following result:
\begin{corollary}
\label{cor:gaussian-init}
Suppose that $p^*$ has bounded covariance: $\Cov_{p^*}[x] \preceq \sigma^2 I$. 
Then, for $n \geq \frac{d}{\goodfrac}$, with probability $1 - \exp\p{-\Omega(\goodfrac n)}$ it is 
possible to output $m \leq \oo\p{\frac{1}{\goodfrac}}$ 
candidate means $\hat{\mu}_1, \ldots, \hat{\mu}_m$ such that
$\min_{j=1}^m \|\mu - \hat{\mu}_j\|_2 \leq \oo\Big({\sigma\sqrt{\frac{\log(2/\goodfrac)}{\goodfrac}}}\Big)$.
Moreover, if $\goodfrac \geq 0.51$ then we can take $m = 1$.
\end{corollary}
We can compare to the results of \citet{lai2016agnostic} and 
\citet{diakonikolas2016robust}, who study mean estimation when
$\goodfrac > \frac{1}{2}$ and one is required to output a single parameter (i.e., $m=1$). 
For simplicity take $\goodfrac = \frac{3}{4}$.
Roughly, \citet{lai2016agnostic} obtain error $\oo({\sigma\sqrt{\log(d)}})$ with sample 
complexity $n = \oo\p{d}$, while requiring a bound on the fourth moments of $p^*$; 
\citet{diakonikolas2016robust} obtain error 
$\oo\p{\sg}$ with sample complexity $n = \oo\p{d^3}$, and require $p^*$ to be Gaussian.
Corollary~\ref{cor:gaussian-init} improves both of these by yielding error $\oo\p{\sg}$ 
with sample complexity $n = \oo\p{d}$, and only requires $p^*$ to have bounded second moments.%
\footnote{
Some fine print: \citeauthor{diakonikolas2016robust} also estimate the covariance matrix, and 
their recovery results are stronger if $\Sigma = \Cov[x]$ is highly skewed; they roughly show 
$\|\hat{\mu} - \mu\|_{\Sigma^{-1}} = \oo(1)$.
The adversary model considered in both of these other papers is also slightly more general than ours:
first $n$ points are drawn from $p^*$ and then an adversary is allowed to corrupt 
$(1-\goodfrac)n$ of the points. However, it is straightforward to show (by monotonicity 
of the operator norm) that our bounds will be worse by at most a $1/\sqrt{\goodfrac}$ 
factor in this stricter setting, which is a constant if $\goodfrac = \frac{3}{4}$.
} 
We note that in contrast to our results, these other results obtain error that vanishes 
as $\goodfrac \to 1$ (at a rate of $(1-\goodfrac)^{1/2}$ in the first case and 
$\oot(1-\goodfrac)$ in the second case). We thus appear to incur some looseness when 
$\goodfrac \approx 1$, in exchange for obtaining results in the previously unexplored 
setting $\goodfrac \leq \frac{1}{2}$.
It would be interesting to obtain a single algorithm that both applies when 
$\goodfrac \ll 1$ and achieves vanishing error as $\goodfrac \to 1$.


\paragraph{Learning mixture of distributions.}
In addition to robust mean estimation, we can use our results to efficiently learn 
mixtures of distributions, by thinking of a single mixture component as 
the good data and the remaining mixture components as bad data.
%
Again applying Theorem~\ref{thm:main-init} to $\func_i(\param) = \|\param - x_i\|_2^2$, 
we obtain the following result, which says that we can successfully cluster 
samples from a mixture of distributions, even in the presence of 
arbitrary corruptions, provided the cluster means are separated in $\ell_2$ distance 
by $\tilde{\Omega}({{\sigma}/{\sqrt{\goodfrac}}})$.
\begin{corollary}
\label{cor:mixture-init}
\vskip -0.01in
Suppose we are given $n$ samples, where each sample either comes from one of 
$k$ distributions $p_1^*, \ldots, p_k^*$ 
(with $\Cov_{p_i^*}[x] \preceq \sigma^2 I$ for all $i$), or is arbitrary. 
Let $\mu_i$ be the mean of $p_i^*$, let $\goodfrac_i$ be the fraction of points from $p_i^*$, and let 
$\goodfrac = \min_{i=1}^k \goodfrac_i$. 
%
Then if $n \geq \frac{d}{\goodfrac}$, with probability $1-k\exp(-\Omega(\goodfrac \opfrac^2 n))$ 
we can obtain a partition 
$T_1, \ldots, T_m$ of $[n]$ and corresponding candidate means 
$\hat{\mu}_1, \ldots, \hat{\mu}_m$ such that:
for all but $\opfrac \goodfrac_i n$ of the points drawn from $p_i^*$, 
the point is assigned a set $T_j$ and candidate mean $\hat{\mu}_j$ with
$\|\mu_i - \hat{\mu}_j\|_2 \leq \oo\p{\frac{\sigma}{\opfrac}\sqrt{\frac{\log(\frac{2}{\goodfrac})}{\goodfrac}}}$.
Moreover, $m \leq \oo\p{\frac{1}{\goodfrac}}$.
\end{corollary}
The $\frac{1}{\opfrac}$ dependence can be replaced with $\sqrt{\log(n)/\opfrac}$, 
or with $\sqrt{\log(2/\opfrac)}$ if the $x_i$ are sub-Gaussian.
The only difference is in which matrix concentration bound we apply to the $x_i$.
Corollary~\ref{cor:mixture-init} says that we can partition the points 
into $\oo\p{\frac{1}{\goodfrac}}$ sets, such that two points from 
well-separated clusters are unlikely to end up in the same set. Note 
however that one cluster might be partitioned into multiple sets. 
In the adversarial setting, this seems unavoidable, since an adversary 
could create a fake cluster very close to a real cluster, and one would 
be forced to either combine the clusters or risk splitting the real cluster 
in two.


For comparison, the best computationally efficient algorithm for 
clustering a mixture of $k$ distributions (with few/no corruptions) 
requires mean separation roughly $\tilde{\oo}({\sigma \sqrt{k}})$ 
\citep{awasthi2012improved,achlioptas2005spectral}, which our result 
matches if $\goodfrac = \Omega({1}/{k})$.


\paragraph{Planted partitions.}
We next consider implications of our results in a version of 
the planted partition model \citep{mcsherry2001spectral}. 
In this model we observe a random directed graph, represented as a matrix 
$A \in \{0,1\}^{n \times n}$. For disjoint subsets $I_1, \ldots, I_k$ of $[n]$, we generate 
edges as follows:
(i) If $u, v \in I_i$, then $p(A_{uv} = 1) = \frac{a}{n}$.
(ii) If $u \in I_i, v \not\in I_i$, then $p(A_{uv} = 1) = \frac{b}{n}$.
(iii) If $u \not\in \cup_{i=1}^k I_i$, the edges emanating from $u$ can be arbitrary.
In contrast to the typical planted partition model, we allow
some number of corrupted vertices not belonging to any of the $I_i$.
In general $a$ and $b$ could depend on the partition indices $i$, $j$, 
but we omit this for simplicity.

Note that the distribution over the row $A_{u}$ is the same for all $u \in I_i$.
By taking this distribution to be the distribution $p^*$, 
Theorem~\ref{thm:main-init-convex} yields the following result:
\begin{corollary}
\label{cor:partition-init}
For the planted partition model above, let 
$\goodfrac = \min_{i=1}^k \frac{|I_i|}{n}$.
Then, with probability $1 - \exp(-\Omega(\goodfrac n))$,
we can obtain sets 
$T_1, \ldots, T_m \subseteq [n]$, 
with $m \leq \oo\p{\frac{1}{\goodfrac}}$, such that 
for all $i \in [k]$, there is a $j \in [m]$ with
$|I_i \triangle T_j| \leq \oo\p{\frac{a\log(\frac{2}{\goodfrac})}{\goodfrac^2(a-b)^2}}n$, 
where $\triangle$ denotes symmetric difference.
\end{corollary}
This shows that we can approximately recover the planted partition, 
even in the presence of arbitrary corruptions, provided 
$\frac{(a-b)^2}{a} \gg \frac{\log(2/\goodfrac)}{\goodfrac^3}$ 
(since the bound on $|I_i \triangle T_j|$ needs 
to be less than $\goodfrac n$ to be meaningful). 
In contrast, the best efficient methods (assuming no corruptions) 
roughly require 
$\frac{(a-b)^2}{a + (k-1)b} \gg k$ in the case of 
$k$ equal-sized communities \citep{abbe2015community,abbe2015detection}. 
In the simplifying setting 
where $b = \frac{1}{2}a$, our bounds require 
$a \gg k^3\log(k)$ while existing bounds require $a \gg k^2$.
The case of unequal size communities is more complex, 
but roughly, our bounds require $a \gg \frac{\log(2/\goodfrac)}{\goodfrac^3}$ 
in contrast to $a \gg \frac{1}{\goodfrac^2}$.


\paragraph{Summary.} For robust mean estimation, 
we match the best existing error bounds of $\oo\p{\sigma}$ when 
$\goodfrac = \frac{3}{4}$, under weaker assumptions. 
For learning mixtures distributions, we 
match the best bound of $\tilde{\oo}(\sigma\sqrt{k})$ when $\goodfrac = \Omega(1/k)$. 
For recovering planted partitions, we require average degree 
$k^3\log(k)$, in contrast to the best known bound of $k^2$.
It is pleasing that a single meta-algorithm 
is capable of matching or nearly matching the best rate 
in these settings, despite allowing for arbitrary corruptions.
We can also achieve bounds for robust density estimation, 
presented in Section~\ref{sec:applications}.

\section{Algorithm}
\label{sec:setting}
\label{sec:algorithm}

In this section we present our algorithm, which consists of 
an SDP coupled with an outlier removal step.
At a high level, our algorithm works as follows: first, we give 
\emph{each function} $\func_i$ its own parameter vector $\param_i$, 
and minimize $\sum_{i=1}^n \func_i(\param_i)$ subject to regularization 
which ensures the $\param_i$ remain close to each other; formally, 
we bound the $\param_i$ to lie within a small ellipse. The reason 
for doing this is that the different $\param_i$ are now only coupled 
via this regularization, and so the influence of adversarial data on the 
good parameters can only come from its effect on the shape of the ellipse. 
We will show that whenever the adversaries affect the shape of the ellipse 
more than a small amount, they are necessarily outliers that can be 
identified and removed. In the remainder of this section, we elaborate 
on these two steps of regularization and outlier removal, and provide 
pseudocode.

\begin{algorithm}[b!]
\setlength{\abovedisplayskip}{0pt}
\setlength{\belowdisplayskip}{0pt}
\setlength{\abovedisplayshortskip}{0pt}
\setlength{\belowdisplayshortskip}{0pt}
\caption{Algorithm for fitting $p^*$}
\label{alg:main}
\begin{algorithmic}[1]
\State Input: $f_1,\ldots,f_n$
\State Initialize $\weight \gets [1; \cdots; 1] \in \bR^n$
\State Set $\lambda \gets \frac{\sqrt{8\goodfrac}n\maxop}{\radius}$
\While{{\bfseries true}}
  \State Let $\hat{\param}_{1:n}, \hat{Y}$ be the solution to
    \begin{align}
    \notag \underset{\param_1,\ldots,\param_n,Y}{\text{minimize}} \ & \sum_{i=1}^n \weight_i \func_i(\param_i) + \lambda \tr(Y) \\
    \label{eq:optimization-alg}
    \text{subject to} \ & \param_i\param_i^{\top} \preceq Y \text{ for all } i = 1, \ldots, n.
    \end{align}
  \If{$\tr(\hat{Y}) \leq \frac{6\radius^2}{\goodfrac}$} \Comment{Check for outliers}
    \State \Return $\hat{\param}_{1:n}$, $\hat{Y}$ \Comment{Not many outliers, can return}
  \Else
    \State $\weight \gets \textsc{UpdateWeights}(\weight, \hat{\param}_{1:n}, \hat{Y})$ \Comment{Re-weight points to down-weight outliers}
  \EndIf
\EndWhile
\end{algorithmic}
\end{algorithm}

\paragraph{Per-function adaptivity.}
If the functions $\func_1, \ldots, \func_n$ were all drawn from $p^*$ (i.e., there are 
no adversaries), then a natural approach would be to let $\hat{\param}$ be the minimizer 
of $\sum_{i=1}^n \func_i(\param)$, which will approximately minimize 
$\Efunc(\param)$ by standard concentration results. 

The problem with using this approach 
in the adversarial setting is that even a single adversarially chosen function $\func_i$ 
could substantially affect the value of $\hat{\param}$. 
To minimize this influence, we give each $\func_i$ 
its own parameter $\param_i$, and 
minimize $\sum_{i=1}^n \func_i(\param_i)$, subject to a regularizer which encourages 
the $\param_i$ to be close together. 
%
The 
adversary now has no influence on the good $\param_i$ except via the regularizer, 
so the key challenge 
is to find a regularizer which sufficiently controls statistical error while also bounding 
the influence of the adversary.

It turns out that the right choice of regularizer in this case is to constrain the 
$\param_i$ to lie within an \emph{ellipse with small trace}. Formally, the centerpiece 
of our algorithm is the following convex optimization problem:
\begin{align}
\notag \underset{\param_1,\ldots,\param_n,Y}{\text{minimize}} \ & \sum_{i=1}^n \weight_i \func_i(\param_i) + \lambda \tr(Y) \\
\label{eq:optimization-1}
\text{subject to} \ & \param_i\param_i^{\top} \preceq Y \text{ for all } i = 1, \ldots, n.
\end{align}
Here the coefficients $\weight_i$ are non-negative weights which will eventually 
be used to downweight outliers (for now it is fine to imagine that $\weight_i = 1$).
Note that $w_iw_i^{\top} \preceq Y$ is equivalent to the semidefinite constraint 
$\left[ \begin{array}{cc} Y & w_i \\ w_i^{\top} & 1 \end{array} \right] \succeq 0$.
The problem \eqref{eq:optimization-1} can be solved in polynomial time in $n$ and $d$ 
assuming oracle access to the gradients $\nabla \func_i$.

Note that the semidefinite constraint $\param_i\param_i^{\top} \preceq Y$ is equivalent 
to $\param_i^{\top}Y^{-1}\param_i \leq 1$, which says that $\param_i$ lies within the 
ellipse centered at $0$ defined by $Y$. 
The regularizer is thus the trace of the minimum ellipse containing the $\param_i$; 
penalizing this trace will tend to push the $\param_i$ closer together, 
but is there any intuition behind its geometry? 
The following lemma shows that $\tr(Y)$ is related to the trace norm 
of $\left[ \param_1 \ \cdots \ \param_n \right]$:
\begin{lemma}
\label{lem:trace}
For any points $\param_1, \ldots, \param_n \in \bR^d$, suppose that $Y \succeq \param_i\param_i^{\top}$ for all 
$i$. Then, letting $\|\cdot\|_*$ denote the trace norm (i.e., sum of singular values) and 
$\Param_T = \left[\param_i\right]_{i \in T}$, we have
$\tr(Y) \geq \frac{\|\Param_T\|_*^2}{|T|}$ for all sets $T \subseteq [n]$.
\end{lemma}
The appearance of the trace norm makes sense in light of the intuition that we should 
be clustering the functions $\func_i$; indeed, trace norm regularization is a key ingredient 
in spectral algorithms for clustering (see e.g. \citet{zha2001spectral,chen2014clustering,chen2014improved}).
Lemma~\ref{lem:trace} says that $\tr(Y)$ simultaneously bounds the trace norm on 
every subset $T$ of $[n]$, which ends up yielding better results than are 
obtained by simply penalizing the overall trace norm; we believe that this 
\emph{local trace norm regularization} likely leads to better results even 
in non-adversarial spectral learning settings.
The most important property of $\tr(Y)$ is that it admits a certain type of 
local H\"{o}lder's inequality which we will explain in Section~\ref{sec:local-holder}.

\begin{algorithm}[b!]
\setlength{\abovedisplayskip}{6pt}
\setlength{\belowdisplayskip}{0pt}
\setlength{\abovedisplayshortskip}{0pt}
\setlength{\belowdisplayshortskip}{0pt}
\caption{Algorithm for updating weights $\weight$ to downweight outliers.}
\label{alg:outlier}
\begin{algorithmic}[1]
\Procedure{UpdateWeights}{$\weight$, $\hat{\param}_{1:n}$, $\hat{Y}$}
\For{$i = 1, \ldots, n$}
\State Let $\tilde{\param}_i$ be the solution to
    \begin{align}
    \notag \underset{\tilde{\param}_i, a_{i1}, \ldots, a_{in}}{\text{minimize}} \ \ & \func_i(\tilde{\param}_i) \\[-1ex]
    \label{eq:param-tilde-alg}
    \text{subject to} \ \ & \tilde{\param}_i = \sum_{j=1}^n a_{ij} \hat{\param}_j, \quad 0 \leq a_{ij} \leq \frac{2}{\goodfrac n}, \quad\quad \sum_{j=1}^n a_{ij} = 1.
    \end{align}
\State Let $z_i \gets \func_i(\tilde{\param}_i) - \func_i(\hat{\param}_i)$
\EndFor
\State $z_{\max} \gets \max\{z_i \mid \weight_i \neq 0\}$
\State $\weight_i' \gets \weight_i \cdot \frac{z_{\max} - z_i}{z_{\max}}$ for $i=1,\ldots,n$
\State \Return $\weight'$
\EndProcedure
\end{algorithmic}
\end{algorithm}

\paragraph{Removing outliers.}
Solving \eqref{eq:optimization-1} is not by itself sufficient to achieve 
robustness. The problem is that a single function $\func_i$ 
could strongly push $\param_i$ to a given value $\param_{\mathrm{target}}$ (e.g. if 
$\func_i(\param) = 10^{100} \|\param - \param_{\mathrm{target}}\|_2^2$) 
which allows the adversaries to arbitrarily expand the ellipse defined by $Y$. To 
combat this, we need some way of removing outlier functions $\func_i$ from 
our dataset. We will do this in a soft way, by assigning a weight 
$\weight_i$ to each function $\func_i$, and downweighting functions that seem 
likely to be outliers.

How can we tell that a function is an outlier? Intuitively, if a function 
$\func_i$ is really drawn from $p^*$, then there should be many other functions 
$\func_j$, $j \neq i$, that are ``similar'' to $\func_i$. We can quantify 
this by considering whether there are a large number of $j \neq i$ for 
which the parameter $\param_j$ for $\func_j$ does a 
good job of minimizing $\func_i$. Formally, 
given a solution $(\hat{\param}_1, \ldots, \hat{\param}_n)$ to 
\eqref{eq:optimization-1}, we compare $\hat{\param}_i$ to $\tilde{\param}_i$, 
which is defined as the solution to the following optimization:
\vskip -0.22in
\begin{align}
\notag \underset{\tilde{\param}_i, a_{i1}, \ldots, a_{in}}{\text{minimize}} \ \ & \func_i(\tilde{\param}_i) \\[-0.8ex]
\label{eq:param-tilde-def}
\text{subject to} \ \ & \tilde{\param}_i = \sum_{j=1}^n a_{ij} \hat{\param}_j, \quad 0 \leq a_{ij} \leq \frac{2}{\goodfrac n}, \quad\quad \sum_{j=1}^n a_{ij} = 1.
\end{align}
\vskip -0.06in
The optimization \eqref{eq:param-tilde-def} roughly asks for a parameter $\tilde{\param}_i$ that 
minimizes $\func_i$, subject to $\tilde{\param}_i$ being the average of at least 
$\frac{\goodfrac n}{2}$ distinct parameters $\hat{\param}_j$.
Given the solution $\tilde{\param}_i$ to \eqref{eq:param-tilde-def}, we then downweight 
the influence of the $i$th data point based on the value of 
$\func_i(\tilde{\param}_i) - \func_i(\hat{\param}_i)$. In particular, 
we will multiply the weight $\weight_i$ by
$1 - \eta \p{\func_i(\tilde{\param}_i) - \func_i(\hat{\param}_i)}$ 
for some appropriate $\eta$. Hopefully, this will downweight any outliers
by a large amount while only downweighting good points by a small amount 
(this hope is verified in Lemma~\ref{lem:outlier-1} below).

Pseudocode for our algorithm is 
given in Algorithms~\ref{alg:main} and \ref{alg:outlier}.

\section{Approach and Proof Outline}
\label{sec:proof-outline}

We now provide an outline of the proof of Theorem~\ref{thm:main-init}, by analyzing the 
output of Algorithm~\ref{alg:main}. 
The structure of our proof has analogies to classical 
uniform convergence arguments, so we will start by reviewing that case.

\subsubsection*{Warm-up: Uniform Convergence}
 
In uniform convergence arguments, we assume that all of 
$\func_1, \ldots, \func_n$ are drawn from $p^*$, which brings us into the realm of classical 
learning theory. The analogue to the optimization \eqref{eq:optimization-alg} in Algorithm~\ref{alg:main} 
is regularized empirical risk minimization:
\begin{equation}
\label{eq:erm}
\hat{\param} = \underset{w \in \sH}{\argmin} \ \sum_{i=1}^n \func_i(\param) + \lambda h(\param),
\end{equation}
where $h(\param)$ is a non-negative regularizer.
Uniform convergence arguments involve two parts:
\begin{enumerate}
\item \textbf{Bound the optimization error}:
      Use the definition of $\hat{\param}$ to conclude that 
      $\sum_{i=1}^n \func_i(\hat{\param}) \leq \sum_{i=1}^n \func_i(\param^*) + \lambda h(\param^*)$ 
      (since $\hat{\param}$ minimizes \eqref{eq:erm}). This step shows that 
      $\hat{\param}$ does almost as well as $\param^*$ at minimizing the empirical risk 
      $\sum_{i=1}^n \func_i(\param)$.
\item \textbf{Bound the statistical error}: Show, via an appropriate concentration inequality, 
      that $\frac{1}{n} \sum_{i=1}^n \func_i(\param)$ is close to $\Efunc(\param)$ for all 
      $\param \in \sH$. Therefore, $\hat{\param}$ is nearly as good as $\param^*$ in terms 
      of the true risk $\Efunc$.
\end{enumerate}
We will see next that the proof of Theorem~\ref{thm:main-init} contains steps similar 
to these, though bounding the statistical error in the presence of adversaries requires 
an additional step of removing outliers.
 
\subsubsection*{Proof Overview}

We will establish 
a stronger version of Theorem~\ref{thm:main-init}, 
which exhibits an explicit $\param \in \sE_{\hat{Y}}$ with small error:
\begin{theorem}
\label{thm:main}
Let $\hat{\param}_{1:n}$, $\hat{Y}$ be the output of 
Algorithm~\ref{alg:main}, and let 
$\hat{\param}_{\avg} = \p{\sum_{i \in \goodset} \weight_i\hat{\param}_i}/\p{\sum_{i \in \goodset} \weight_i}$. Then, $\Efunc(\hat{\param}_{\avg}) - \Efunc(\param^*) \leq 18\frac{\maxop\radius}{\sqrt{\goodfrac}}$. Furthermore, $\hat{\param}_{\avg} \in \sE_{\hat{Y}}$ and $\tr(\hat{Y}) \leq \frac{6\radius^2}{\goodfrac}$.
\end{theorem}
To prove Theorem~\ref{thm:main}, recall that Algorithm~\ref{alg:main}
has at its core the following optimization problem:
\begin{align}
\notag \underset{\param_1,\ldots,\param_n,Y}{\text{minimize}} \ & \sum_{i=1}^n \weight_i \func_i(\param_i) + \lambda \tr(Y) \\
\label{eq:optimization-2}
\text{subject to} \ & \param_i\param_i^{\top} \preceq Y \text{ for all } i = 1, \ldots, n.
\end{align}
This optimization asks to minimize $\sum_{i=1}^n \weight_i \func_i(\param_i)$ while constraining 
the $\param_i$ to lie within the ellipse defined by $Y$. As in the uniform convergence argument 
above, there are two sources of 
error that we need to bound: the \emph{optimization error} 
$\sum_{i \in \goodset} \weight_i \p{\func_i(\hat{\param}_i) - \func_i(\param^*)}$, 
and the \emph{statistical error} $\sum_{i \in \goodset} \weight_i \p{\Efunc(\hat{\param}_{\avg}) - \func_i(\hat{\param}_i)}$. 
Note that the statistical error now measures two quantities: the distance from $\func_i(\hat{\param}_i)$ 
to $\func_i(\hat{\param}_{\avg})$, and from $\func_i(\hat{\param}_{\avg})$ to $\Efunc(\hat{\param}_{\avg})$.

Bounding the optimization error requires showing that the 
ellipse defined by $\hat{Y}$ is not too small (so that it contains $\param^*$), 
while bounding the statistical error requires showing that the ellipse is not too large (so that 
we cannot overfit too much). 
The former turns out to be easy and is 
shown in Lemma~\ref{lem:optimization-error-1}. The latter is more involved and requires 
several steps. First, we show in Lemma~\ref{lem:statistical-error-1} that 
the statistical error can be bounded in terms of $\tr(Y)$ and $\maxop$, which verifies 
the intuition that bounding the statistical error reduces to bounding $Y$. 
Next, in Lemma~\ref{lem:w-tilde-bound-1}, we show that the parameters $\tilde{\param}_i$ 
found in Algorithm~\ref{alg:outlier} are bounded by an ellipse $\tilde{Y}$ with small trace, 
and that $\func_i(\tilde{\param}_i) \approx \func_i(\hat{\param}_i)$ for $i \in \goodset$. 
By the optimality of $(\hat{\param}_{1:n}$, $\hat{Y})$ for \eqref{eq:optimization-2}, the only way that 
$\tr(\hat{Y})$ can be much larger than $\tr(\tilde{Y})$ is therefore if 
$\func_i(\hat{\param}_i) \ll \func_i(\tilde{\param}_i)$ for $i \not\in \goodset$. 
In this case, we can identify outliers $i \not\in \goodset$ by considering the value of
$\func_i(\tilde{\param}_i) - \func_i(\hat{\param}_i)$, and Lemma~\ref{lem:outlier-1} 
verifies that we can use this to perform outlier removal. We expand on both the 
optimization error and statistical error bounds below.

\paragraph{Bounding optimization error on $\goodset$.} 
Throughout the argument, we will make use of the optimality of $(\hat{\param}_{1:n}, \hat{Y})$ 
for \eqref{eq:optimization-2}, which implies that
\begin{equation}
\label{eq:optimality-feasibility}
\sum_{i=1}^n \weight_i \func_i(\hat{\param}_i) + \lambda \tr(\hat{Y}) \leq 
\sum_{i=1}^n \weight_i \func_i(\param_i) + \lambda \tr(Y)
\text{ for any feasible $(\param_{1:n}, Y)$.}
\end{equation}
We wish to bound 
$\sum_{i \in \goodset} \weight_i \func_i(\hat{\param}_i)$, but the preceding bound involves 
all of $\sum_{i=1}^n \weight_i \func_i(\hat{\param}_i)$, not just the 
$\func_i$ for $i \in \goodset$. 
However, because the $\hat{\param}_i$ are free to vary independently, 
we can bound $\sum_{i \in \goodset} \weight_i \p{\func_i(\hat{\param}_i) - \func_i(\param^*)}$ 
in terms of the amount that $\tr(\hat{Y})$ would need to increase before 
$\param^*(\param^*)^{\top} \preceq \hat{Y}$. In particular, by taking 
$Y = \hat{Y} + (\param^*)(\param^*)^{\top}$ in \eqref{eq:optimality-feasibility}, 
we can obtain the following bound on the optimization error:
\begin{lemma}
\label{lem:optimization-error-1}
The solution $\hat{\param}_{1:n}$ to \eqref{eq:optimization-2} satisfies
\begin{equation}
\label{eq:optimization-error-1}
\sum_{i \in \goodset} \weight_i \p{\func_i(\hat{\param}_i) - \func_i(\param^*)} \leq \lambda \|\param^*\|_2^2.
\end{equation}
\end{lemma}

\paragraph{Bounding the statistical error.}
We next consider the statistical error. 
We cannot bound this error via standard uniform convergence techniques, 
because each $\func_i$ has a different argument $\hat{\param}_i$. However, 
it turns out that the operator norm bound $\maxop$, together 
with a bound on $\tr(\hat{Y})$, yield concentration of the $\func_i$ to $\Efunc$. 
In particular, we have:
\begin{lemma}
\label{lem:statistical-error-1}
Let $\hat{\param}_{\avg} \eqdef \frac{\sum_{i \in \goodsete} \weight_i \hat{\param}_i}{\sum_{i \in \goodset} \weight_i}$. 
Then the solution $\hat{\param}_{1:n}$, $\hat{Y}$ to \eqref{eq:optimization-2} satisfies
\begin{align}
\label{eq:statistical-error-1a}
\sum_{i \in \goodset} \weight_i \p{\func_i(\hat{\param}_{\avg}) - \func_i(\hat{\param}_i)} &\leq \goodfrac n \maxop \p{\sqrt{\tr(\hat{Y})} + \radius}, \text{ and } \\
\label{eq:statistical-error-1b}
\sum_{i \in \goodset} \weight_i \p{\Efunc(\hat{\param}_{\avg}) - \Efunc(\param^*)} &\leq \sum_{i \in \goodset} \weight_i \p{\func_i(\hat{\param}_{\avg}) - \func_i(\param^*)} + 2\goodfrac n\radius\maxop.
\end{align}
\end{lemma}
Lemma~\ref{lem:statistical-error-1} relates $\func_i(\hat{\param}_i)$ to 
$\func_i(\hat{\param}_{\avg})$ in \eqref{eq:statistical-error-1a}, 
and then relates $\func_i(\hat{\param}_{\avg})$ to $\Efunc(\hat{\param}_{\avg})$ 
in \eqref{eq:statistical-error-1b}. 
Together these allow us to bound the statistical error in terms of $\tr(\hat{Y})$ and $\maxop$.
The proof is an application of the matrix H\"{o}lder's inequality 
$|\tr(A^{\top}B)| \leq \|A\|_*\|B\|_{\op}$, with $A_i = \hat{\param}_i - \hat{\param}_{\avg}$ and 
$B_i = \nabla \func_i(\hat{\param}_{\avg}) - \nabla \Efunc(\hat{\param}_{\avg})$.

\paragraph{Bounding the trace.}
We next need to bound $\tr(\hat{Y})$ itself. We again exploit the optimality 
constraint \eqref{eq:optimality-feasibility}, which implies that 
$\tr(\hat{Y}) \leq \tr(Y) + \frac{1}{\lambda}\p{\sum_{i=1}^n \weight_i \p{\func_i(\param_i) - \func_i(\hat{\param}_i)}}$ 
for any feasible $(\param_{1:n}, Y)$.
We will take $\param_{1:n}$ to be 
$\tilde{\param}_{1:n}$ as defined in equation \eqref{eq:param-tilde-alg} of 
Algorithm~\ref{alg:outlier}; we then take $Y$ to be 
$\frac{2}{\goodfrac n} \hat{\Param}\hat{\Param}^{\top}$, where 
$\Param = \left[ \param_1 \ \cdots \ \param_n \right]$. Lemma~\ref{lem:w-tilde-bound-1} 
asserts that $(\param_{1:n}, Y)$ is feasible, and uses this to 
``almost'' bound $\tr(\hat{Y})$:
\begin{lemma}
\label{lem:w-tilde-bound-1}
For $\tilde{\param}_i$ as defined in \eqref{eq:param-tilde-def}, and 
$\tilde{Y} \eqdef \frac{2}{\goodfrac n}\hat{\Param}\hat{\Param}^{\top}$, we have 
$\tilde{\param}_i\tilde{\param}_i^{\top} \preceq \tilde{Y}$ for all $i$, and also 
$\tr(\tilde{Y}) \leq \frac{2\radius^2}{\goodfrac}$. In addition, 
\begin{equation}
\label{eq:trace-tilde-bound}
\tr(\hat{Y}) \leq \frac{2\radius^2}{\goodfrac} + \frac{1}{\lambda}\p{\sum_{i=1}^n \weight_i \p{\func_i(\tilde{\param}_i) - \func_i(\hat{\param}_i)}},
\end{equation}
and if $\sum_{i \in \goodset} \weight_i \geq \frac{\goodfrac n}{2}$ then 
\begin{equation}
\label{eq:w-tilde-bound-1}
\sum_{i \in \goodset} \weight_i\p{\func_i(\tilde{\param}_i) - \func_i(\hat{\param}_i)} 
\leq \goodfrac n \p{\sqrt{\tr(\hat{Y})} + \radius}.
\end{equation}
\end{lemma}
\noindent This ``almost'' bounds $\tr(\hat{Y})$ in the following sense: 
if instead of $\sum_{i \in \goodset} \weight_i\p{\func_i(\tilde{\param}_i) - \func_i(\hat{\param}_i)}$, 
\eqref{eq:w-tilde-bound-1} gave a bound on $\sum_{i=1}^n \weight_i\p{\func_i(\tilde{\param}_i) - \func_i(\hat{\param}_i)}$, then 
we could plug in to \eqref{eq:trace-tilde-bound} to obtain (e.g.) 
$\tr(\hat{Y}) \leq \frac{2\radius^2}{\goodfrac} + \frac{n}{\lambda}\Big({\sqrt{\tr(\hat{Y})} + \radius}\Big)$, 
after which solving the quadratic for $\tr(\hat{Y})$ would yield a bound.
The problem is that $\func_i(\tilde{\param}_i)$, for $i \not\in \goodset$, 
could be arbitrarily large, so additional work is needed.

\paragraph{Outlier removal.}
This brings us to our final idea of \emph{outlier removal}. The intuition is the following: 
let $z_i \eqdef {\func_i(\tilde{\param}_i) - \func_i(\hat{\param}_i)}$. Then 
either: (i) the average of $z_i$ over all of $[n]$ is not much larger than over $\goodset$ 
(in which case the bound \eqref{eq:w-tilde-bound-1} extends from $\goodset$ to $[n]$), or 
(ii) the average of $z_i$ is much larger on $[n]$ than on $\goodset$, in which case it 
should be possible to downweight the points in $[n]\backslash \goodset$ a substantial 
amount relative to the points in $\goodset$. This is the role that the outlier removal 
step
(Algorithm~\ref{alg:outlier}) plays, and Lemma~\ref{lem:outlier-1} formalizes its effect 
on the weights $\weight_i$. 
\begin{lemma}
\label{lem:outlier-1}
Suppose that $\frac{1}{n} \sum_{i=1}^n \weight_i \p{\func_i(\tilde{\param}_i) - \func_i(\hat{\param}_i)} \geq \frac{2}{\goodfrac n} \sum_{i \in \goodset} \weight_i \p{\func_i(\tilde{\param}_i) - \func_i(\hat{\param}_i)}$.
Then, the update step in 
Algorithm~\ref{alg:outlier} satisfies
\begin{equation}
\label{eq:outlier-1}
\frac{1}{\goodfrac n} \sum_{i \in \goodset} \weight_i - \weight_i' \leq \frac{1}{2n}\sum_{i=1}^n \weight_i - \weight_i'.
\end{equation}
Moreover, the above supposition holds if 
$\lambda = \frac{\sqrt{8\goodfrac}n\maxop}{\radius}$ and 
$\tr(\hat{Y}) > \frac{6\radius^2}{\goodfrac}$. 
\end{lemma}

Lemma~\ref{lem:outlier-1} says that, if the average value of $z_i$ is at least twice as large over 
$[n]$ as over $\goodset$, then the weights $\weight_i$ decrease at most half as quickly on 
$\goodset$ as on $[n]$. Moreover, this holds whenever $\tr(\hat{Y}) > \frac{6\radius^2}{\goodfrac}$. 

\paragraph{Combining the results.}
Lemma~\ref{lem:outlier-1} ensures that eventually we have 
$\tr(\hat{Y}) \leq \oo\p{\frac{\radius^2}{\goodfrac}}$, which allows us to bound
the overall statistical error (using Lemma~\ref{lem:statistical-error-1}) by 
$\oo\p{\sqrt{\goodfrac}n\radius\maxop}$. In addition, since 
$\lambda = \oo\p{\sqrt{\goodfrac} n \maxop / \radius}$, the optimization error is 
bounded (via Lemma~\ref{lem:optimization-error-1}) by 
$\oo\p{\sqrt{\goodfrac} n \radius \maxop}$, as well. Combining the various bounds, 
we end up obtaining
\begin{equation}
\label{eq:combined-bound}
\Big(\sum_{i \in \goodset} \weight_i\Big)\p{\Efunc\p{\hat{\param}_{\avg}} - \Efunc\p{\param^*}} \leq \oo\p{\sqrt{\goodfrac} n \radius \maxop}.
\end{equation}
Then, since \eqref{eq:outlier-1} ensures that the $\weight_i$ decrease twice as quickly over 
$[n]$ as over $\goodset$, we decrease $\sum_{i \in \goodset} \weight_i$ by at most a factor 
of $2$ over all iterations of the algorithm, so that 
$\sum_{i \in \goodset} \weight_i \geq \frac{\goodfrac n}{2}$. 
Dividing \eqref{eq:combined-bound} through by $\sum_{i \in \goodset} \weight_i$ then yields Theorem~\ref{thm:main}.

In the remainder of this section, we will prove Lemmas~\ref{lem:optimization-error-1} 
through \ref{lem:outlier-1}, then show more formally how Theorem~\ref{thm:main} 
follows from these lemmas.


\subsubsection*{Proof of Lemma~\ref{lem:optimization-error-1}}

We want to bound $\sum_{i \in \goodset} \weight_i \func_i(\hat{\param}_i)$. As noted above, 
the optimality of $(\hat{\param}_{1:n}, \hat{Y})$ implies that 
$\sum_{i=1}^n \weight_i \func_i(\hat{\param}_i) + \lambda \tr(\hat{Y}) \leq \sum_{i=1}^n \weight_i \func_i(\param_i) + \lambda \tr(Y)$
for any feasible point $\param_{1:n}, Y$. We will use the particular choice of 
\begin{equation}
\param_i = \left\{ \begin{array}{ccl} \param^* & : & i \in \goodset \\ \hat{\param}_i & : & i \not\in \goodset \end{array} \right., \quad Y = \hat{Y} + w^*(w^*)^{\top}.
\end{equation}
It is easy to see that this is a feasible point for the optimization: if $i \in \goodset$, 
then $\param_i\param_i^{\top} = w^*(w^*)^{\top} \preceq \hat{Y} + w^*(w^*)^{\top}$, and 
if $i \not\in \goodset$ then $\param_i\param_i^{\top} = \hat{\param}_i\hat{\param}_i^{\top} \preceq \hat{Y} \preceq \hat{Y} + w^*(w^*)^{\top}$. Therefore, we have
\begin{equation}
\sum_{i=1}^n \weight_i \func_i(\hat{\param}_i) + \lambda \tr(\hat{Y}) \leq \sum_{i \in \goodset} \weight_i \func_i(\param^*) + \sum_{i \not\in \goodset} \weight_i \func_i(\hat{\param}_i) + \lambda \p{\tr(\hat{Y}) + \|w^*\|_2^2}.
\end{equation}
Re-arranging yields $\sum_{i \in \goodset} \weight_i\func_i(\hat{\param}_i) \leq \sum_{i \in \goodset} \weight_i\func_i(\param^*) + \lambda \|w^*\|_2^2$, as was to be shown.

\subsubsection*{Proof of Lemma~\ref{lem:statistical-error-1}}

First, by using H\"{o}lder's inequality $\tr(A^{\top}B) \leq \|A\|_{\op}\|B\|_*$, we 
can show the following:
\begin{lemma}
\label{lem:holder-1}
For any $\param_0$ and any $\param_{1:n}$ satisfying $\param_i\param_i^{\top} \preceq Y$ 
for all $i$, we have
\begin{equation}
\label{eq:holder-1}
\Big|\sum_{i \in \goodset} \weight_i \langle \nabla \func_i(\param_0) - \nabla \Efunc(\param_0), \param_i \rangle\Big| \leq \goodfrac n\sqrt{\tr(Y)}\maxop.
\end{equation}
\end{lemma}
\noindent To establish \eqref{eq:statistical-error-1a} from Lemma~\ref{lem:holder-1}, note that
\begin{align}
\sum_{i \in \goodset} \weight_i \p{\func_i(\hat{\param}_{\avg}) - \func_i(\hat{\param}_i)} 
 &\stackrel{(i)}{\leq} \sum_{i \in \goodset} \weight_i \langle \nabla \func_i(\hat{\param}_{\avg}), \hat{\param}_{\avg} - \hat{\param}_i \rangle \\
 &\stackrel{(ii)}{=} \sum_{i \in \goodset} \weight_i \langle \nabla \func_i(\hat{\param}_{\avg}) - \nabla \Efunc(\hat{\param}_{\avg}), \hat{\param}_{\avg} - \hat{\param}_i \rangle \\
 &\stackrel{(iii)}{\leq} \goodfrac n\p{\sqrt{\tr(\hat{Y})} + \radius^2}\maxop,
\end{align}
where (i) is by convexity of $\func_i$, (ii) is because 
\begin{align*}
\sum_{i \in \goodset} \weight_i \langle \nabla \Efunc(\hat{\param}_{\avg}), \hat{\param}_{\avg} - \hat{\param}_i \rangle &= \langle \nabla \Efunc(\hat{\param}_{\avg}), \sum_{i \in \goodset} \weight_i\p{\hat{\param}_{\avg} - \hat{\param}_i} \rangle \\
 &= 0,
\end{align*}
and (iii) is two applications of Lemma~\ref{lem:holder-1}, with
$(\param_i, Y) = (\hat{\param}_{\avg}, \hat{\param}_{\avg}\hat{\param}_{\avg}^{\top})$ and $(\param_i, Y) = (\hat{\param}_i, \hat{Y})$.

To establish \eqref{eq:statistical-error-1b}, 
we expand $\Efunc(\hat{\param}_{\avg}) - \Efunc(\param^*)$ as an integral; letting 
$\param(t) = t\hat{\param}_{\avg} + (1-t)\param^*$, we have
\begin{align}
\sum_{i \in \goodset} \weight_i \p{\Efunc(\hat{\param}_{\avg}) - \Efunc(\param^*)} 
 &= \int_{0}^{1} \sum_{i \in \goodset} \weight_i \langle \nabla \Efunc(\param(t)), \hat{\param}_{\avg} - \param^* \rangle dt \\
 &= \int_{0}^{1} \sum_{i \in \goodset} \weight_i \langle \nabla \func_i(\param(t)), \hat{\param}_{\avg} - \param^* \rangle dt \\
\notag &\phantom{++} + \int_{0}^{1} \sum_{i \in \goodset} \weight_i \langle \nabla \Efunc(\param(t)) - \nabla \func_i(\param(t)), \hat{\param}_{\avg} - \param^* \rangle dt \\
 &\stackrel{(iv)}{\leq} \int_{0}^{1} \sum_{i \in \goodset} \weight_i \langle \nabla \func_i(\param(t)), \hat{\param}_{\avg} - \param^* \rangle dt + \int_{0}^{1} 2\goodfrac n\radius\maxop dt \\
 &= \p{\sum_{i \in \goodset} \weight_i \p{\func_i(\hat{\param}_{\avg}) - \func_i(\param^*)}} + 2\goodfrac n\radius\maxop,
\end{align}
as claimed.
Here (iv) invokes Lemma~\ref{lem:holder-1} at
$\param_0 = \param(t)$, $\param_i = \hat{\param}_{\avg} - \param^*$, 
$Y = (\hat{\param}_{\avg} - \param^*)(\hat{\param}_{\avg} - \param^*)^{\top}$.

\subsubsection*{Proof of Lemma~\ref{lem:w-tilde-bound-1}}

Let $\tilde{\param}_i = \sum_{j=1}^n a_{ij} \hat{\param}_j$ as in \eqref{eq:param-tilde-alg}. We have
\begin{align}
\tilde{\param}_i\tilde{\param}_i^{\top} 
 &= \p{\sum_{j=1}^n a_{ij} \hat{\param}_j}\p{\sum_{j=1}^n a_{ij}\hat{\param}_j}^{\top} \\
 &\stackrel{(i)}{\preceq} \sum_{j=1}^n a_{ij} \hat{\param}_j\hat{\param}_j^{\top} \\
 &\stackrel{(ii)}{\preceq} \sum_{j=1}^n \frac{2}{\goodfrac n} \hat{\param}_j\hat{\param}_j^{\top} \\
 &= \frac{2}{\goodfrac n} \hat{\Param}\hat{\Param}^{\top} = \tilde{Y}.
\end{align}
Here (ii) is because $a_{ij} \leq \frac{2}{\goodfrac n}$, while (i) is an 
instantiation of the inequality $\bE[x]\bE[x]^{\top} \preceq \bE[xx^{\top}]$ (using 
the fact that $\sum_j a_{ij} = 1$).
This completes the first claim.

Next, $\tr(\tilde{Y}) = \frac{2}{\goodfrac n}\tr(\hat{\Param}\hat{\Param}^{\top}) = 
\frac{2}{\goodfrac n}\|\hat{\Param}\|_F^2 \leq \frac{2}{\goodfrac n} \cdot n\radius^2 
= \frac{2\radius^2}{\goodfrac}$. This completes the second claim. The inequality 
\eqref{eq:trace-tilde-bound} then follows from \eqref{eq:optimality-feasibility}, 
by optimality of $(\hat{\param}_{1:n}, \hat{Y})$ and feasibility of $(\tilde{\param}_{1:n}, \tilde{Y})$.

Finally, by taking $a_{ij} = \frac{\weight_j}{\sum_{j' \in \goodset} \weight_{j'}}$, we see that 
$\hat{\param}_{\avg} = \sum_{j \in \goodset} \p{\frac{\weight_j}{\sum_{j' \in \goodset} \weight_{j'}}} \hat{\param}_j$ is a 
feasible point for $\tilde{\param}_i$ (note that $a_{ij} \leq \frac{2}{\goodfrac n}$, since $c_j \leq 1$ 
and $\sum_{j' \in \goodset} \weight_{j'} \geq \frac{\goodfrac n}{2}$ by assumption). We therefore
have $\func_i(\tilde{\param}_i) \leq \func_i(\hat{\param}_{\avg})$ for all $i$, and in particular 
for $i \in \goodset$. Therefore, 
$\sum_{i \in \goodset} \weight_i\p{\func_i(\tilde{\param}_i) - \func_i(\hat{\param}_i)} 
\leq \sum_{i \in \goodset} \weight_i\p{\func_i(\hat{\param}_{\avg}) - \func_i(\hat{\param}_i)}$, 
which is bounded by $\goodfrac n \maxop \p{\sqrt{\tr(\hat{Y})} + \radius}$ by 
Lemma~\ref{lem:statistical-error-1}. This establishes \eqref{eq:w-tilde-bound-1} and completes the lemma.

\subsubsection*{Proof of Lemma~\ref{lem:outlier-1}}
First, note that $\tilde{\param}_i\tilde{\param}_i \preceq \hat{Y}$, since it is 
a convex combination of the $\hat{\param}_j$, and so
replacing $\hat{\param}_i$ with $\tilde{\param}_i$ yields a feasible 
point for the optimization. Therefore, $\func_i(\tilde{\param}_i) \geq \func_i(\hat{\param}_i)$, 
and so $z_i \geq 0$.

Now, remember that 
$\weight_i' = \weight_i \cdot \frac{z_{\max} - z_i}{z_{\max}}$. 
Then for any $S \subseteq [n]$, we have
\begin{align}
\sum_{i \in S} \weight_i - \weight_i' &= \sum_{i \in S} \weight_i \cdot \frac{z_i}{z_{\max}} \\
 &= \frac{1}{z_{\max}} \sum_{i \in S} \weight_iz_i.
\end{align}
Taking $S = \goodset$ and $S = [n]$, it follows that
\begin{align}
\frac{\sum_{i \in \goodset} \weight_i - \weight_i'}{\sum_{i=1}^n \weight_i - \weight_i'} &= \frac{\sum_{i \in \goodset} \weight_iz_i}{\sum_{i=1}^n \weight_iz_i} \\
 &\stackrel{(i)}{\leq} \frac{\goodfrac}{2},
\end{align}
where (i) is by the supposition 
$\frac{1}{n}\sum_{i=1}^n \weight_i z_i \geq \frac{2}{\goodfrac n} \sum_{i \in \goodset} \weight_i z_i$.
This establishes \eqref{eq:outlier-1}.

Next, suppose that $\lambda = \frac{\sqrt{8\goodfrac}n\maxop}{\radius}$ and 
$\tr(\hat{Y}) > \frac{6\radius^2}{\goodfrac}$. 
In this case we must show that the supposition
$\frac{1}{n} \sum_{i=1}^n \weight_i z_i \geq \frac{2}{\goodfrac n} \sum_{i \in \goodset} \weight_i z_i$ 
necessarily holds. Note that
\begin{align}
\sum_{i \in \goodset} \weight_iz_i 
 &= \sum_{i \in \goodset} \weight_i\p{\func_i(\tilde{\param}_i) - \func_i(\hat{\param}_i)} \\
 &\leq \sum_{i \in \goodset} \weight_i\p{\func_i(\hat{\param}_{\avg}) - \func_i(\hat{\param}_i)} \\
\label{eq:x-bound} &\leq \goodfrac n \maxop \p{\sqrt{\tr(\hat{Y})} + \radius}
\end{align}
by Lemma~\ref{lem:statistical-error-1} and \eqref{eq:statistical-error-1a}.
Now let $x = \sum_{i \in \goodset} \weight_i z_i$, and assume for the sake of contradiction 
that $\sum_{i=1}^n \weight_i z_i < \frac{2}{\goodfrac} x$. Then by \eqref{eq:trace-tilde-bound} 
in Lemma~\ref{lem:w-tilde-bound-1} we have
\begin{align}
\tr(\hat{Y}) &\stackrel{\eqref{eq:trace-tilde-bound}}{\leq} \frac{2\radius^2}{\goodfrac} + \frac{1}{\lambda}\p{\sum_{i=1}^n \weight_iz_i} \\
 &< \frac{2\radius^2}{\goodfrac} + \frac{1}{\lambda}\p{\frac{2}{\goodfrac} x} \\
 &\stackrel{\eqref{eq:x-bound}}{\leq} \frac{2\radius^2}{\goodfrac} + \frac{2n \maxop}{\lambda}\p{\sqrt{\tr(\hat{Y})} + \radius}.
\end{align}
Solving the quadratic for $\hat{Y}$, we obtain (see Lemma~\ref{lem:quadratic})
$\tr(\hat{Y}) \leq \frac{4\radius^2}{\goodfrac} + \frac{4\radius n\maxop}{\lambda} + \frac{4n^2\maxop^2}{\lambda^2}$. Using the value $\lambda = \frac{\sqrt{8\goodfrac}n\maxop}{\radius}$, 
we have $\tr(\hat{Y}) \leq \frac{6\radius^2}{\goodfrac}$, which is a ruled out by assumption 
and is thus a contradiction. This establishes the claim that $\sum_{i=1}^n \weight_iz_i \geq \frac{2}{\goodfrac}\sum_{i \in \goodset} \weight_iz_i$ and completes the lemma.

\subsubsection*{Proof of Theorem~\ref{thm:main}}

First, we note 
that $\sum_{i \in \goodset} \weight_i \geq \frac{\goodfrac n}{2}$; this is because 
the invariant $\sum_{i \in \goodset} \weight_i \geq \frac{\goodfrac n}{2} + \frac{\goodfrac}{2} \sum_{i=1}^n \weight_i$ holds throughout the algorithm by Lemma~\ref{lem:outlier-1}.
In particular, Algorithm~\ref{alg:main} does eventually return, since Algorithm~\ref{alg:outlier} 
zeros out at least one additional $\weight_i$ each time, and this can happen at most 
$n - \frac{\goodfrac n}{2}$ times before $\sum_{i \in \goodset} \weight_i$ would drop below 
$\frac{\goodfrac n}{2}$ (which we established is impossible).

Now, let ($\hat{\param}_{1:n}$, $\hat{Y}$) be the return value of Algorithm~\ref{alg:main}.
By Lemma~\ref{lem:optimization-error-1}, we have $\sum_{i \in \goodset} \weight_i \p{\func_i(\hat{\param}_i) - \func_i(\param^*)} \leq \lambda \radius^2$. Also, by 
assumption we have $\tr(\hat{Y}) \leq \frac{6\radius^2}{\goodfrac}$. 
By Lemma~\ref{lem:statistical-error-1} we then have
\begin{align}
\sum_{i \in \goodset} \weight_i \p{\Efunc(\hat{\param}_{\avg}) - \Efunc(\param^*)} 
 &\stackrel{\eqref{eq:statistical-error-1b}}{\leq} \sum_{i \in \goodset} \weight_i \p{\func_i(\hat{\param}_{\avg}) - \func_i(\param^*)} + 2\goodfrac n \radius \maxop \\
 &\stackrel{\eqref{eq:statistical-error-1a}}{\leq} \sum_{i \in \goodset} \weight_i \p{\func_i(\hat{\param}_i) - \func_i(\param^*)} + 3\goodfrac n \radius\maxop + \sqrt{6\goodfrac} n \radius\maxop \\
 &\stackrel{\eqref{eq:optimization-error-1}}{\leq} \lambda \radius^2 + 3\goodfrac n \radius\maxop + \sqrt{6\goodfrac} n \radius \maxop \\
 &= 3\goodfrac n \radius\maxop + (\sqrt{6} + \sqrt{8})\sqrt{\goodfrac} n \radius\maxop \\
 &\leq 9\sqrt{\goodfrac} n \radius\maxop.
\end{align}
Since $\sum_{i \in \goodset} \weight_i \geq \frac{\goodfrac n}{2}$, dividing through by 
$\sum_{i \in \goodset} \weight_i$ yields
$\Efunc(\hat{\param}_{\avg}) - \Efunc(\param^*) \leq 18\frac{\maxop \radius}{\sqrt{\goodfrac}}$, 
as was to be shown.

\section{Concentration of Errors: A Local H\"{o}lder's Inequality}
\label{sec:local-holder}

Most of the bounds in Section~\ref{sec:proof-outline} are bounds on an average 
error: for instance, Lemma~\ref{lem:optimization-error-1} bounds the average difference 
between $\func_i(\hat{\param}_i)$ and $\func_i(\param^*)$, and 
Lemma~\ref{lem:statistical-error-1} bounds the average difference between 
$\func_i(\hat{\param}_{\avg})$ and $\func_i(\hat{\param}_i)$. 
One might hope for 
a stronger bound, showing that the above quantities are close together for 
almost all $i$, rather than only close in expectation. This is relevant, for 
instance, in a clustering setting, where we would like to say that almost all 
points are assigned a parameter $\hat{\param}_i$ that is close to the true cluster center. 
Even beyond this relevance, asking whether we obtain concentration of errors in this 
adversarial setting seems like a conceptually natural question: 

\begin{quote}
\emph{If the good data is sub-Gaussian, then can we obtain sub-Gaussian concentration of 
the errors, or can the adversaries force the error to only be small in expectation? 
What properties of the good data affect concentration of errors in the presence of an adversary?}
\end{quote}

In this section, we will show that we can indeed obtain sub-Gaussian concentration, 
at least for the statistical error. In particular, we will characterize the 
concentration behavior of the errors $\func_i(\hat{\param}_{\avg}) - \func_i(\hat{\param}_i)$ 
using a \emph{local H\"{o}lder's inequality}, which depends upon a refined notion of 
$\maxop$ that we denote by $\maxopi$. Before defining $\maxopi$, we will state the 
local H\"{o}lder's inequality:
\begin{lemma}
\label{lem:local-holder}
Suppose that the weights $b_i \in [0,1]$ satisfy 
$\sum_{i \in \goodset} b_i \geq \opfrac \goodfrac n$, 
and that the parameters $\param_i$ satisfy $\param_i\param_i^{\top} \preceq Y$. 
Then, for any $\param_0 \in \sH$, we have
\begin{equation}
\label{eq:local-holder}
\left|\sum_{i \in \goodset} b_i \langle \param_i, \nabla \func_i(\param_0) - \nabla \Efunc(\param_0) \rangle\right| \leq \p{\sum_{i \in \goodset} b_i}\sqrt{\tr(Y)}\maxopi.
\end{equation}
\end{lemma}
We call this a local H\"{o}lder's inequality because it is a sharpening of 
Lemma~\ref{lem:holder-1}, which established via H\"{o}lder's inequality that
\begin{equation}
\left|\sum_{i \in \goodset} c_i \langle \param_i, \nabla \func_i(\param_0) - \nabla \Efunc(\param_0) \rangle\right| \leq \goodfrac n \sqrt{\tr(Y)}\maxop.
\end{equation}
By taking $b_i = \bI[\langle \param_i, \nabla \func_i(\param_0) - \nabla \Efunc(\param_0) \rangle > \sqrt{\tr(Y)}\maxopi]$, 
Lemma~\ref{lem:local-holder} implies in particular that 
$\langle \param_i, \nabla \func_i(\param_0) - \nabla \Efunc(\param_0) \rangle \leq \sqrt{\tr(Y)}\maxopi$ 
for all but $\opfrac \goodfrac n$ values of $i \in \goodset$.

\paragraph{A local spectral norm bound.}
We now define $\maxopi$. The quantity $\maxopi$ is essentially the maximum value of 
$\maxop$ over any subset of $\goodset$ of size at least $\opfrac |\goodset|$:
\begin{equation}
\label{eq:def-maxopi}
\maxopi \eqdef \max_{\param \in \sH} \max_{T \subseteq \goodset, |T| \geq \lfloor \opfrac \goodfrac n \rfloor} \frac{1}{\sqrt{|T|}} \left\|\left[ \nabla \func_i(\param) - \nabla \Efunc(\param) \right]_{i \in T}\right\|_{\op}.
\end{equation}
(As a special case note that $\maxop_1 = \maxop$.)
The quantity $\maxopi$ bounds not just the operator norm of all of the 
points in $\goodset$, but also the operator norm on any large subset of $\goodset$. We will 
see later that it is often possible to obtain good bounds on $\maxopi$.

\paragraph{Concentration of statistical error.}
Using $\maxopi$, we can obtain an improved version of 
the bounds \eqref{eq:statistical-error-1a} and \eqref{eq:statistical-error-1b} 
from Lemma~\ref{lem:statistical-error-1}, 
showing that $\func_i(\hat{\param}_i)$ is close to 
a nominal value $\func_i(\hat{\param}_{\avg}^b)$ for almost all $i \in \goodset$:

\begin{lemma}
\label{lem:statistical-error-2}
Let the weights $b_i \in [0,1]$ satisfy
$\sum_{i \in \goodset} b_i \geq \opfrac \goodfrac n$, 
and define $\hat{\param}_{\avg}^b \eqdef \frac{\sum_{i \in \goodsete} b_i \hat{\param}_i}{\sum_{i \in \goodset} b_i}$. Then 
the solution $\hat{\param}_{1:n}$, $\hat{Y}$ to \eqref{eq:optimization-2} satisfies
\begin{align}
\label{eq:statistical-error-2a}
\sum_{i \in \goodset} b_i \p{\func_i(\hat{\param}_{\avg}^b) - \func_i(\hat{\param}_i)} &\leq \sum_{i \in \goodset} b_i \langle \nabla \func_i(\hat{\param}_{\avg}^b), \hat{\param}_{\avg}^b - \hat{\param}_i \rangle \leq \p{\sum_{i \in \goodset} b_i} \maxopi \p{\sqrt{\tr(\hat{Y})} + \radius}.
\end{align}
Moreover, for any $\param, \param' \in \sH$, we have
\begin{align}
\label{eq:statistical-error-2b}
\left|\sum_{i \in \goodset} b_i \p{\Efunc(\param) - \Efunc(\param')} - \sum_{i \in \goodset} b_i \p{\func_i(\param) - \func_i(\param')}\right| &\leq 2\p{\sum_{i \in \goodset} b_i}\radius\maxopi.
\end{align}
\end{lemma}
Relative to Lemma~\ref{lem:statistical-error-1}, the main differences are: 
The bounds now hold for any weights $b_i$ (with $\hat{\param}_{\avg}$ 
replaced by $\hat{\param}_{\avg}^b$), and both \eqref{eq:statistical-error-2a} and 
\eqref{eq:statistical-error-2b} have been strengthened in some minor ways relative to 
\eqref{eq:statistical-error-1a} and \eqref{eq:statistical-error-1b} --- 
in \eqref{eq:statistical-error-2a} we are now bounding the linearization 
$\langle \nabla \func_i(\hat{\param}_{\avg}^b), \hat{\param}_{\avg}^b - \hat{\param}_i \rangle$, 
and \eqref{eq:statistical-error-2b} holds at all $\param, \param'$ instead of just 
$\hat{\param}_{\avg}, \param^*$. 
These latter strengthenings are trivial and also hold in Lemma~\ref{lem:statistical-error-1}, 
but were omitted earlier for simplicity. The important difference is that the inequalities 
hold for any $b_i$, rather than just for the original weights $\weight_i$.

It is perhaps unsatisfying that \eqref{eq:statistical-error-2a} holds relative to 
$\hat{\param}_{\avg}^b$, rather than $\hat{\param}_{\avg}$. Fortunately, by 
exploiting the fact that $\hat{\param}_{\avg}$ is nearly optimal for $\Efunc$, 
we can replace $\hat{\param}_{\avg}^b$ with $\hat{\param}_{\avg}$ at the 
cost of a slightly weaker bound:
\begin{corollary}
\label{cor:statistical-error-2}
Let the weights $b_i \in [0,1]$ satisfy $\sum_{i \in \goodset} b_i \geq \opfrac\goodfrac n$, and 
suppose that $\sum_{i \in \goodset} c_i \geq \frac{1}{2}\goodfrac n$. Then 
the solution $\hat{\param}_{1:n}, \hat{Y}$ to \eqref{eq:optimization-2} satisfies
\begin{equation}
\label{eq:cor-statistical-error-2}
\sum_{i \in \goodset} b_i \p{\func_i(\hat{\param}_{\avg}) - \func_i(\hat{\param}_i)} \leq \p{\sum_{i \in \goodset} b_i}\p{\maxopi\p{3\sqrt{\tr(\hat{Y})} + 9\radius} + \frac{2\lambda \radius^2}{\goodfrac n}}.
\end{equation}
In particular, if $\tr(\hat{Y}) = \oo\p{\frac{\radius^2}{\goodfrac}}$ 
and $\lambda = \oo\p{\sqrt{\goodfrac}n\maxopi/\radius}$, 
then $\func_i(\hat{\param}_{\avg}) - \func_i(\hat{\param}_i) \leq \oo\p{{\maxopi\radius}/{\sqrt{\goodfrac}}}$ 
for all but $\opfrac \goodfrac n$ values of $i \in \goodset$.
\end{corollary}
Corollary~\ref{cor:statistical-error-2} shows that no matter 
what the adversary does, the function errors 
$\func_i(\hat{\param}_{\avg}) - \func_i(\hat{\param}_i)$ will be relatively tightly concentrated 
(at least assuming $\maxopi$ is small; we will address the typical size of $\maxopi$ later).
Looking ahead, we will also be able to show that, in the case that the
$\func_i$ are strongly convex, $\|\param_i - \param^*\|_2^2$ is also small for 
almost all $i \in \goodset$. We give this result as Lemma~\ref{lem:statistical-error-3} in 
Section~\ref{sec:strongly-convex}.

\paragraph{Preserving inliers.}
Our outlier removal step can be modified based on $\maxopi$ so that almost none 
of the good points are removed. This is not strictly necessary for any 
of our later results, but is an intuitively appealing property for our 
algorithm to have. 
That we can preserve the good points is unsurprising in light of
Corollary~\ref{cor:statistical-error-2}, which says that the 
good points concentrate, and hence should be cleanly separable from any outliers. 
The modified outlier removal step is given as Algorithm~\ref{alg:outlier-2}.

\begin{algorithm}[b!]
\setlength{\abovedisplayskip}{6pt}
\setlength{\belowdisplayskip}{4pt}
\setlength{\abovedisplayshortskip}{0pt}
\setlength{\belowdisplayshortskip}{0pt}
\caption{Algorithm for updating weights $\weight$ to downweight outliers.}
\label{alg:outlier-2}
\begin{algorithmic}[1]
\Procedure{UpdateWeights}{$\weight$, $\hat{\param}_{1:n}$, $\hat{Y}$}
\State $\tau \gets \maxopi\p{3\sqrt{\tr(\hat{Y})} + 9\radius}$
\For{$i = 1, \ldots, n$}
\State Let $\tilde{\param}_i$ be the solution to \eqref{eq:param-tilde-alg} as in Algorithm~\ref{alg:outlier}.
\State Let $z_i \gets \max\p{\func_i(\tilde{\param}_i) - \func_i(\hat{\param}_i) - \tau, 0}$
\EndFor
\State $z_{\max} \gets \max\{z_i \mid \weight_i \neq 0\}$
\State $\weight_i' \gets \weight_i \cdot \frac{z_{\max} - z_i}{z_{\max}}$ for $i=1,\ldots,n$
\State \Return $\weight'$
\EndProcedure
\end{algorithmic}
\end{algorithm}

Algorithm~\ref{alg:outlier-2} is almost identical to Algorithm~\ref{alg:outlier}.
The only difference from Algorithm~\ref{alg:outlier} is that, instead of setting 
$z_i$ to $\func_i(\tilde{\param}_i) - \func_i(\hat{\param}_i)$, we set 
$z_i$ to $\max\p{\func_i(\tilde{\param}_i) - \func_i(\hat{\param}_i) - \tau, 0}$ 
for an appropriately chosen $\tau$. This creates a buffer such that we do not start 
to downweight points until their function error passes the threshold $\tau$, which helps 
to make sure that very little mass is removed from the good points (because we do not start 
to take away mass until we are fairly sure that a point is bad).
Formally, we have the following result for Algorithm~\ref{alg:outlier-2},
which is analogous to Lemma~\ref{lem:outlier-1} for Algorithm~\ref{alg:outlier}:

\begin{lemma}
\label{lem:outlier-2}
Suppose that $\lambda = \frac{\sqrt{8\goodfrac}n\maxop}{\radius}$ and 
$\tr(\hat{Y}) > \frac{35\radius^2}{\goodfrac}$. Then, the update step in 
Algorithm~\ref{alg:outlier-2} satisfies
\begin{equation}
\label{eq:outlier-2}
\sum_{i \in \goodset} \weight_i - \weight_i' \leq \frac{\opfrac\goodfrac}{2}\sum_{i=1}^n \weight_i - \weight_i'.
\end{equation}
\end{lemma}
This shows that the rate at which mass is removed from $\goodset$ is at most 
$\frac{\opfrac}{2}$ the rate at which mass is removed overall.

\paragraph{Interpreting $\maxopi$.} We end this section by giving some intuition for 
the typical scale of $\maxopi$. Recall that Lemma~\ref{lem:sg-1} shows that, when 
the gradients of $\func_i$ are sub-Gaussian with parameter $\sg$, then 
$\maxop \leq \oo({\sg})$ assuming $n \gg d/\goodfrac$. A similar bound holds for 
$\maxopi$, with an additional factor of $\sqrt{\log(2/\opfrac)}$:
\begin{lemma}
\label{lem:sg-2}
If $\nabla \func_i(\param) - \nabla \Efunc(\param)$ is $\sg$-sub-Gaussian and $L$-Lipschitz, 
then with probability $1-\delta$ we have 
\begin{equation}
\label{eq:sg-2}
\maxopi = \oo\p{\sg\p{\sqrt{\log(2/\opfrac)} + \sqrt{\frac{d\max\p{1, \log\p{\frac{\radius L}{\sg}}} + \log(1/\delta)}{\opfrac \goodfrac n}}}}.
\end{equation}
In particular, if 
$n \geq \frac{1}{\opfrac \goodfrac}\p{d\max\p{1, \log(\tfrac{\radius L}{\sg})} + \log(1/\delta)}$, 
then $\maxopi = \oo\p{\sg\sqrt{\log(2/\opfrac)}}$ with probability $1-\delta$, where $\oo(\cdot)$ 
masks only absolute constants.
\end{lemma}
Lemma~\ref{lem:sg-2} together with Corollary~\ref{cor:statistical-error-2} show that, 
if the gradients of $\func_i$ are sub-Gaussian, then the errors between 
$\func_i(\hat{\param}_{\avg})$ and $\func_i(\hat{\param}_i)$ are also sub-Gaussian, in the sense 
that the fraction of $i$ for which 
$\func_i(\hat{\param}_{\avg}) - \func_i(\hat{\param}_i) \geq \Omega(\sigma\sqrt{\log(2/\opfrac)/\goodfrac})$ 
is at most $\opfrac$. Inverting this, for sufficiently large 
$t$ the fraction of $i$ for which $\func_i(\hat{\param}_{\avg}) - \func_i(\hat{\param}_i) \geq t\sigma/\sqrt{\goodfrac}$ 
is at most $\exp(-\Omega(t^2))$.
In other words, no matter what the adversary does, it cannot prevent the function errors 
from concentrating in a sub-Gaussian manner, provided the good data itself is sub-Gaussian.

\paragraph{A general Chebyshev bound.}
What happens if the function errors are not sub-Gaussian, but we still have a 
bound on $\maxop = \maxop_1$? We can then bound $\maxopi$ in 
terms of $\maxop$ by exploiting the monotonicity of the operator norm.
\begin{lemma}
\label{lem:chebyshev}
For any $\opfrac_1 \leq \opfrac_2$, $\maxop_{\opfrac_1} \leq \sqrt{\frac{\lfloor \opfrac_2 \goodfrac n \rfloor}{\lfloor \opfrac_1 \goodfrac n \rfloor}}\maxop_{\opfrac_2} \leq 2\sqrt{\frac{\opfrac_2}{\opfrac_1}}\maxop_{\opfrac_2}$.
\end{lemma}
When coupled with Corollary~\ref{cor:statistical-error-2}, this shows that the function 
errors concentrate in a Chebyshev-like manner: The fraction of $i$ for which 
$\func_i(\hat{\param}_{\avg}) - \func_i(\hat{\param}_i)$ exceeds 
$\Omega\p{\sigma/\sqrt{\goodfrac\opfrac}}$ 
is at most $\opfrac$, and so the fraction of $i$ for which 
$\func_i(\hat{\param}_{\avg}) - \func_i(\hat{\param}_i) \geq t \sigma/\sqrt{\goodfrac}$ is $\oo\p{\frac{1}{t^2}}$. 
Note that this is {already} a strengthening of the na\"{i}ve bound from Markov's inequality, 
which would only say that the fraction is $\oo\p{\frac{1}{t}}$. The local 
H\"{o}lder's inequality in Lemma~\ref{lem:local-holder} thus leads to a tighter analysis even 
without any further bounding of $\maxopi$.

\subsubsection*{Proof of Lemma~\ref{lem:local-holder}}

Throughout this section we will for convenience use 
$\Delta_i$ to denote $\nabla \func_i(\param_0) - \nabla \Efunc(\param_0)$.
We start with a helper lemma that translates information 
about $\maxopi$ into a more useful form:
\begin{lemma}
\label{lem:local-op-bound}
For any $b_i \in [0,1]$ with $\sum_{i \in \goodset} b_i \geq \lfloor \opfrac \goodfrac n \rfloor$, 
and any matrix $Y \succeq 0$, we have
\begin{equation}
\label{eq:local-op-bound}
\sum_{i \in \goodset} b_i\Delta_i^{\top}Y\Delta_i \leq \p{\sum_{i \in \goodset} b_i}\tr(Y)\maxopi^2.
\end{equation}
\end{lemma}
\begin{proof}
Since the inequality is linear in $b_i$, it is enough to check it 
at the extreme values of $b_i$. These consist of $b_i \in \{0,1\}$ 
such that $\sum_{i \in \goodset} b_i = \lfloor \opfrac \goodfrac n \rfloor$, 
which exactly correspond to subsets $S$ of $\goodset$ of size 
$\lfloor \opfrac \goodfrac n \rfloor$.\footnote{There is also an extreme point corresponding 
to $S = \goodset$, which we ignore as its analysis is identical.}
It thus suffices to check \eqref{eq:local-op-bound} at these points, 
which amounts to showing that
\begin{equation}
\sum_{i \in T} \Delta_i^{\top}Y\Delta_i \leq |T|\tr(Y)\maxopi^2 \text{\ \ for all sets $T$ of size $\lfloor \opfrac \goodfrac n \rfloor$.}
\end{equation}
Letting $D_T = [\Delta_i]_{i \in T}$, we show this as follows:
\begin{align}
\sum_{i \in T} \Delta_i^{\top}Y\Delta_i
 &= \tr\p{Y \sum_{i \in T} \Delta_i\Delta_i^{\top}} \\
 &= \tr\p{Y D_TD_T^{\top}} \\
 &\leq \tr\p{Y}\|D_TD_T^{\top}\|_{\op} \\
 &\leq \tr\p{Y}\|D_T\|_{\op}^2 \\
 &\stackrel{(i)}{\leq} \tr(Y)|T|\maxopi^2,
\end{align}
as was to be shown. Here (i) invokes the definition of 
$\maxopi$, which implies that $\maxopi \geq \max_{|T| = \lfloor \opfrac \goodfrac n \rfloor} \frac{\|D_T\|_{\op}}{\sqrt{|T|}}$.
\end{proof}

We are now ready to prove Lemma~\ref{lem:local-holder}.

\begin{proof}[Proof of Lemma~\ref{lem:local-holder}]
We have
\begin{align}
\left|\sum_{i \in \goodset} b_i \langle \param_i, \nabla \func_i(\param_0) - \nabla \Efunc(\param_0) \rangle\right|
 &= \left|\sum_{i \in \goodset} b_i \langle \param_i, \Delta_i \rangle\right|  \\
 &\stackrel{(i)}{\leq} \sqrt{\p{\sum_{i \in \goodset} b_i} \sum_{i \in \goodset} b_i \langle \param_i, \Delta_i \rangle^2} \\
 &= \sqrt{\p{\sum_{i \in \goodset} b_i} \sum_{i \in \goodset} b_i \Delta_i^{\top} \param_i\param_i^{\top}\Delta_i} \\
 &\stackrel{(ii)}{\leq} \sqrt{\p{\sum_{i \in \goodset} b_i} \sum_{i \in \goodset} b_i \Delta_i^{\top} Y \Delta_i} \\
 &\stackrel{\eqref{eq:local-op-bound}}{\leq} \sqrt{\p{\sum_{i \in \goodset} b_i}^2 \tr(Y)\maxopi^2} \\
 &= \p{\sum_{i \in \goodset} b_i}\maxopi\sqrt{\tr(Y)},
\end{align}
as was to be shown. Here (i) is Cauchy-Schwarz and (ii) uses the condition $\param_i\param_i^{\top} \preceq Y$.
\end{proof}

\subsubsection*{Proof of Lemma~\ref{lem:statistical-error-2}}

First, note that $\p{\func_i(\hat{\param}_{\avg}^b) - \func_i(\hat{\param}_i)} \leq \langle \nabla \func_i(\hat{\param}_{\avg}^b), \hat{\param}_{\avg} - \hat{\param}_i \rangle$ by convexity of 
$\func_i$, so only the second inequality in \eqref{eq:statistical-error-2a} is interesting.
Similarly to how we established \eqref{eq:statistical-error-1a} in Lemma~\ref{lem:statistical-error-1}, we have
\begin{align}
\sum_{i \in \goodset} b_i \langle \nabla \func_i(\hat{\param}_{\avg}^b), \hat{\param}_{\avg}^b - \hat{\param}_i \rangle 
 &= \sum_{i \in \goodset} b_i \langle \nabla \func_i(\hat{\param}_{\avg}^b) - \nabla \Efunc(\hat{\param}_{\avg}^b), \hat{\param}_{\avg}^b - \hat{\param}_i \rangle \\
 &\stackrel{\eqref{eq:local-holder}}{\leq} \p{\sum_{i \in \goodset} b_i}\maxopi\p{\sqrt{\tr(\hat{Y})} + \radius}.
\end{align}
In the final step we invoke \eqref{eq:local-holder} twice, identically to 
Lemma~\ref{lem:statistical-error-1}.

For \eqref{eq:statistical-error-2b}, we follow an identical argument to 
\eqref{eq:statistical-error-1b} from Lemma~\ref{lem:statistical-error-1}. 
Defining $\param(t) = t \param + (1-t)\param'$, we have 
%
\begin{align}
\sum_{i \in \goodset} b_i \p{\Efunc(\param) - \Efunc(\param')} 
 &= \int_{0}^{1} \sum_{i \in \goodset} b_i \langle \nabla \Efunc(\param_t), \param - \param' \rangle dt \\
 &= \int_{0}^{1} \sum_{i \in \goodset} b_i \langle \nabla \func_i(\param_t), \param - \param' \rangle dt + \int_{0}^{1} \sum_{i \in \goodset} b_i \langle \nabla \Efunc(\param_t) - \nabla \func_i(\param_t), \param - \param' \rangle dt \\
 &\stackrel{\eqref{eq:local-holder}}{\leq} \int_{0}^{1} \sum_{i \in \goodset} b_i \langle \nabla \func_i(\param_t), \param - \param' \rangle dt + \int_{0}^{1} 2\p{\sum_{i \in \goodset} b_i}\radius\maxopi dt \\
 &= \p{\sum_{i \in \goodset} b_i \p{\func_i(\param) - \func_i(\param')}} + 2\p{\sum_{i \in \goodset} b_i}\radius\maxopi.
\end{align}
This yields \eqref{eq:statistical-error-2b} 
and completes the lemma.

\subsubsection*{Proof of Corollary~\ref{cor:statistical-error-2}}
We need to bound $\sum_{i \in \goodset} b_i \p{\func_i(\hat{\param}_{\avg}) - \func_i(\hat{\param}_{\avg}^b)}$. 
By \eqref{eq:statistical-error-2b} in Lemma~\ref{lem:statistical-error-2}, 
we have 
\begin{align}
\sum_{i \in \goodset} b_i \p{\func_i(\hat{\param}_{\avg}) - \func_i(\hat{\param}_{\avg}^b)} 
 &\stackrel{\eqref{eq:statistical-error-2b}}{\leq} \p{\sum_{i \in \goodset} b_i}\p{\Efunc(\hat{\param}_{\avg}) - \Efunc(\hat{\param}_{\avg}^b) + 2\maxopi\radius} \\
\label{eq:cor-se2-target} &\leq \p{\sum_{i \in \goodset} b_i}\p{\Efunc(\hat{\param}_{\avg}) - \Efunc(\param^*) + 2\maxopi\radius}.
\end{align}
On the other hand, by Lemmas~\ref{lem:statistical-error-1} and \ref{lem:optimization-error-1}, we also have
\begin{align}
\p{\sum_{i \in \goodset} \weight_i}\p{\Efunc(\hat{\param}_{\avg}) - \Efunc(\param^*)}
 &\stackrel{\eqref{eq:statistical-error-1b}}{\leq} \sum_{i \in \goodset} \weight_i\p{\func_i(\hat{\param}_{\avg}) - \func_i(\param^*)} + 2\goodfrac n\maxop\radius \\
 &\stackrel{\eqref{eq:statistical-error-1a}}{\leq} \sum_{i \in \goodset} \weight_i\p{\func_i(\hat{\param}_i) - \func_i(\param^*)} + \goodfrac n\maxop\p{\sqrt{\tr(\hat{Y})} + 3\radius} \\
 &\stackrel{\eqref{eq:optimization-error-1}}{\leq} \lambda \radius^2 + \goodfrac n\maxop\p{\sqrt{\tr(\hat{Y})} + 3\radius}.
\end{align}
Dividing through by $\sum_{i \in \goodset} \weight_i$ and using the condition $\sum_{i \in \goodset} \weight_i \geq \frac{1}{2}\goodfrac n$, 
we obtain $\Efunc(\hat{\param}_{\avg}) - \Efunc(\param^*) \leq \frac{2\lambda\radius^2}{\goodfrac n} + 2\maxop\p{\sqrt{\tr(\hat{Y})} + 3\radius}$. 
Using $\maxop \leq \maxopi$ and substituting into \eqref{eq:cor-se2-target} yields 
$$\sum_{i \in \goodset} b_i\p{\func_i(\hat{\param}_{\avg}) - \func_i(\hat{\param}_{\avg}^b)} \leq \p{\sum_{i \in \goodset} b_i}\p{\maxopi\Big({2\sqrt{\tr(\hat{Y})} + 8\radius}\Big) + \frac{2\lambda\radius^2}{\goodfrac n}}.$$ 
Combining this with the bound on $\sum_{i \in \goodset} b_i\p{\func_i(\hat{\param}_{\avg}^b) - \func_i(\hat{\param}_i)}$ 
from \eqref{eq:statistical-error-2a} then gives the desired inequality \eqref{eq:cor-statistical-error-2}.
The rest of Corollary~\ref{cor:statistical-error-2} follows by setting 
$b_i = \bI[\func_i(\hat{\param}_{\avg}) - \func_i(\hat{\param}_i) > C \cdot \maxopi\radius/\sqrt{\goodfrac}]$ 
for an appropriate absolute constant $C$.

\subsubsection*{Proof of Lemma~\ref{lem:outlier-2}}
Recall that $z_i = \max\p{\func_i(\tilde{\param}_i) - \func_i(\hat{\param}_i) - \tau, 0}$. 
Our goal is to show that
$\sum_{i \in \goodset} \weight_i z_i \leq \frac{\opfrac\goodfrac}{2} \sum_{i=1}^n \weight_i z_i$. 
If we can show this, then the result follows identically to the first part of 
Lemma~\ref{lem:outlier-1}.

To start, we have
\begin{align}
\sum_{i \in \goodset} \weight_i z_i 
 &= \sum_{i \in \goodset} \p{\weight_i\bI[z_i>0]} \cdot \p{\func_i(\tilde{\param}_i) - \func_i(\hat{\param}_i) - \tau} \\
 &\leq \sum_{i \in \goodset} \p{\weight_i\bI[z_i>0]} \cdot \p{\func_i(\hat{\param}_{\avg}) - \func_i(\hat{\param}_i) - \tau} \\
 &\stackrel{(i)}{\leq} \max\p{\opfrac \goodfrac n, \sum_{i \in \goodset} \weight_i\bI[z_i>0]}\tau - \sum_{i \in \goodset} \weight_i\bI[z_i>0]\tau \\
 &= \max\p{\opfrac \goodfrac n - \sum_{i \in \goodset} \weight_i\bI[z_i>0], 0}\tau \\
 &\leq \opfrac \goodfrac n \tau.
\end{align}
Here (i) is because $\sum_{i \in \goodset} (\weight_i\bI[z_i>0])\p{\func_i(\hat{\param}_{\avg}) - \func_i(\hat{\param}_i)} \leq \p{\sum_i \weight_i\bI[z_i>0]}\tau$ 
by Lemma~\ref{lem:statistical-error-2} 
if $\sum_{i \in \goodset} \weight_i\bI[z_i>0] \geq \opfrac \goodfrac n$, 
and if the sum is less than $\opfrac \goodfrac n$, we can 
always arbitrarily increase the weights beyond $\weight_i\bI[z_i>0]$ until their 
sum is large enough 
(since $\func_i(\hat{\param}_{\avg}) - \func_i(\hat{\param}_i) \geq 0$, this 
will only make the resulting inequality stronger).

Now let $x = \sum_{i \in \goodset} \weight_iz_i$, and assume for the sake of 
contradiction that $\sum_{i=1}^n \weight_iz_i < \frac{2}{\opfrac\goodfrac}x$.
By Lemma~\ref{lem:w-tilde-bound-1} we have
\begin{align}
\tr(\hat{Y}) 
 &\leq \frac{2\radius^2}{\goodfrac} + \frac{1}{\lambda}\p{\sum_{i=1}^n \weight_i \p{\func_i(\tilde{\param}_i) - \func_i(\hat{\param}_i)}} \\
 &\leq \frac{2\radius^2}{\goodfrac} + \frac{1}{\lambda}\p{\sum_{i=1}^n \weight_i \p{\tau + \max\p{\func_i(\tilde{\param}_i) - \func_i(\hat{\param}_i) - \tau, 0}}} \\
 &\leq \frac{2\radius^2}{\goodfrac} + \frac{1}{\lambda}\p{n\tau + \frac{2}{\opfrac\goodfrac}x} \\
 &\leq \frac{2\radius^2}{\goodfrac} + \frac{3n\tau}{\lambda}
   = \frac{2\radius^2}{\goodfrac} + \frac{3\radius}{\sqrt{8\goodfrac}}\p{3\sqrt{\tr(\hat{Y})} + 9\radius} \\
 &\leq \frac{12\radius^2}{\goodfrac} + \sqrt{\frac{11\radius^2\tr(\hat{Y})}{\goodfrac}}.
\end{align}
By Lemma~\ref{lem:quadratic} applied to $\tr(\hat{Y})$, we have
$\tr(\hat{Y}) \leq \frac{35\radius^2}{\goodfrac}$. 
This contradicts the assumption of the lemma, so we indeed have 
$\sum_{i=1}^n \weight_iz_i \geq \frac{\opfrac \goodfrac}{2}\sum_{i \in \goodset} \weight_iz_i$, 
from which the desired result follows.

\subsubsection*{Proof of Lemma~\ref{lem:chebyshev}}
Let $M$ be any matrix, and $T_1 \subseteq T_2$ be subsets of the rows of $M$. 
Then it is easy to see that $\|M_{T_1}\|_{\op} \leq \|M_{T_2}\|_{\op}$. Using this, 
it is then clear from the definition \eqref{eq:def-maxopi} that 
$\maxop_{\opfrac_1} \leq \sqrt{\frac{\lfloor \opfrac_2 \goodfrac n \rfloor}{\lfloor \opfrac_1 \goodfrac n \rfloor}}\maxop_{\opfrac_2}$, 
since any subset $T$ appearing in the definition of $\maxop_{\opfrac_1}$ can be extended to a 
set $T'$ appearing in the definition of $\maxop_{\opfrac_2}$, whose size is at most 
$\frac{\lfloor \opfrac_2 \goodfrac n \rfloor}{\lfloor \opfrac_1 \goodfrac n \rfloor}$ times 
bigger.

\section{Bounds for Strongly Convex Losses}
\label{sec:strongly-convex}

We now turn our attention to the special case that 
the functions $\func_i$ are strongly convex in $\param$, in the sense that
for all $\param, \param' \in \sH$,
\begin{equation}
\func_i(\param') \geq \langle \param' - \param, \nabla \func_i(\param) \rangle + \frac{\kappa}{2}\|\param' - \param\|_2^2.
\end{equation}

In this case, we will obtain stronger bounds by iteratively clustering 
the output $\hat{\param}_{1:n}$ of Algorithm~\ref{alg:main} and re-running the algorithm on each cluster.
The main theorem in this section is a recovery result in the list decoding model, for 
an algorithm (Algorithm~\ref{alg:cluster}) that formalizes this clustering intuition:
\begin{theorem}
\label{thm:cluster-main}
Suppose that $\opfrac \leq \frac{1}{2}$ and let $\sU$, $\hat{\param}_{1:n}$ 
be the output of Algorithm~\ref{alg:cluster}. 
Then $\sU$ has size at most $\lfloor \frac{1}{(1-\opfrac)\goodfrac} \rfloor$, and
$\min_{u \in \sU} \|u - \param^*\|_2 \leq \oo\p{\frac{\maxopi\sqrt{\log(2/\goodfrac)}}{\kappa \sqrt{\goodfrac}}}$. 
Moreover, 
$\|\hat{\param}_i - \param^*\|_2 \leq \oo\p{\frac{\maxopi\sqrt{\log(2/\goodfrac)}}{\kappa \sqrt{\goodfrac}}}$ 
for all but $\opfrac \goodfrac n$ values of $i \in \goodset$.
\end{theorem}

\newcommand{\cand}[1]{\bar{#1}}
\begin{algorithm}[b!]
\caption{Iterative clustering algorithm for approximating $\param^*$}
\label{alg:cluster}
\begin{algorithmic}[1]
\State $B(u; s)$ denotes the ball of radius $s$ centered at $u$
\State $A(u; s)$ denotes the output of Algorithm~\ref{alg:main} with 
       hypothesis space $\sH \cap B(u; s)$, radius $s$, origin shifted to $u$
\Procedure{FindClusters}{}
\State $\hat{\param}_{1:n}^{(1)} \gets A(0; \radius)$ \Comment{initialize $\hat{\param}$}
\State $\radius^{(1)} \gets \radius$
\For{$t = 1, 2, \ldots$}
  \State $\sW \gets \{\hat{\param}_i^{(t)} \mid \hat{\param}_i^{(t)} \text{ is assigned}\}$
  \If{$\radius^{(t)} < C_1 \cdot \frac{\maxopi\log(2/\goodfrac)}{\kappa \sqrt{\goodfrac}}$} \Comment{if radius is small, compute a clustering and exit}
    \State Let $\rf = C_2 \cdot \frac{\maxopi\sqrt{\log(2/\goodfrac)}}{\kappa \sqrt{\goodfrac}}$ \Comment{$C_1$, $C_2$ are absolute constants}
    \State Greedily find a maximal set of points $u_1, \ldots, u_m$ such that:
    \State \indent (i) $|B(u_j; 2\rf) \cap \sW| \geq (1-\opfrac)\goodfrac n$ for all $j$
    \State \indent (ii) $\|u_j - u_{j'}\|_2 > 4\rf$ for all $j \neq j'$.
    \State \Return $\sU = \{u_1, \ldots, u_m\}$ as well as $\hat{\param}_{1:n}^{(t)}$.
  \EndIf
  \State
  \For{$h = 1, \ldots, 112\log\p{\frac{t(t+1)}{\delta}}$} \Comment{compute candidate assignment $\cand{\param}_{1:n}(h)$}
    \State $\cand{\param}_{1:n}(h) \gets \text{unassigned}$
    \State Let $\sP_h$ be a $(\rho, 2r^{(t)}, \frac{7}{8})$-padded decomposition of $\sW$ with $\rho = \oo\p{r^{(t)}\log(\frac{2}{\goodfrac})}$.
    \For{each $T \in \sP_h$} \Comment{run Algorithm~\ref{alg:main} on each piece of the decomposition}
      \State Let $B(u,\rho)$ be a ball containing $T$
      \State For $i$ with $\hat{\param}_i^{(t)} \in T$, assign $\cand{\param}(h)_{i}$ based on the output of $A(u; \rho + \radius^{(t)})$
    \EndFor
  \EndFor
  \For{$i = 1, \ldots, n$} \Comment{pick an assignment that most of the $\cand{\param}_i(h)$ agree on}
    \State Find a $h_0$ such that 
           $\|\cand{\param}_i(h_0) - \cand{\param}_i(h)\|_2 \leq \frac{1}{3}r^{(t)}$ for at least 
           $\frac{1}{2}$ of the $h$'s.
    \State Set $\hat{\param}_i^{(t+1)} \gets \cand{\param}_i(h_0)$ (leave unassigned if $h_0$ does not exist)
  \EndFor
  \State $\radius^{(t+1)} \gets \frac{1}{2}\radius^{(t)}$
\EndFor
\EndProcedure
\end{algorithmic}
\end{algorithm}

Note, interestingly, that the bound does not depend on the radius $\radius$. 
Since the list decoding model can be reduced to the semi-verified model, 
Theorem~\ref{thm:cluster-main} also yields strengthened results in the semi-verified 
model when the functions are strongly convex (we omit these for brevity).

\paragraph{Algorithm and proof overview.}
Algorithm~\ref{alg:cluster} works at a high level as follows:
first, run Algorithm~\ref{alg:main} to obtain $\hat{\param}_i$ that are (as we will show in 
Proposition~\ref{prop:cluster-concentration}) 
relatively close to $\param^*$ for most $i \in \goodset$. Supposing that the good $\hat{\param}_i$ 
are within distance $\radius' \ll \radius$ of $\param^*$, we can cluster $\hat{\param}_{1:n}$ into balls of radius 
$\tilde{\oo}\p{\radius'}$, and re-run Algorithm~\ref{alg:main} on each cluster; Theorem~\ref{thm:main} 
will now yield bounds in terms of $\radius'$ instead of $\radius$. By repeating 
this enough times, we can shrink our hypothesis space to a small ball around 
$\param^*$, thus obtaining substantially better bounds. A key piece of machinery which 
will allow us to obtain a satisfactory clustering is the notion of a
\emph{padded decomposition}, originally due to \citet{fakcharoenphol2003tight}, 
which we explain in more detail later in this section.

Pseudocode for Algorithm~\ref{alg:cluster} is provided above: We keep track of an upper bound 
$\radius^{(t)}$ on the distance from the $\hat{\param}_i$ to $\param^*$, which is initially 
$\radius$ and decreases by a factor of $2$ each time. If this radius drops below a threshold, 
then we perform a final greedy clustering and exit. Otherwise we use padded decompositions to 
cluster the points, and run Algorithm~\ref{alg:cluster} on each cluster to obtain new 
assignments for each $\hat{\param}_i$ (since the padded decomposition is randomized, we repeat 
this several times to ensure correctness with high probability). We can show (Lemma~\ref{lem:cluster-invariant}) 
that these new assignments $\hat{\param}_i$ will be within distance $\frac{1}{2}\radius^{(t)}$ 
to $\param^*$ for almost all $i \in \goodset$, which is the 
key to proving correctness of the algorithm.

The rest of this section consists of three parts: First, we will show that 
if the $\func_i$ are strongly convex, and $\hat{\param}_{1:n}$ is the output of 
Algorithm~\ref{alg:main}, then $\|\hat{\param}_i - \param^*\|_2$ is small for 
most $i \in \goodset$ (this requires some work, since applying 
Theorem~\ref{thm:main} directly would only imply that 
$\|\hat{\param}_{\avg} - \param^*\|_2$ is small).
Next, we will introduce the notion of a padded decomposition, and show (following 
ideas in \citet{fakcharoenphol2003tight}) that padded decompositions of small diameter 
exist in our setting. Finally, we will combine these two results to analyze 
Algorithm~\ref{alg:cluster} and establish Theorem~\ref{thm:cluster-main}.

\paragraph{Establishing concentration of $\|\hat{\param}_i - \param^*\|_2$.}
We will first show that $\hat{\param}_i$ is close to $\param^*$ for almost all 
$i \in \goodset$. This is captured in Proposition~\ref{prop:cluster-concentration}:
\begin{proposition}
\label{prop:cluster-concentration}
For some absolute constant $C$ and for any $\omega \geq 1$, 
the output $\hat{\param}_{1:n}$ of Algorithm~\ref{alg:main} satisfies
$\|\hat{\param}_i - \param^*\|_2^2 \leq C\omega \cdot \frac{\radius \maxopi}{\kappa\sqrt{\goodfrac}}$ 
for all but $\frac{\opfrac \goodfrac n}{\omega^2}$ values of $i \in \goodset$.
\end{proposition}
The key to establishing Proposition~\ref{prop:cluster-concentration} 
lies in leveraging the bound on the statistical error 
from Lemma~\ref{lem:statistical-error-2}, together with the strong convexity of 
$\func_i$. Recall that Lemma~\ref{lem:statistical-error-2} says that
for any $b_i \in [0,1]$ satisfying $\sum_{i \in \goodset} b_i \geq \opfrac \goodfrac n$, we have
\begin{equation}
\sum_{i \in \goodset} b_i \langle \nabla \func_i(\hat{\param}_{\avg}^b), \hat{\param}_{\avg}^b - \hat{\param}_i \rangle \leq \p{\sum_{i \in \goodset} b_i} \maxopi \p{\sqrt{\tr(\hat{Y})} + \radius}.
\end{equation}
By strong convexity of $\func_i$, we then have
\begin{align}
0 &\leq \sum_{i \in \goodset} b_i \p{\func_i(\hat{\param}_{\avg}^b) - \func_i(\hat{\param}_i)} \\
 &\leq \sum_{i \in \goodset} b_i \p{\langle \nabla \func_i(\hat{\param}_{\avg}^b), \hat{\param}_{\avg}^b - \hat{\param}_i \rangle - \frac{\kappa}{2}\|\hat{\param}_i - \hat{\param}_{\avg}^b\|_2^2} \\
 &\leq \p{\sum_{i \in \goodset} b_i}\maxopi \p{\sqrt{\tr(\hat{Y})} + \radius} - \frac{\kappa}{2}\sum_{i \in \goodset} b_i \|\hat{\param}_i - \hat{\param}_{\avg}^b\|_2^2.
\end{align}
Therefore:
\begin{lemma}
\label{lem:statistical-error-3}
For any $b_i \in [0,1]$ satisfying $\sum_{i \in \goodset} b_i \geq \opfrac \goodfrac n$, 
we have
\begin{equation}
\frac{\sum_{i \in \goodset} b_i \|\hat{\param}_i - \hat{\param}_{\avg}^b\|_2^2}{\sum_{i \in \goodset} b_i} \leq \frac{2}{\kappa}\p{\sqrt{\tr(\hat{Y})} + \radius}\maxopi.
\end{equation}
\end{lemma}
By applying Lemma~\ref{lem:statistical-error-3} to $b_i' = \frac{1}{2}\p{b_i + \frac{\sum_j b_j}{\sum_j c_j}c_i}$, 
we obtain the following corollary, which gives bounds in terms of $\hat{\param}_{\avg}$ rather than $\hat{\param}_{\avg}^b$:
\begin{corollary}
\label{cor:statistical-error-3}
For any $b_i \in [0,1]$ satisfying $\opfrac \goodfrac n \leq \sum_{i \in \goodset} b_i \leq \sum_{i \in \goodset} c_i$, we have
\begin{equation}
\frac{\sum_{i \in \goodset} b_i \|\hat{\param}_i - \hat{\param}_{\avg}\|_2^2}{\sum_{i \in \goodset} b_i} \leq \frac{16}{\kappa}\p{\sqrt{\tr(\hat{Y})} + \radius}\maxopi.
\end{equation}
In particular, at most $\opfrac \goodfrac n$ points $i \in \goodset$ have 
$\|\hat{\param}_i - \hat{\param}_{\avg}\|_2^2 \geq \frac{16}{\kappa}\p{\sqrt{\tr(\hat{Y})} + \radius}\maxopi$.
\end{corollary}
Corollary~\ref{cor:statistical-error-3} is crucial because it shows that 
all but an $\opfrac$ fraction of the $\hat{\param}_i$, for $i \in \goodset$, concentrate 
around $\hat{\param}_{\avg}$.

Note that we also have
$\|\hat{\param}_{\avg} - \param^*\|_2^2 \leq \frac{2}{\kappa}\p{\Efunc(\hat{\param}_{\avg}) - \Efunc(\param^*)}$, 
which is bounded by Theorem~\ref{thm:main}; moreover, Theorem~\ref{thm:main} 
also bounds $\tr(\hat{Y})$. Finally, we have 
$\maxop_{\opfrac/\omega^2} \leq 2\omega \maxopi$ by Lemma~\ref{lem:chebyshev}. 
Combining all of these inequalities, we can obtain 
Proposition~\ref{prop:cluster-concentration}.

\paragraph{Padded decompositions.}
Proposition~\ref{prop:cluster-concentration} says that 
the output $\hat{\param}_1, \ldots, \hat{\param}_n$ of 
Algorithm~\ref{alg:main} satisfies $\|\hat{\param}_i - \param^*\|_2 \leq s$ 
for almost all $i \in \goodset$, for some 
$s \ll \radius$. We would then ideally like to partition the 
$\hat{\param}_i$ into sets of small diameter (say $2s$), such that all of 
$\goodset$ is in a single piece of the partition (so that we can then 
run Algorithm~\ref{alg:cluster} on each piece of the partition, and be 
guaranteed that at least one piece has most of $\goodset$).

In general, 
this may not be possible, but we can obtain a probabilistic version 
of the hoped for result: We will end up finding a partition into sets of 
diameter $\oo\p{s\log(2/\goodfrac)}$ such that, with probability $\frac{7}{8}$, all of 
$\goodset$ is in a single piece of the partition. This leads us to the definition 
of a padded decomposition:

\begin{definition}
Let $x_1, \ldots, x_n$ be points in a metric space.
A \emph{($\rho, \tau, \delta$)-padded decomposition} is 
a (random) partition $\sP$ of $\{x_1,\ldots,x_n\}$ such that:
(i) each element of $\sP$ has diameter at most $\rho$, and (ii) 
for each $x_i$, with probability $1-\delta$ all points within 
distance $\tau$ of $x_i$ lie in a single element of $\sP$.
\end{definition}

\citet{fakcharoenphol2003tight} show that for any $\tau$ and 
$\delta$, a $(\rho, \tau, \delta)$-padded decomposition exists 
with $\rho = \oo\p{\frac{\tau\log(n)}{\delta}}$. Moreover, 
the same proof shows that, if every $x_i$ is within distance 
$\tau$ of at least $\Omega(\goodfrac n)$ other $x_i$, then 
we can actually take $\rho = \oo\p{\frac{\tau\log(2/\goodfrac)}{\delta}}$. 
In particular, in our case we can obtain a 
$(\oo\p{s\log(2/\goodfrac)}, 2s, 7/8)$-padded decomposition of the 
$\hat{\param}_i$ output by Algorithm~\ref{alg:main};
we show this formally in Lemma~\ref{lem:padded}.
This probabilistic notion of clustering turns out to be sufficient 
for our purposes.

\newcommand{\numiter}{l}
\paragraph{Analyzing Algorithm~\ref{alg:cluster}.} 
We are now prepared to analyze Algorithm~\ref{alg:cluster}.
In each iteration, Algorithm~\ref{alg:cluster} independently samples $\numiter = 112\log\p{\frac{t(t+1)}{\delta}}$ 
padded decompositions of the $\hat{\param}_i$. For each decomposition $\sP_h$, it then runs Algorithm~\ref{alg:main} 
on each component of the resulting partition, and thereby obtains candidate values 
$\cand{\param}_1(h), \ldots, \cand{\param}_n(h)$. Finally, it updates $\hat{\param}_i$ by finding a point 
close to at least $\frac{1}{2}$ of the candidate values $\cand{\param}_i(h)$, across $h = 1,\ldots,\numiter$.

The idea for why this works is the following: since $\frac{7}{8}$ of the time, the 
padded decomposition $\sP_h$ succeeds in preserving $\goodset$, it is also the 
case that roughly $\frac{7}{8}$ of the candidate assignments $\cand{\param}_i(h)$ are ``good'' 
assignments close to $\param^*$. Therefore, any point that is close to at least 
$\frac{1}{2}$ of the $\cand{\param}_i(h)$ is, by pigeonhole, close to one of the good 
$\cand{\param}_i(h)$, and therefore also close to $\param^*$.

If we formalize this argument, then we obtain
Lemma~\ref{lem:cluster-invariant}, which controls the behavior of the 
update $\hat{\param}_{1:n}^{(t)} \to \hat{\param}_{1:n}^{(t+1)}$ on each 
iteration of the loop:

\begin{lemma}
\label{lem:cluster-invariant}
Algorithm~\ref{alg:cluster} satisfies the following property:
at the beginning of iteration $t$ of the outer loop, let 
$\goodsete^{(t)}$ denote the set of points $i \in \goodset$ for which 
$\|\hat{\param}_i^{(t)} - \param^*\|_2 \leq \radius^{(t)}$. 
Also suppose that $|\goodsete^{(t)}| \geq (1-\opfrac)\goodfrac n$ and 
$\opfrac \leq \frac{1}{2}$. Then, with probability 
$1-\frac{\delta}{t(t+1)}$ over the randomness in the padded decompositions, 
$\|\hat{\param}_i^{(t+1)} - \param^*\|_2 \leq \frac{1}{2}\radius^{(t)}$ 
for all but $C_0 \cdot \p{\frac{\maxopi\log(2/\goodfrac)}{\kappa \radius^{(t)} \sqrt{\goodfrac}}}^2 \cdot \opfrac\goodfrac n$ points in $\goodsete^{(t)}$, 
for some absolute constant $C_0$.
\end{lemma}
Essentially, Lemma~\ref{lem:cluster-invariant} shows that if almost all of the good 
points are within $r^{(t)}$ of $\param^*$ at the beginning of a loop iteration, then almost all 
of the good points are within $\frac{1}{2}r^{(t)}$ of $\param^*$ at the end of 
that loop iteration, provided $\radius^{(t)}$ is large enough.
Using Lemma~\ref{lem:cluster-invariant}, we can establish Theorem~\ref{thm:cluster-main}.



\subsubsection*{Proof of Theorem~\ref{thm:cluster-main}}
Using Lemma~\ref{lem:cluster-invariant} we can prove Theorem~\ref{thm:cluster-main}.
The key is to show that the precondition $|\goodsete^{(t)}| \geq (1-\opfrac)\goodfrac n$ 
of Lemma~\ref{lem:cluster-invariant} holds throughout Algorithm~\ref{alg:cluster}, for appropriate 
choices of the absolute constants $C_1$ and $C_2$. 

We will actually prove the stronger claim that $|\goodsete^{(t)}| \geq (1-\frac{2}{3}\opfrac)\goodfrac n$ for all $t$. 
Lemma~\ref{lem:cluster-invariant} shows that $\goodsete^{(t)}$ decreases in size by at most 
$C_0\p{\frac{\maxopi\log(2/\goodfrac)}{\kappa \radius^{(t)} \sqrt{\goodfrac}}}^2 \opfrac \goodfrac n$ at iteration $t$. 
Now take $C_1 = \sqrt{2C_0}$.
In the final iteration before halting, since $\radius^{(t)} \geq \sqrt{2C_0} \frac{\maxopi\log(2/\goodfrac)}{\kappa \sqrt{\goodfrac}}$, 
we can conclude by Lemma~\ref{lem:cluster-invariant} that at most $\frac{1}{2}\opfrac n$ points 
leave $\goodsete^{(t)}$ in this iteration of the algorithm. But if we look back to previous 
iterations of the algorithm, the bound on the number of points leaving from $\goodsete^{(t)}$ decreases by a 
factor of $4$ each time $t$ decreases (because $\radius^{(t)}$ increases by a factor of $2$). 
Therefore, the total number of points that leave $\goodsete^{(t)}$ over all iterations of the algorithm is at most 
$\frac{1}{2}\opfrac n \p{1 + \frac{1}{4} + \frac{1}{16} + \cdots} = \frac{2}{3}\opfrac n < \opfrac n$. 
This establishes that $|\goodsete^{(t)}| \geq (1-\frac{2}{3}\opfrac)\goodfrac n$ for all iterations 
of the algorithm.

Now, consider the points $\hat{\param}_{1:n}^{(T)}$ obtained in the final iteration $T$ of Algorithm~\ref{alg:cluster}.
By invoking Proposition~\ref{prop:cluster-concentration} with $\omega = \sqrt{3}$, we have for an appropriate 
choice of $C_2$ that 
$\|\hat{\param}_{i}^{(T)} - \param^*\|_2 \leq \oo\p{\sqrt{\frac{r^{(T-1)} \maxopi}{\kappa \sqrt{\goodfrac}}}} = C_2 \cdot \frac{\maxopi\sqrt{\log(2/\goodfrac)}}{\kappa\sqrt{\goodfrac}} = \rf$ 
for all but $\frac{1}{3}\opfrac \goodfrac n$ values of $i \in \goodsete^{(T-1)}$, and hence for all 
but $\opfrac\goodfrac n$ values of $i \in \goodset$. Call the set of points satisfying this 
condition $\goodsete^{\mathrm{final}}$.

To finish, observe that 
by construction, $\sU$ is a maximal set of centers $u$ such that:
(i) $|B(u; 2\rf) \cap \sW| \geq (1-\opfrac)\goodfrac n$ and 
(ii) $B(u; 2\rf) \cap B(u'; 2\rf) = \emptyset$ for $u \neq u'$.
But for $i \in \goodsete^{\mathrm{final}}$, we have 
$B(\hat{\param}_i^{(T)}; 2\rf) \supseteq B(\param^*; \rf) \supseteq \{\hat{\param}_i^{(T)} \mid i \in \goodsete^{\mathrm{final}}\}$, and hence 
$|B(\hat{\param}_i^{(T)}; 2\rf) \cap \sW| \geq (1-\opfrac)\goodfrac n$. Therefore, 
$B(\hat{\param}_i^{(T)}; 2\rf)$ must intersect 
$B(u; 2\rf)$ for some $u \in \sU$, as otherwise we could add 
$\hat{\param}_i^{(T)}$ to $\sU$. This implies that 
$\|\hat{\param}_i^{(T)} - u\|_2 \leq 4\rf$, so
that $\|\param^* - u\|_2 \leq 5\rf$, and hence $\|\hat{\param}_{i'}^{(T)} - u\|_2 \leq 6\rf$ 
for all $i' \in \goodsete^{\mathrm{final}}$. Therefore, the center $u$ 
satisfies the required property that $\|u-\param^*\|_2 = \oo(\rf)$ and 
$\|\hat{\param}_i^{(T)} - u\|_2 = \oo(\rf)$ for all but 
$\opfrac \goodfrac n$ elements of $\goodset$.

We finally note that, for $u \in \sU$, the sets $B(u; 2\rf)$ are 
disjoint and each contain at least $(1-\opfrac)\goodfrac n$ points, so 
there can be at most $\lfloor \frac{1}{(1-\opfrac)\goodfrac} \rfloor$ elements in $\sU$.

\subsubsection*{Proof of Proposition~\ref{prop:cluster-concentration}}

By Corollary~\ref{cor:statistical-error-3}, we have that 
$\|\hat{\param}_i - \hat{\param}_{\avg}\|_2^2 \leq \oo\p{\frac{\sqrt{\tr(\hat{Y}) + \radius^2}\maxop_{\opfrac/\omega^2}}{\kappa}}$ for all but $\frac{\opfrac \goodfrac}{\omega^2} n$ points $i \in \goodset$. But $\tr(\hat{Y}) \leq \oo\p{\frac{\radius^2}{\goodfrac}}$, and 
$\maxop_{\opfrac/\omega^2} \leq 2\omega \maxopi$, so we have
$\|\hat{\param}_i - \hat{\param}_{\avg}\|_2^2 \leq \oo\p{\omega \cdot \frac{\radius\maxopi}{\kappa \sqrt{\goodfrac}}}$ for all but $\frac{\opfrac \goodfrac}{\omega^2} n$ points.

In addition, by Theorem~\ref{thm:main-init} we have 
$\Efunc(\hat{\param}_{\avg}) - \Efunc(\param^*) \leq \oo\p{\frac{\radius\maxop}{\sqrt{\goodfrac}}}$,
which by the strong convexity of $\Efunc$ implies that
$\|\hat{\param}_{\avg} - \param^*\|_2^2 \leq \oo\p{\frac{\radius\maxop}{\kappa\sqrt{\goodfrac}}}$. 
Combining these, we have
$\|\hat{\param}_i - \param^*\|_2^2 \leq 2\p{\|\hat{\param}_i - \hat{\param}_{\avg}\|_2^2 + \|\hat{\param}_{\avg} - \param^*\|_2^2} \leq \oo\p{\omega \cdot \frac{\radius\maxopi}{\kappa\sqrt{\goodfrac}} + \frac{\radius\maxop}{\kappa\sqrt{\goodfrac}}} = \oo\p{\omega \cdot \frac{\radius\maxopi}{\kappa\sqrt{\goodfrac}}}$, as was to be shown.

\subsubsection*{Proof of Corollary~\ref{cor:statistical-error-3}}

For notational convenience, 
let $B = \sum_{i \in \goodset} b_i$ and $C = \sum_{i \in \goodset} c_i$. We have 
\begin{align}
\sum_{i \in \goodset} b_i\|\hat{\param}_i - \hat{\param}_{\avg}^2\|_2^2
 &\stackrel{(i)}{\leq} \sum_{i, j \in \goodset} b_i \cdot \frac{c_j}{C} \|\hat{\param}_i - \hat{\param}_j\|_2^2 \\
 &= \frac{1}{B} \sum_{i, j \in \goodset} b_i \cdot (\tfrac{B}{C}c_j) \|\hat{\param}_i - \hat{\param}_j\|_2^2 \\
 &\stackrel{(ii)}{\leq} \frac{1}{B} \sum_{i,j \in \goodset} (b_i + \tfrac{B}{C}c_i)(b_j + \tfrac{B}{C}c_j)\|\hat{\param}_i - \hat{\param}_j\|_2^2,
\end{align}
where the first inequality is convexity of $\|\cdot\|_2^2$, and 
the second is because we only added positive terms to the sum.

Now, we will apply Lemma~\ref{lem:statistical-error-3} at the value 
$b_i' = \frac{1}{2}\p{b_i + \frac{B}{C}c_i}$. 
This satisfies $b_i' \in [0,1]$ (since $\sum_j c_j \geq \sum_j b_j$), and also 
$\sum_i b_i' = \sum_i b_i \geq \opfrac\goodfrac n$. We can therefore invoke Lemma~\ref{lem:statistical-error-3} 
to obtain
\begin{align}
\frac{1}{B} \sum_{i,j \in \goodset} (b_i + \tfrac{B}{C}c_i)(b_j + \tfrac{B}{C}c_j)
 &= \frac{4}{B}\sum_{i,j \in \goodset} b_i'b_j' \|\hat{\param}_i - \hat{\param}_j\|_2^2  \\
 &= 8\sum_{i \in \goodset} b_i' \|\hat{\param}_i - \hat{\param}_{\avg}^{b'}\|_2^2  \\
 &\leq \frac{16}{\kappa}\p{\sqrt{\tr(\hat{Y})} + \radius}\maxopi,
\end{align}
as was to be shown.
 
\subsubsection*{Proof of Lemma~\ref{lem:cluster-invariant}}

Call a partition $\sP_h$ \emph{good} if the set $\goodsete^{(t)}$ lies within 
a single piece of $\sP_h$, and let $H$ be the set of iterations for which 
$\sP_h$ is good, and $\numiter = 112\log\p{\frac{t(t+1)}{\delta}}$ be the total number 
of iterations. Note that each $h$ is good with probability $\frac{7}{8}$ independently, 
so by the Chernoff bound, the probability that $|H| < \frac{3}{4}\numiter$ is 
at most $\exp\p{-\frac{\numiter}{112}} = \frac{\delta}{t(t+1)}$. 

Note that, for any good $h$, if $T$ is the element of $\sP_h$ containing 
$\goodsete^{(t)}$, then $\param^*$ is within distance $\radius^{(t)}$ of 
$T$, and hence if $T \subseteq B(u, \rho)$ then $\param^* \in B(u, \rho + \radius^{(t)})$. 
Therefore, the hypothesis space for $A(u, \rho + \radius^{(t)})$ contains $\param^*$ and 
so Proposition~\ref{prop:cluster-concentration} implies that, for any good $h$,
\begin{equation}
\label{eq:cluster-invariant-1}
\|\cand{\param}_i(h) - \param^*\|_2 \leq \oo\p{\sqrt{\omega \cdot \frac{(\rho + \radius^{(t)})\maxopi}{\kappa\sqrt{\goodfrac}}}} = \oo\p{\sqrt{\omega \cdot \frac{\radius^{(t)}\maxopi\log(\frac{2}{\goodfrac})}{\kappa\sqrt{\goodfrac}}}}
\end{equation}
for all but $\frac{\opfrac \goodfrac n}{\omega^2}$ values 
of $i \in \goodsete^{(t)}$. 
We will take $\omega$ such that the right-hand-side of \eqref{eq:cluster-invariant-1} is $\frac{1}{6}\radius^{(t)}$, which 
yields $\omega = \Omega\p{\frac{\kappa \radius^{(t)} \sqrt{\goodfrac}}{\maxopi\log(\frac{2}{\goodfrac})}}$.

Next, call an $i \in \goodsete^{(t)}$ \emph{bad} if $\|\cand{\param}_i(h) - \param^*\|_2 > \frac{1}{6}\radius^{(t)}$ 
for more than $\frac{\numiter}{8}$ of the good $h$. 
Averaging over the good $h$, 
there are therefore at most $\frac{8\opfrac \goodfrac n}{\omega^2}$ 
bad values of $i$ (since each such bad value would constitute 
$\frac{\numiter}{8}$ violations of \eqref{eq:cluster-invariant-1}, 
and we know that there are at most $\numiter \cdot \frac{\opfrac \goodfrac n}{\omega^2}$ 
violations in total).

Now if $i$ is not bad, then $\|\cand{\param}_i(h) - \param^*\|_2 \leq \frac{1}{6}\radius^{(t)}$ 
for at least $\frac{5\numiter}{8}$ values of $h$ (because it is small for all but $\frac{\numiter}{8}$ of the at least $\frac{3\numiter}{4}$ good $h$). 
Now, consider any $\cand{\param}_i(h_0)$ that is within distance $\frac{1}{3}\radius^{(t)}$ of at least $\frac{\numiter}{2}$ of the $\cand{\param}_i(h)$.
By the pigeonhole principle, $\cand{\param}_i(h_0)$ is therefore close to at least one of these good $\cand{\param}_i(h)$, and so satisfies 
$\|\cand{\param}_i(h_0) - \param^*\|_2 \leq \|\cand{\param}_i(h_0) - \cand{\param}_i(h)\|_2 + \|\cand{\param}_i(h) - \param^*\|_2 \leq \frac{1}{2}\radius^{(t)}$.
Moreover, such a $h_0$ exists since any of the good $\cand{\param}_i(h)$ 
be within distance $\frac{1}{3}\radius^{(t)}$ of the other good $\cand{\param}_i(h')$.

In sum, the assigned value $\hat{\param}_i^{(t+1)} = \cand{\param}_i(h_0)$ has distance 
at most $\frac{1}{2}\radius^{(t)}$ to $\param^*$ for all but $\frac{8\opfrac\goodfrac n}{\omega^2}$ values 
of $i \in \goodsete^{(t)}$, which yields the desired result.

\section{Lower Bounds}
\label{sec:lower-bounds}

In this section we prove lower bounds showing that the dependence of our 
bounds on $\maxop$ 
is necessary even if $p^*$ is a multivariate Gaussian. For $\goodfrac$, we are only 
able to show a necessary dependence of $\sqrt{\log(\frac{1}{\goodfrac})}$, 
rather than the $\sqrt{{1}/{\goodfrac}}$ appearing in our bounds. 
Determining the true worst-case dependence 
on $\goodfrac$ is an interesting open problem.

One natural question is whether $\maxop$, which typically depends on the maximum 
singular value of the covariance matrix, is really the right dependence, or whether 
we could achieve bounds based on e.g. the average singular value instead.
Lemma~\ref{lem:lower-bound-cover} rules this out, showing that it would require 
$\Omega(2^k)$ candidates in the list-decodable setting to achieve dependence 
on even the $k$th singular value of the covariance matrix.

\begin{lemma}
\label{lem:lower-bound-cover}
Suppose that $p^*$ is known to be a multivariate Gaussian 
$\sN(\mu, \Sigma)$ with covariance $\Sigma \preceq \Sigma_0$, where 
$\mu$ and $\Sigma$ are otherwise unknown. 
Also let $\sigma_k^2$ denote the $k$th largest singular value of $\Sigma_0$.
Then, given any amount of data, an $\goodfrac$-fraction of which 
is drawn from $p^*$, any procedure for outputting at most $m = 2^{k-1}$ candidate 
means $\hat{\mu}_1, \ldots, \hat{\mu}_m$ must have, with 
probability at least $\frac{1}{2}$, $\min_{j=1}^m \|\hat{\mu}_j - \mu\|_2 \geq \frac{\sigma_k}{4}\cdot \frac{\log(\frac{1}{\goodfrac})}{\sqrt{1 + \log(\frac{1}{\goodfrac})}}$.
\end{lemma}

By the reduction from the semi-verified to the list-decodable model, we 
obtain a lower bound in the semi-verified setting as well:

\begin{corollary}
\label{cor:lower-bound-semiverified}
Any algorithm that has $\|\hat{\mu} - \mu\|_2 < \frac{\sigma_k}{4}\cdot \frac{\log(\frac{1}{\goodfrac})}{\sqrt{1 + \log(\frac{1}{\goodfrac})}}$ 
with probability at least $\frac{1}{2}$ in the semi-verified model requires at least
$\frac{k-2}{\log_2(\frac{1}{\goodfrac})}$ verified samples.
\end{corollary}

We remark that the same proofs show lower bounds for non-strongly convex losses as well. 
For instance, while mean estimation corresponds to $\func_i(\param) = \|\param - x_i\|_2^2$, 
we could also estimate the direction of a vector by taking $\func_i(\param) = -\param^{\top}x_i$, 
and constraining $\|\param\|_2 \leq \radius$; in this case the minimizer of $\Efunc(\param)$ 
is $\param^* = \radius \cdot \frac{\mu}{\|\mu\|_2}$, where $\mu = \bE[x]$. An essentially 
identical argument to Lemma~\ref{lem:lower-bound-cover} and Corollary~\ref{cor:lower-bound-semiverified} 
shows that at least $2^{\Omega(k)}$ candidates $\hat{\param}_j$ 
are necessary to obtain 
$\min_{j} \Efunc(\hat{\param}_j) - \Efunc(\param^*) \leq \radius \cdot \frac{\sigma_k}{4} \cdot \frac{\log(\frac{1}{\goodfrac})}{\sqrt{1 + \log(\frac{1}{\goodfrac})}}$ 
with probability greater than $\frac{1}{2}$, and at least $\Omega\p{\frac{k}{\log_2(\frac{1}{\goodfrac})}}$ 
verified samples are needed in the semi-verified model.
To avoid repetition we omit these.

\subsubsection*{Proof of Lemma~\ref{lem:lower-bound-cover}}
Let us suppose that the unverified data has distribution $\hat{p} = \sN(0, \Sigma_0)$, 
which is possible iff $\hat{p} \geq \goodfrac p^*$. 
We start with a lemma characterizing when it is possible that the true distribution 
$p^*$ has mean $\mu$:
\begin{lemma}
\label{lem:gaussian-cover}
Let $\hat{p} = \sN(0, \Sigma_0)$ and $p^* = \sN(\mu, \Sigma_0 - \lambda \mu\mu^{\top})$. 
Then $\hat{p} \geq \goodfrac p^*$ provided that 
$\mu^{\top}\Sigma_0^{-1}\mu \leq \frac{\log^2(\frac{1}{\goodfrac})}{1 + \log(\frac{1}{\goodfrac})}$ and $\lambda = \frac{1}{\log(\frac{1}{\goodfrac})}$.
\end{lemma}
As a consequence, if $\mu^{\top}\Sigma_0^{-1}\mu \leq t^2$ for $t = \frac{\log(\frac{1}{\goodfrac})}{\sqrt{1 + \log(\frac{1}{\goodfrac})}}$, 
then $p^*$ could have mean $\mu$. Now, consider the space spanned by the $k$ largest eigenvectors of $\Sigma_0$, and 
let $B_k$ be the ball of radius $t \sigma_k$ in this space. Also let $\sP$ be a maximal packing of $B_k$ of radius $\frac{t}{4}\sigma_k$.
A simple volume argument shows that $|B_k| \geq 2^k$. On the other hand, every element $\mu$ of $B_k$ is a potential mean because 
it satisfies $\mu^{\top}\Sigma_0^{-1}\mu \leq \frac{\|\mu\|_2^2}{\sigma_k^2} \leq t^2$. Therefore, if the true mean $\mu$
is drawn uniformly from $B_k$, any $2^{k-1}$ candidate means $\hat{\mu}_j$ must miss at least half of the elements of $B_k$ (in the sense of 
being distance at least $\frac{t}{4}\sigma_k$ away) and so with probability at least $\frac{1}{2}$, $\|\hat{\mu}_j - \mu\|_2$ is 
at least $\frac{t}{4}\sigma_k$ for all $j$, as was to be shown.

\subsubsection*{Proof of Corollary~\ref{cor:lower-bound-semiverified}}
Suppose that we can obtain $\hat{\mu}$ satisfying $\|\hat{\mu} - \mu\|_2 < \frac{\sigma_k}{4} \cdot \frac{\log(\frac{1}{\goodfrac})}{\sqrt{1+\log(\frac{1}{\goodfrac})}}$
 with $m \leq \frac{k-2}{\log_2(\frac{1}{\goodfrac})}$ samples, with probability at least $\frac{1}{2}$.
If we sample $m$ elements uniformly from the unverified samples, with probability $\goodfrac^m$ we will obtain 
only samples from $p^*$, and therefore with probability $\frac{1}{2}\goodfrac^m$ we can obtain (by assumption)
a candidate mean $\hat{\mu}$ with $\|\hat{\mu} - \mu\|_2 < \frac{\sigma_k}{4} \cdot \frac{\log(\frac{1}{\goodfrac})}{\sqrt{1 + \log(\frac{1}{\goodfrac})}}$.
If we repeat this $\frac{2}{\goodfrac^m}$ times, then with probability $1 - \p{1 - \frac{1}{2}\goodfrac^m}^{\frac{2}{\goodfrac^m}} \geq 1-\frac{1}{e} > \frac{1}{2}$,
at least one of the candidate means will be close to $\mu$. But by Lemma~\ref{lem:lower-bound-cover}, 
this implies that $\frac{2}{\goodfrac^m} \geq 2^{k-1}$, and so $m \geq \frac{k-2}{\log_2(\frac{1}{\goodfrac})}$, as claimed.

\section{Intuition: Stability Under Subsets}
\label{sec:intuition}

In this section, we establish a sort of ``duality for robustness'' that provides intuition 
underlying our results. Essentially, we will show the following:
\begin{quotation}
\noindent \emph{If a statistic of a dataset is approximately preserved across every large 
subset of the data, then it can be robustly recovered even in the presence 
of a large amount of additional arbitrary/adversarial data.}
\end{quotation}
To be a bit more formal, suppose that for a set of points 
$\{x_1, \ldots, x_n\}$ lying in $\bR^d$, there is a subset 
$I \subseteq [n]$ with the following property: \emph{for any subset 
$T \subseteq I$ of size at least $\frac{1}{2}\goodfrac^2n$, the mean over 
$T$ is $\varepsilon$-close to the mean over $I$}. In symbols,
\begin{equation}
\label{eq:I-stable}
\left\|\mu_T - \mu_I\right\|_2 \leq \varepsilon \text{ for all } T \subseteq I \text{ with } |T| \geq \frac{1}{2}\goodfrac^2n, \text{ where } \mu_T \eqdef \frac{1}{|T|} \sum_{i \in T} x_i.
\end{equation}
In such a case, can we approximate the mean of $I$, even if $I$ is not known and the points 
in $[n] \backslash I$ are arbitrary?

If we do not care about computation, the answer is yes, via a simple exponential-time 
algorithm. Call a set of points $J$ $\goodfrac$-stable if it satisfies the following properties:
$|J| \geq \goodfrac n$, and \eqref{eq:I-stable} holds with $I$ replaced by $J$ (i.e., the 
mean moves by at most $\varepsilon$ when taking subsets of $J$ of size at least 
$\frac{1}{2}\goodfrac^2 n$). 
Then, we can run the following (exponential-time) algorithm:
\begin{enumerate}
\item Initialize $\sU = []$.
\item Find an $\goodfrac$-stable set $J$ which has overlap 
      at most $\frac{1}{2}\goodfrac^2n$ with all elements of $\sU$.
\item Append $J$ to $\sU$ and continue until no more such sets exist.
\end{enumerate}
A simple inclusion-exclusion argument\footnote{
Specifically, $k$
sets of size $\goodfrac n$ with pairwise overlap $\frac{1}{2}\goodfrac^2n$ must have a 
union of size at least $k\goodfrac n - \frac{1}{2}\binom{k}{2}\goodfrac^2 n$, 
which implies $k \leq \frac{2}{\goodfrac}$. 
}  
shows that $k \leq \frac{2}{\goodfrac}$. 
Therefore, the above 
algorithm terminates with $|\sU| \leq \frac{2}{\goodfrac}$. On the other hand, 
by construction, $I$ must have overlap at least $\frac{1}{2}\goodfrac^2 n$ with 
at least one element of $\sU$ (as otherwise we could have also added $I$ to $\sU$). 
Therefore, for some $J \in \sU$, $|I \cap J| \geq \frac{1}{2}\goodfrac^2 n$. 
But then, letting $T = I \cap J$, we have
$\|\mu_{I} - \mu_J\|_2 \leq \|\mu_{I} - \mu_T\|_2 + \|\mu_T - \mu_J\|_2 \leq 2\varepsilon$.
Therefore, the mean over $I$ is within distance $2\varepsilon$ of at least one of the 
$\mu_J$, for $J \in \sU$.

We have therefore established our stated duality property for robustness: 
\emph{if a statistic is stable under taking subsets of data, then it can also be 
recovered if the data is itself a subset of some larger set of data}.

Can we make the above algorithm computationally efficient? First, can we even 
\emph{check} if a set $J$ is $\goodfrac$-stable? This involves checking the following 
constraint:
\begin{equation}
\left\|\sum_{i \in J} c_i(x_i - \mu_J)\right\|_2 \leq \varepsilon \|c\|_1 
\text{for all $c$ such that }  c_i \in \{0,1\}, \sum_{i \in J} c_i \geq \frac{1}{2}\goodfrac^2n.
\end{equation}
It is already unclear how to check this constraint efficiently. However, defining the matrix 
$C_{ij} = c_ic_j \in [0,1]^{J \times J}$, we can take a semi-definite relaxation of 
$C$, resulting in the 
constraints $C_{ij} \in [0,1]$ and $\tr(C) \geq \frac{1}{2}\goodfrac^2n$. 
Letting $A_{ij} = (x_i - \mu_J)^{\top}(x_j - \mu_J)$, 
this results in the following sufficient condition for $\goodfrac$-stability:
\begin{align}
\notag \tr(A^{\top}C) &\leq \varepsilon^2 \tr(C)^2 \text{ for all $C$ such that } \\
C \succeq 0,  \
&C_{ij} \in [0,1],  \
\tr(C) \geq \frac{1}{2}\goodfrac^2n.
\label{eq:constraint-A}
\end{align}
This is still non-convex due to the presence of $\tr(C)^2$.
However, a sufficient condition for \eqref{eq:constraint-A} to hold is that
$\tr(A^{\top}C) \leq \frac{1}{2}\varepsilon^2\goodfrac^2 n \tr(C)$ for all $C \succeq 0$, 
which is equivalent to $\|A\|_{\op} \leq \frac{1}{2}\varepsilon^2\goodfrac^2 n$, or, 
letting $X$ be the matrix with rows $x_i - \mu_J$, 
$\varepsilon \geq \sqrt{\frac{2}{\goodfrac}} \cdot \p{\frac{1}{\sqrt{\goodfrac n}}\|X\|_{\op}}$, 
where $\frac{1}{\sqrt{\goodfrac n}}\|X\|_{\op}$ is essentially our definition of 
$\maxop$. From this, we can directly see why $\maxop / \sqrt{\goodfrac}$ appears in our bounds: 
it is a convex relaxation of the $\goodfrac$-stability condition.
Our main optimization \eqref{eq:optimization-1} can (somewhat less directly) also be thought of 
as a semidefinite relaxation which attempts to represent the means of all $\frac{2}{\goodfrac}$ 
of the $\goodfrac$-stable sets simultaneously (this is done implicitly, via the major axes 
of the ellipse $Y$).

As an aside, we note that the above discussion, together with the bound 
(Lemma~\ref{lem:subgaussian}) on the operator norm of samples from a sub-Gaussian distribution, 
implies that a set of samples from a sub-Gaussian distribution is $\goodfrac$-stable 
with high probability, and in particular the mean of any large subset of samples is close 
to the mean of the overall set.
Moreover, even if a distribution has only bounded second moments, 
Proposition~\ref{prop:bss} shows that a suitable subset of samples from the 
distribution will be $\goodfrac$-stable.

\section{Applications}
\label{sec:applications}

In this section we present several applications of 
our main results.
\subsection{Gaussians and Mixtures of sub-Gaussians}
For robustly learning the mean of a distribution with bounded variance, 
we have the following result:
\begin{corollary}
\label{cor:subgaussian}
\label{cor:mean}
Let $p^*$ be a distribution on $\bR^d$ 
with mean $\mu$ and bounded covariance: $\Cov_{p^*}[x] \preceq \sigma^2 I$. 
Suppose that we observe 
$n \geq \frac{d}{\goodfrac}$ points, of which $\goodfrac n$ are drawn from 
$p^*$. Then, for any $\opfrac \leq \frac{1}{2}$, with probability $1-\exp(-\Omega(\opfrac^2\goodfrac n))$, it is possible to output parameters 
$\hat{\param}_1, \ldots, \hat{\param}_m$, where $m \leq \lfloor \frac{1}{\goodfrac - \opfrac} \rfloor$, 
such that $\min_{j=1}^m \|\hat{\param}_j - \mu\|_2 \leq \oo\p{\frac{\sg}{\opfrac}\sqrt{\frac{\log(2/\goodfrac)}{\goodfrac}}}$.
\end{corollary}
\begin{proof}
Define $\func_i(\param) = \|\param - x_i\|_2^2$, 
which is $1$-strongly convex and has $\param^* = \mu$. In this case, 
$\nabla \func_i(\param) - \nabla \Efunc(\param) = x_i - \mu$, and so Proposition~\ref{prop:bss} 
implies that with probability $1-\exp(-\Omega(\opfrac^2\goodfrac n))$ there is a subset of $(1-\opfrac/2)\goodfrac n$ of the good points 
satisfying $\maxop = \oo\p{\sigma/\sqrt{\opfrac}}$, and hence (by Lemma~\ref{lem:chebyshev}) 
$\maxop_{\opfrac/2} = \oo\p{\sigma/\opfrac}$. 
Theorem~\ref{thm:cluster-main} 
then says that we can output parameters 
$\hat{\param}_1, \ldots, \hat{\param}_m$, where 
$m \leq \lfloor \frac{1}{(1-\opfrac/2)^2\goodfrac} \rfloor \leq \lfloor \frac{1}{(1-\opfrac)\goodfrac} \rfloor$, such that 
$\min_{j=1}^m \|\hat{\param}_j - \mu\|_2 \leq \oo\p{\frac{\sg}{\opfrac} \sqrt{\frac{\log(2/\goodfrac)}{\goodfrac}}}$, 
as was to be shown.
\end{proof}

For robustly learning a mixture of distributions, we have the following result:
\begin{corollary}
\label{cor:mixture}
Suppose we are given $n$ samples, where each sample either comes from one of 
$k$ sub-Gaussian distributions $p_1^*, \ldots, p_k^*$ 
(with $\Cov_{p_i^*}[x] \preceq \sigma^2 I$), or is arbitrary. 
Let $I_i$, $\goodfrac_i$, and $\mu_i$ denote respectively the set of 
points drawn from $p_i^*$, the fraction of points drawn from $p_i^*$, and 
the mean of $p_i^*$, and let $\goodfrac = \min_{i=1}^k \goodfrac_i$.
Suppose that $n \geq \frac{d}{\goodfrac}$. Then for any 
$\opfrac \leq \frac{1}{2}$, with probability 
$1 - k\exp(-\Omega(\opfrac^2\goodfrac n))$, 
we can output a partition 
$T_1, \ldots, T_m$ of $[n]$ and candidate means 
$\hat{\mu}_1, \ldots, \hat{\mu}_m$ such that:
\begin{enumerate}
\item $m \leq \lfloor \frac{1}{(1-\opfrac)\goodfrac} \rfloor$.
\item For all but $\opfrac \goodfrac_i n$ of the elements of $h$ of $I_i$, 
      $h$ is assigned a partition $T_j$ such that $\|\mu_i - \hat{\mu}_j\|_2 \leq \oo\Big({\frac{\sg}{\opfrac}\sqrt{\frac{\log(2/\goodfrac)}{\goodfrac}}}\Big)$,
      where $\oo(\cdot)$ masks only absolute constants.
\end{enumerate}
\end{corollary}
\begin{proof}
We apply Theorem~\ref{thm:cluster-main} in the same way as in Corollary~\ref{cor:mean}, 
letting the candidate means $\hat{\mu}_1, \ldots, \hat{\mu}_m$ be the cluster centers $\sU$ 
output by Algorithm~\ref{alg:cluster}.
For each $i$, we then know that 
$\|\hat{\mu}_j - \mu_i\|_2 \leq \oo\Big(\frac{\sigma}{\opfrac}\sqrt{\frac{\log(2/\goodfrac)}{\goodfrac}}\Big)$ 
for at least one of the means $\hat{\mu}_j$.\footnote{
One might worry that, because different hyper-parameters are needed 
for different values of $\goodfrac$, a single instance of Algorithm~\ref{alg:cluster} 
cannot find all of the $\mu_i$ for different $i$. However, Algorithm~\ref{alg:cluster} 
will run correctly given any $\goodfrac \leq |\goodset|/n$, so we can 
in fact recover all the $\mu_i$ simultaneously.
}
Moreover, we know that the $\hat{\param}_h$ output by Algorithm~\ref{alg:cluster} 
satisfy $\|\hat{\param}_h - \mu_i\|_2 \leq \oo\Big(\frac{\sigma}{\opfrac}\sqrt{\frac{\log(2/\goodfrac)}{\goodfrac}}\Big)$ 
for all but $\opfrac \goodfrac_i n$ elements of $h \in I_i$. 
We can therefore define $T_j$ to be the set of $h$ for which $\hat{\param}_h$ is closer to $\hat{\mu}_j$ than 
to any of the other $\hat{\mu}_{j'}$. It is then easy to see that any $h \in I_i$ which is close 
to $\param_i$ will be assigned a mean $\hat{\mu}_j$ that is also close to 
$\param_i$. This yields the desired result.
\end{proof}

For the planted partition model, we have the following result:
\begin{corollary}
\label{cor:partition}
Let $I_1, \ldots, I_k$ be disjoint subsets of $[n]$, and consider the 
following random directed graph with vertex set $[n]$:
\begin{itemize}
\item If $u, v \in I_i$, the probability of an edge from $u$ to $v$ is $\frac{a}{n}$.
\item If $u \in I_i, v \not\in I_i$, the probability of an edge from $u$ to $v$ is $\frac{b}{n}$.
\item If $u \not\in \cup_{i=1}^k I_i$, the edges emanating from $u$ can be arbitrary.
\end{itemize}
Let $\goodfrac = \min_{i=1}^k \frac{|I_i|}{n}$.
Then with probability $1 - k\exp(-\Omega(\goodfrac n))$, 
we can output sets $T_1, \ldots, T_m \subseteq [n]$, 
with $m \leq \frac{4}{\goodfrac}$, such that 
for all $i \in [k]$, there is a $j \in [m]$ with
$|I_i \triangle T_j| \leq \oo\p{\frac{a\log(2/\goodfrac)}{\alpha^2(a-b)^2}n}$, 
where $\triangle$ denotes the symmetric difference of two sets.
\end{corollary}
\begin{proof}
It turns out that this problem reduces to mean estimation as well, 
though we will need to be more careful in how we apply our matrix 
concentration bound. 

Let $A$ be the adjacency matrix of the graph. 
We will let $\func_u(\param) = \|\param - A_u\|_2^2 = \sum_{v=1}^n (\param_v - A_{uv})^2$.
Note that for $u \in I_i$, we have $\mu_i = \bE[A_{uv}] = \frac{b}{n} + \frac{a-b}{n}\bI[v \in I_i]$, 
and $\Cov[A_{u}]$ is a diagonal matrix with entries that are all either $\frac{a}{n}$ or $\frac{b}{n}$, 
so that in particular $\|\Cov[A_{u}]\|_{\op} \leq \frac{a}{n}$, and hence $\sigma = \sqrt{\frac{a}{n}}$.

In this case we have $d = n$, so Proposition~\ref{prop:bss} implies that we 
can find a subset of $I_i$ of size at least $|I_i|/2 \geq \frac{\goodfrac}{2}n$ 
such that $\maxop = \oo\p{\sigma\sqrt{1 + \frac{d}{\goodfrac n}}} = \oo\p{\sqrt{\frac{a}{\goodfrac n}}}$.
Invoking Theorem~\ref{thm:cluster-main} with $\opfrac = \frac{1}{2}$, 
we obtain candidate means $\hat{\mu}_1, \ldots, \hat{\mu}_m$, with $m \leq \frac{4}{\goodfrac}$, 
such that $\|\mu_i - \hat{\mu}_j\|_2^2 \leq \oo\p{\frac{\maxop^2\log(2/\goodfrac)}{\goodfrac}} = \oo\p{\frac{a\log(2/\goodfrac)}{\goodfrac^2 n}}$.

Now, we define the set $T_j$ such that $u \in T_j$ iff $(\hat{\mu}_j)_u \geq \frac{a+b}{2n}$. 
Suppose for some $u$, $u \in I_i$ but $u \not\in T_j$. Then we must have 
$(\hat{\mu}_j)_u < \frac{a+b}{2n}$, but $(\mu_i)_u = \frac{a}{n}$. Therefore, 
any $u \in I_i$ with $u \not\in T_j$ contributes at least $\frac{(a-b)^2}{4n^2}$ 
to $\|\mu_i - \hat{\mu}_j\|_2^2$. Similarly, any $u \in T_j$ with $u \not\in I_i$ 
similarly contributed at least $\frac{(a-b)^2}{4n^2}$. We can conclude that
$\|\mu_i - \hat{\mu}_j\|_2^2 \geq \frac{(a-b)^2}{4n^2} |I_i \triangle T_j|$. 
In particular, if $\|\mu_i - \hat{\mu}_j\|_2^2 = \oo\p{\frac{a\log(2/\goodfrac)}{\goodfrac^2 n}}$, 
then $|I_i \triangle T_j| = \oo\p{\frac{a\log(2/\goodfrac)}{\goodfrac^2(a-b)^2}n}$, 
as was to be shown.
\end{proof}

Finally, we have the following corollary for density modeling:
\begin{corollary}
\label{cor:density}
Let $p_{\theta}(x) = \exp\p{\theta^{\top}\phi(x) - A(\theta)}$, 
where $A(\theta)$ is the log-normalization constant. Let $x_1, \ldots, x_n$ 
be points such that $\goodfrac n$ are drawn from $p^*$, and 
$\Cov_{p^*}[\phi(x)] \preceq \sigma^2 I$. Also suppose that 
$\phi(x) \in \bR^d$ and $n \geq \frac{d}{\goodfrac}$.
Then with probability $1 - \exp(-\Omega(\goodfrac n))$, the following hold:
\begin{itemize}
\item Given a single verified sample $x \sim p^*$, we can output 
      $\hat{\theta}$ such that $\bE[\KL{p_{\theta^*}}{p_{\hat{\theta}}} \leq \oo\p{\frac{\sigma\radius}{\sqrt{\goodfrac}}}]$, 
      where $\theta^*$ is the minimizer of $\bE[-\log p_{\theta}(x)]$.
\item We can output $m = \oo\p{\frac{\log(2/\delta)}{\goodfrac}}$ candidates 
      $\hat{\theta}_j$ such that, with probability $1-\delta$, 
      $\min_{j=1}^m \bE[\KL{\smash{p_{\theta^*}}}{\smash{p_{\hat{\theta}_j}}}]$ is at most 
      $\oo\p{\frac{\sigma\radius}{\sqrt{\goodfrac}}}$.
\end{itemize}
\end{corollary}
Note that $\KL{p_{\theta^*}}{p_{\theta}} = (\theta^* - \theta)^{\top}\mu + A(\theta) - A(\theta^*)$, 
where $\mu = \bE[\phi(x)]$, so Corollary~\ref{cor:density} roughly asks for $\theta$ to have 
large inner product with $\mu$, but with the additional goal of keeping $A(\theta)$ small.
\begin{proof}[Proof of Corollary~\ref{cor:density}]
Define the loss $\func_i(\theta) = -\log p_{\theta}(x_i) = A(\theta) - \theta^{\top}\phi(x_i)$. 
We observe that $\Efunc(\theta) - \Efunc(\theta^*) = \bE_{p^*}[(\theta^* - \theta)^{\top}\phi(x) + A(\theta) - A(\theta^*)] 
= (\theta^*-\theta)^{\top}\mu + A(\theta) - A(\theta^*) = \KL{p_{\theta^*}}{p_{\hat{\theta}}}$, 
so bounding $\KL{p_{\theta^*}}{p_{\theta}}$ is equivalent to bounding $\Efunc(\theta) - \Efunc(\theta^*)$.

Note that $\nabla \func_i(\theta) - \nabla \Efunc(\theta) = x_i - \mu$. Therefore,
Proposition~\ref{prop:bss} implies that with probability $1 - \exp(-\Omega(\goodfrac n))$ 
there is a set of at least $\frac{\goodfrac}{2}n$ samples $x_i$ with 
spectral norm bound $S = \oo\p{\sigma}$.
Theorem~\ref{thm:main} then implies that we obtain an ellipse 
$\sE_{\hat{Y}}$ and samples $\hat{\theta}_{1:n}$ such that 
$\hat{\theta}_i\hat{\theta}_i^{\top} \preceq \hat{Y}$ and 
$\Efunc(\hat{\theta}_{\avg}) - \Efunc(\theta^*) \leq \oo\p{\frac{\sigma\radius}{\sqrt{\goodfrac}}}$. 

Now take a single verified sample $x \sim p^*$, and define 
$\hat{\theta} = \argmin_{\theta \in \sE_Y} A(\theta) - \theta^{\top}\phi(x)$.
Similarly to Lemma~\ref{lem:semi-verified}, we have
\begin{align}
\Efunc(\hat{\theta}) - \Efunc(\theta^*)
 &= (\hat{\theta}^*-\theta)^{\top}\mu + A(\hat{\theta}) - A(\theta^*) \\
\label{eq:density-1} &= (\theta^*-\hat{\theta})^{\top}(\mu-\phi(x)) + (\theta^*-\hat{\theta})^{\top}\phi(x) + A(\hat{\theta}) - A(\theta^*) \\
 &\stackrel{(i)}{\leq} (\theta^*-\hat{\theta})^{\top}(\mu-\phi(x)),
\end{align}
where (i) is because $\hat{\theta}$ minimizes the right-hand term in \eqref{eq:density-1}.
On the other hand, we have
\begin{align}
\bE_{x}[((\theta^*-\hat{\theta})^{\top}(\mu-\phi(x)))^2]
 &= \bE_x[(\mu-\phi(x))^{\top}(\theta^*-\hat{\theta})(\theta^*-\hat{\theta})^{\top}(\mu-\phi(x))] \\
 &\leq 4\bE_x[(\mu-\phi(x))^{\top}\hat{Y}(\mu-\phi(x))] \\
 &= 4\tr(\Cov[\phi(x)]^{\top}\hat{Y}) \\
 &\leq 4\sigma^2\tr(\hat{Y}) \\
 &= \oo\p{\frac{\sigma^2\radius^2}{\goodfrac}}.
\end{align}
In particular, $\bE[(\theta^*-\hat{\theta})^{\top}(\mu-\phi(x))] = \oo\p{\frac{\sigma\radius}{\sqrt{\goodfrac}}}$, 
which together with the preceding inequality yields the desired bound in $\Efunc(\hat{\theta}) - \Efunc(\theta^*)$.

The above gives us the desired result in the semi-verified model. But 
then, we can reduce the semi-verified model to the list decodable model 
by sampling random points from the untrusted data, which gives us the 
list decodable result as well.
\end{proof}


\appendix

\section{Padded Decompositions}

The following lemma is essentially from \citet{fakcharoenphol2003tight}.

\newcommand{\km}{k_{\max}}
\begin{lemma}
\label{lem:padded}
Let $x_1, \ldots, x_n$ be points in a metric space, and let $I \subseteq [n]$ be 
such that $d(x_i, x_j) \leq r$ for all $i, j \in I$. Consider the following 
procedure:
\begin{itemize}
\item Initialize $X = \{1,\ldots,n\}$, $\sP = \emptyset$, and sample $k$ uniformly from 
      $2$ to $\km$, where $\km \geq 1 + \frac{1}{\delta}\log(\tfrac{n}{|I|})$.
\item While there are points remaining in $S$:
      \begin{itemize}
      \item Sample $i$ uniformly from $\{1,\ldots,n\}$
      \item Let $T$ be the set of points in $X$ within distance $kr$ of $x_i$.
      \item Update $X \gets X \backslash T$, $\sP = \sP \cup \{T\}$.
      \end{itemize}
\item Output the partition $\sP$.
\end{itemize}
Then, with probability at least $1-\delta$, all elements of 
$I$ are in a single element of $\sP$.
\end{lemma}
\begin{proof}
It suffices to show the following: if $T$ contains any element of $I$, then 
with probability at least $\frac{3}{4}$ it contains all elements of $I$.

To see this, let $N(k)$ denote the set of points within distance $kr$ of at least one 
point in $I$. 
Then, for a fixed $k$, the probability that $T$ intersects $I$ is $\frac{|N(k)|}{n}$; but 
since all points in $I$ are within distance $r$ of each other, 
the probability that $T$ contains all of $I$ is at least $\frac{|N(k-1)|}{n}$. Letting 
$a_k \eqdef |N(k)|$, the 
conditional probability that $T$ contains $I$, given that it intersects $I$, is therefore 
at least $\frac{a_{k-1}}{a_k}$. Marginalizing over $k$, the probability is at least
\begin{align}
\frac{1}{\km-1} \sum_{k=2}^{\km} \frac{a_{k-1}}{a_k} 
 &\stackrel{(i)}{\geq} \p{\prod_{k=2}^{\km} \frac{a_{k-1}}{a_k}}^{\frac{1}{\km-1}} \\
 &\geq \p{\frac{|I|}{n}}^{\frac{\delta}{\log(n/|I|)}} \\ 
 &= e^{-\delta} \geq 1-\delta,
\end{align}
as was to be shown. Here (i) is the good old AM-GM inequality.
\end{proof}

\section{Matrix Concentration Bounds}

\subsection{Concentration from Second Moments}
The following is proved using techniques from
\citet{batson2012twice}. However, the bound itself appears to be 
novel.

\begin{proposition}
\label{prop:bss}
Suppose that $p$ is a distribution on $\bR^d$ with $\bE_p[X] = \mu$ and 
$\Cov_p[X] \preceq \sigma^2 I$ for some $\sigma$.
Then, given $n$ samples from $p$, with probability $1 - \exp\p{-\frac{\bsserr^2 n}{16}}$ there is a subset 
$J \subseteq [n]$ of size at least $(1-\bsserr)n$ such that 
$\lambda_{\max}\p{\frac{1}{|J|} \sum_{i \in J} (x_i-\mu)(x_i-\mu)^{\top}} \leq \frac{4\sigma^2}{\bsserr}\p{1 + \frac{d}{(1-\bsserr)n}}$.
\end{proposition}

\begin{proof}
Without loss of generality we will take $\mu = 0$.
Our proof relies on the following claim:
\begin{lemma}
\label{lem:bss-induction}
For a matrix $M$, suppose that $M \prec cI$ and
$\tr\p{(cI - M)^{-1}} \leq \frac{1}{4\sigma^2}$. 
Then, for $x \sim p$, with probability at least $1-\frac{\bsserr}{2}$ 
we have $M+\bsserr xx^{\top} \prec (c+4\sigma^2)I$ and 
$\tr\p{((c+4\sigma^2)I - (M + \bsserr xx^{\top}))^{-1}} \leq \frac{1}{4\sigma^2}$.
\end{lemma}
We prove this claim below, first showing how Proposition~\ref{prop:bss} follows 
from the claim.
We will go through the stream of $n$ samples, and keep a running set 
$J$ and matrix $M$. Initially $J = \emptyset$ and $M = 0$. For each index $i$ in the 
stream, we will 
add $i$ to $J$ (and update $M$ to $M + \bsserr x_ix_i^{\top}$) if the 
conclusion of Lemma~\ref{lem:bss-induction} holds for 
$c = 4\sigma^2(d+|J|)$. Note that then, by induction, the precondition of 
Lemma~\ref{lem:bss-induction} always holds for $c = 4\sigma^2(d+|J|)$, since it 
holds when $J = \emptyset$ and the condition for adding $i$ to $J$ ensures 
that it continues to hold.

At the end of the stream, we therefore obtain $J$ and $M$ such that
$M \preceq 4\sigma^2(d+|J|)I$,
which implies that $\lambda_{\max}(\frac{1}{|J|} \sum_{i \in J} x_ix_i^{\top}) = \frac{1}{\bsserr |J|}\lambda_{\max}(M) \leq \frac{4\sigma^2}{\bsserr}\p{1 + \frac{d}{|J|}}$. 
It remains to show that $|J| \geq (1-\bsserr)n$ with high probability.

To see this, note that for each element $i$, $|J|$ increases by $1$ with probability at least 
$1-\frac{\bsserr}{2}$. The distribution of $|J|$ is therefore lower-bounded by the sum 
of $n$ independent Bernoulli variables with bias $1-\frac{\bsserr}{2}$, which by the Chernoff 
bound is greater than $(1-\bsserr)n$ with probability at least $1 - \exp\p{-\frac{\bsserr^2n}{16}}$. 
\end{proof}

\begin{proof}[Proof of Lemma~\ref{lem:bss-induction}]
This is essentially Lemma 3.3 of \citet{batson2012twice}.
Instead of $M + \bsserr xx^{\top}$, it will be helpful later to consider 
consider $M + txx^{\top}$ for arbitrary $t \in [0,\bsserr]$.
By the Sherman-Morrison matrix inversion formula, we have
\begin{equation}
((c+4\sigma^2)I - M - txx^{\top})^{-1} = ((c+4\sigma^2)I - M)^{-1} + t\frac{((c+4\sigma^2)I - M)^{-1}xx^{\top}((c+4\sigma^2)I - M)^{-1}}{1 - tx^{\top}((c+4\sigma^2)I-M)^{-1}x}.
\end{equation}
Therefore, letting $\Phi_{c}(M)$ denote $\tr((cI-M)^{-1})$, we have
\begin{align}
\Phi_{c+4\sigma^2}(M + txx^{\top})
 &= \Phi_{c+4\sigma^2}(M) + \frac{x^{\top}((c+4\sigma^2)I - M)^{-2}x}{1/t - x^{\top}((c+4\sigma^2)I-M)^{-1}x}.
\end{align}
We want to have $\Phi_{c+4\sigma^2I}(M+txx^{\top}) \leq \Phi_c(M)$, 
which in light of the above is equivalent to the condition
\begin{equation}
\label{eq:bss-want}
\frac{1}{t} \stackrel{?}{\geq} x^{\top}((c+4\sigma^2)I - M)^{-1}x + \frac{x^{\top}((c+4\sigma^2)I - M)^{-2}x}{\Phi_c(M) - \Phi_{c+4\sigma^2}(M)}.
\end{equation}
Let us compute the expectation of the right-hand-side. Since $\bE[x^{\top}Ax] = \tr(A^{\top}\Cov[x])$ and $\Cov[x] \preceq \sigma^2 I$, 
the right-hand-side of \eqref{eq:bss-want} has expectation at most 
\begin{align}
\sigma^2\p{\tr\p{((c+4\sigma^2)I - M)^{-1}} + \frac{\tr\p{((c+4\sigma^2)I - M)^{-2}}}{\Phi_c(M) - \Phi_{c+4\sigma^2}(M)}} \leq \sigma^2\p{\Phi_{c+4\sigma^2}(M) + \frac{1}{4\sigma^2}},
\end{align}
where the second step is in light of the inequality
\begin{align*}
\Phi_c(M) - \Phi_{c+4\sigma^2}(M) &= \tr\p{(cI-M)^{-1} - ((c+4\sigma^2)I-M)^{-1}} \\
 &\geq 4\sigma^2 \tr\p{((c+4\sigma^2)I-M)^{-2}}.
\end{align*}
Since 
$\Phi_{c+4\sigma^2}(M) \leq \Phi_{c}(M) \leq \frac{1}{4\sigma^2}$, the right-hand-side 
of \eqref{eq:bss-want} is at most 
$\frac{1}{2}$ in expectation. Therefore, with probability at least 
$1 - \frac{\bsserr}{2}$, it is at most $\frac{1}{\bsserr}$, in which 
case \eqref{eq:bss-want} holds for all $t \in [0,\bsserr]$.

This certainly implies that $\Phi_{c+4\sigma^2}(M+\bsserr xx^{\top}) = \tr\p{((c+4\sigma^2)I - (M + \bsserr xx^{\top}))^{-1}} \leq \frac{1}{4\sigma^2}$. Moreover, 
we also have that $\Phi_{c+4\sigma^2}(M+txx^{\top}) \leq \frac{1}{4\sigma^2} < \infty$ for all 
$t \in [0,\bsserr]$. Since the eigenvalues of $M+txx^{\top}$ are a continuous function of $t$, 
we therefore know that the maximum eigenvalue of $M+txx^{\top}$ never crosses 
$c+4\sigma^2$ for $t \in [0,\bsserr]$, and so $M + \bsserr xx^{\top} \prec (c+4\sigma^2)I$ as well 
(since $M \prec cI \prec (c+4\sigma^2)I$ by assumption).
\end{proof}

\subsection{Concentration for sub-Gaussians}
In this section we provide a matrix concentration result for sub-Gaussian variables, 
which underlies Lemma~\ref{lem:sg-2}.
The proofs closely mirrors Theorem 5.39 of \citet{vershynin2010introduction}.
\begin{lemma}
\label{lem:subgaussian}
Let $p$ be a distribution over vectors $X \in \bR^d$ 
such that (i) $\bE[X] = 0$, 
and (ii) $\bE[\exp\p{\langle \lambda, X \rangle }] \leq \exp\p{\frac{\sg^2}{2}\|\lambda\|_2^2}$ 
for some constant $\sg$.
Then, if $M$ is an $m \times d$ matrix with each row drawn i.i.d. from $p$, 
with probability $1-\delta$ 
the following holds simultaneous for all subsets $T$ of the rows of $M$ with 
$|T| \geq s$:
\begin{equation}
\|M_T\|_{\op}^2 \leq 18\sg^2\p{d + s\log(e/\beta) + \log(1/\delta)},
\end{equation}
where $\beta = s/m$.
\end{lemma}
\begin{proof}
We will bound
\begin{equation}
\label{eq:b-k-1}
\max_{|T| \geq s} \frac{\|M_T\|_{\op}^2 - Q}{|T|},
\end{equation}
where $Q = 18\sigma^2\p{d + \log(1/\delta)}$.

Let $\sN$ be a $(1/4)$-net of the sphere $S^{d-1}$, meaning that all points 
in $S^{d-1}$ are within distance at most $\frac{1}{4}$ of a point in $\sN$. 
As shown in \citet{vershynin2010introduction}, $\sN$ can be taken to have size at most 
$9^d$, and moreover one has $\max_{u \in \sN} 2\|M_Tu\|_2^2 \geq \max_{\|u\|_2 \leq 1} \|M_Tu\|_2^2 = \|M_T\|_{\op}^2$. 
Using these results, we can bound \eqref{eq:b-k-1} as
\begin{align}
\label{eq:b-k-2}
\max_{|T| \geq s} \frac{\|M_T\|_{\op}^2 - Q}{|T|}
 &\leq \max_{u \in \sN} \max_{|T| \geq s} \frac{2\|M_Tu\|_2^2 - Q}{|T|} \\
\label{eq:net} &= \frac{2}{s} \max_{u \in \sN} \max_{|T| = s} \p{\Big(\sum_{i \in T} \langle M_i, u \rangle^2\Big) - Q/2}.
\end{align}
We will focus on bounding the inner expression for a fixed $T$, then union bound over 
the $9^d\binom{m}{s}$ possibilities for $|T|$ and $u$. Let $Z_i = {\langle M_i, u \rangle}^2$. 
Note that $\bE[Z_i] = u^{\top}\Cov[X]u \leq \sg^2$. In addition, 
$\bE[\exp\p{\lambda \langle M_i, u \rangle}] \leq \exp\p{\frac{\sg^2}{2}\lambda^2}$ (since $\|u\|_2 = 1$), 
and so (see Lemma~\ref{lem:sg-se}) we have
$\bE[\exp\p{\lambda\p{Z_i - \sg^2}}] \leq \exp\p{16\sg^4\lambda^2}$ 
for $0 \leq \lambda \leq \frac{1}{4\sg^2}$.
By independence, it follows that $\bE\left[\exp\p{\lambda \sum_{i \in T} (Z_i - \sg^2)}\right] 
\leq \exp\p{16s\sg^4\lambda^2}$, again for 
$0 \leq \lambda \leq \frac{1}{4\sg^2}$. By a Chernoff argument, at 
$\lambda = \frac{1}{4\sg^2}$, we obtain 
\begin{align}
\bP\left[ \sum_{i \in T} \p{\langle M_i, u \rangle}^2 \geq s\sg^2 + t\right]
 &\leq \exp\p{s - \frac{t}{4\sg}}.
\end{align}
We need the right-hand-side to be smaller than $\frac{\delta}{9^d\binom{m}{s}}$, 
for which we can take 
\begin{align}
t &= 4\sg^2\p{d\log(9) + s + s\log(me/s) + \log(1/\delta)} \\
 &\leq 4\sg^2\p{2.2d + s + s\log(e/\beta) + \log(1/\delta)}.
\end{align}
Overall, we have that with probability $1-\delta$, 
\begin{align}
\sum_{i \in T} (\langle M_i, u \rangle)^2 
 &\leq \sg^2\p{8.8d + 5s + 4s\log(e/\beta) + 4\log(1/\delta)} \\
 &\leq 9\sg^2\p{d + s\log(e/\beta) + \log(1/\delta)}
\end{align}
for all $T$ of size $s$ and all $u \in \sN$. 
In particular, \eqref{eq:net} can be bounded by 
$18\sg^2\log(e/\beta)$, which completes the proof.
\end{proof}

\section{Deferred Proofs}

\subsection{Bounding Quadratics}
\begin{lemma}
\label{lem:quadratic}
Suppose that $z \leq a + \sqrt{bz + c^2}$ where $a, b, c, \geq 0$. Then, $z \leq 2a + b + c$.
\end{lemma}
\begin{proof}
Re-arranging to remove the square-root, we have $z^2 - (2a+b)z + (a^2-c^2) \leq 0$.
Now suppose that $z > 2a + b + c$. Then, 
$z^2 - (2a+b)z + (a^2 - c^2) = z(z-2a-b) + (a^2 - c^2) > (2a+b+c)c + (a^2 - c^2) 
 = a^2 + (2a+b)c > 0$, which is a contradiction.
\end{proof}

\subsection{Proof of Lemma~\ref{lem:sg-1}}

We omit this as it is a special case of Lemma~\ref{lem:sg-2}, 
proved below.

\subsection{Proof of Lemma~\ref{lem:semi-verified}}

We start by showing that 
$\bE\left[\sup_{\param \in \sE_Y} |\func(\param) - \Efunc(\param)|^q\right]^{1/q} \leq \sqrt{2\tr(Y)}\sigma_q$.
Let $\mu = \bE[x]$ and $\Sigma = \Cov[x]$. We have 
\begin{align}
\bE_{f}\left[ \sup_{\param \in \sE_Y} |\func(\param) - \Efunc(\param)|^q\right] 
 &= \bE_x\left[ \sup_{\param \in \sE_Y} |\phi(\param^{\top}x) - \bE_{x'}[\phi(\param^{\top}x')]|^q\right]  \\
 &= \bE_x\left[ \sup_{\param \in \sE_Y} |\bE_{x'}[\phi(\param^{\top}x) - \phi(\param^{\top}x')]|^q\right]  \\
 &\stackrel{(i)}{\leq} \bE_x\left[ \sup_{\param \in \sE_Y} \bE_{x'}[|\phi(\param^{\top}x) - \phi(\param^{\top}x')|^2]^{q/2}\right]  \\
 &\stackrel{(ii)}{\leq} \bE_x\left[ \sup_{\param \in \sE_Y} \bE_{x'}[(\param^{\top}(x-x'))^2]^{q/2}\right]  \\
 &= \bE_x\left[ \sup_{\param \in \sE_Y} \Big(\param^{\top}\p{\Sigma + (x-\mu)(x-\mu)^{\top}}\param\Big)^{q/2}\right] \\
 &\stackrel{(iii)}{\leq} \bE_x\left[ \tr(Y^{\top}\p{\Sigma + (x-\mu)(x-\mu)^{\top}})^{q/2} \right] \\
 &\stackrel{(iv)}{\leq} 2^{q/2-1}\p{\tr(Y^{\top}\Sigma)^{q/2} + \bE_x[((x-\mu)^{\top}Y(x-\mu))^{q/2}]},
\end{align}
where (i) and (iv) are applications of the power mean inequality, 
(ii) is because $\phi$ is $1$-Lipschitz, and (iii) is because $\param\param^{\top} \preceq Y$.

On the other hand, we have $\Sigma \preceq \sigma_q^2 I$ by the $q$th moment condition (since $q > 2$), 
so $\tr(Y^{\top}\Sigma)^{q/2} \leq \tr(Y)^{q/2}\sigma_q^q$. Also, letting 
$Y = \sum_{i} \lambda_i v_iv_i^{\top}$ be the eigendecomposition of $Y$, we have 
\begin{align}
\bE_x[((x-\mu)^{\top}Y(x-\mu))^{q/2}]
 &= \bE_x\Big[\Big(\sum_{i} \lambda_i (v_i^{\top}(x-\mu))^2\Big)^{q/2}\Big] \\
 &\stackrel{(v)}{\leq} \tr(Y)^{q/2-1} \bE_x\left[ \sum_i \lambda_i (v_i^{\top}(x-\mu))^q \right] \\
 &\stackrel{(vi)}{\leq} \tr(Y)^{q/2}\sigma_q^q,
\end{align}
where (v) is again the power mean inequality and (vi) uses the $q$th moment 
bound $\bE[(v_i^{\top}(x-\mu))^q] \leq \sigma_q^q$.

Combining these, we obtain 
$\bE_f\left[\sup_{\param \in \sE_Y} |\func(\param) - \Efunc(\param)|^q\right] \leq (2\tr(Y))^{q/2}\sigma_q^q$. 
To finish, we simply apply Markov's inequality to this statistic to get a tail bound on 
$\sup_{\param \in \sE_Y} |\func(\param) - \Efunc(\param)|$, after which the result follows 
from a standard uniform convergence argument.

\subsection{Proof of Lemma~\ref{lem:trace}}

By duality of operator and trace norm, we have (letting $z_i$ be the $i$th row of the matrix $Z$)
\begin{align}
\|\Param_T\|_*^2 
 &= \sup_{\|Z\|_{\op} \leq 1} \tr(\Param_T^{\top}Z)^2 \\
 &= \sup_{\|Z\|_{\op} \leq 1} \p{\sum_{i \in T} \param_i^{\top}z_i}^2 \\
 &\stackrel{(i)}{\leq} \sup_{\|Z\|_{\op} \leq 1} |T|\sum_{i \in T} (\param_i^{\top}z_i)^2 \\
 &= \sup_{\|Z\|_{\op} \leq 1} |T|\sum_{i \in T} \param_i^{\top}\p{z_iz_i^{\top}}\param_i \\
 &\stackrel{(ii)}{\leq} \sup_{\|Z\|_{\op} \leq 1} |T|\sum_{i \in T} \tr(Yz_iz_i^{\top}) \\
 &= \sup_{\|Z\|_{\op} \leq 1} |T|\tr(YZ^{\top}Z) \\
 &= |T|\tr(Y),
\end{align}
as was to be shown. Here (i) is Cauchy-Schwarz and (ii) uses the fact that $\param_i\param_i \preceq Y$.

\subsection{Proof of Lemma~\ref{lem:holder-1}}

Let $F$ be the matrix whose $i$th row is $\nabla \func_i(\param_0) - \nabla \Efunc(\param_0)$, 
and let $W$ be the matrix whose $i$th row is $\param_i$. We consider only the rows $i \in \goodset$, 
so each matrix is $\goodfrac n \times d$.
We have 
\begin{align}
\Big|\sum_{i \in \goodset} \weight_i \langle \nabla \func_i(\param_0) - \nabla \Efunc(\param_0), \param_i \rangle\Big| 
 &= \tr(F^{\top}\diag(c)W) \\
\label{eq:holder-1-target} &\leq \|\diag(c)F\|_{\op}\|W\|_*
\end{align}
by H\"{o}lder's inequality. Furthermore, $\|\diag(c)F\|_{\op} \leq \|\diag(c)\|_{\op}\|F\|_{\op} \leq \|F\|_{\op}$, 
and furthermore $\|F\|_{\op} \leq \sqrt{\goodfrac n}\maxop$ by the definition of $\maxop$. 
Finally, $\|W\|_* \leq \sqrt{|\goodset|\tr(Y)} = \sqrt{\goodfrac n \tr(Y)}$ by 
Lemma~\ref{lem:trace}. Combining these, we see 
that \eqref{eq:holder-1-target} is bounded by $\goodfrac n \maxop \sqrt{\tr(Y)}$, 
as was to be shown.

\subsection{Proof of Lemma~\ref{lem:sg-2}}

\begin{proof}
Define $\maxopi(\param)$ as
\begin{equation}
\maxopi(\param) \eqdef \max_{|T| \geq \lfloor \opfrac \goodfrac n \rfloor} \frac{1}{\sqrt{|T|}} \|\left[ \nabla \func_i(\param) - \nabla \Efunc(\param) \right]_{i \in T}\|_{\op}.
\end{equation}
By Lemma~\ref{lem:subgaussian} below, 
at any fixed $\param$ it is the case that 
\begin{equation}
\maxopi(\param) 
\leq 6\sg\sqrt{\log(e/\opfrac) + \frac{d+\log(1/\delta)}{\opfrac\goodfrac n}}.
\end{equation}
On the other hand, because $\nabla \func_i(\param) - \nabla \Efunc(\param)$ is $L$-Lipschitz 
in $\param$, one can check that $\maxopi(\param)$ is also $L$-Lipschitz in $\param$ (see Lemma~\ref{lem:sg-lipschitz} below). In 
particular, around any fixed $\param$, it holds that 
$\maxopi(\param) \leq \sg\p{1 + 6\sqrt{\log(e/\opfrac) + \frac{d + \log(1/\delta)}{\opfrac \goodfrac n}}}$ for all $\param'$ with $\|\param' - \param\|_2 \leq \frac{\sg}{L}$. But we can 
cover $\sH$ with $\p{1 + \frac{2\radius L}{\sg}}^d$ balls of radius $\frac{\sg}{L}$. 
Therefore, a union bound argument yields that, with probability $1-\delta$, for all 
$\param \in \sH$ we have
\begin{align}
\maxopi(\param)
 &\leq \sg\p{1 + 6\sqrt{\log(e/\opfrac) + \frac{d\p{1 + \log\p{1 + \frac{2\radius L}{\sg}}} + \log(1/\delta)}{\opfrac \goodfrac n}}} \\
 &= \oo\p{\sg\p{\sqrt{\log(2/\opfrac)} + \sqrt{\frac{d\max\p{1, \log\p{\frac{\radius L}{\sg}}} + \log(1/\delta)}{\opfrac \goodfrac n}}}},
\end{align}
as was to be shown.
\end{proof}

\begin{lemma}
\label{lem:sg-lipschitz}
If $\nabla \func_i(\param) - \nabla \Efunc(\param)$ is $L$-Lipschitz in 
$\param$, then $\maxopi(\param)$ is also $L$-Lipschitz in $\param$.
\end{lemma}
\begin{proof}
It suffices to show that $\frac{1}{\sqrt{|T|}}\left\|\left[ \nabla \func_i(\param) - \nabla \Efunc(\param) \right]_{i \in T} \right\|_{\op}$ 
is $L$-Lipschitz for each $T$, since the max of Lipschitz functions is Lipschitz. 
To show this, let $\Delta_i(\param)$ denote $\nabla \func_i(\param) - \nabla \Efunc(\param)$, and note that 
\begin{align}
\frac{1}{\sqrt{|T|}}\left|\left\|\left[ \Delta_i(\param) \right]_{i \in T} \right\|_{\op} - \left\|\left[ \Delta_i(\param')\right]_{i \in T} \right\|_{\op}\right|
 &\stackrel{(i)}{\leq} \frac{1}{\sqrt{|T|}}\left\|\left[ \Delta_i(\param) - \Delta_i(\param') \right]_{i \in T} \right\|_{\op} \\
 &\stackrel{(ii)}{\leq} \frac{1}{\sqrt{|T|}}\left\|\left[ \Delta_i(\param) - \Delta_i(\param') \right]_{i \in T} \right\|_{F} \\
 &= \sqrt{\frac{1}{|T|} \sum_{i \in T} \|\Delta_i(\param) - \Delta_i(\param')\|_2^2} \\
 &\stackrel{(iii)}{\leq} \sqrt{\frac{1}{|T|} \sum_{i \in T} L^2 \|\param - \param'\|_2^2} \\
 &= L \|\param - \param'\|_2,
\end{align}
as was to be shown.
Here (i) is the triangle inequality for the operator norm, 
(ii) is because Frobenius norm is larger than operator norm, 
and (iii) invokes the assumed Lipschitz property for $\Delta_i(\param)$.
\end{proof}

\subsection{Proof of Lemma~\ref{lem:gaussian-cover}}

Recall that $p^* = \sN(\mu, \Sigma_0 - \lambda \mu\mu^{\top})$ and $\hat{p} = \sN(0, \Sigma_0)$. We want to show that 
$\goodfrac p^* \leq \hat{p}$, which is equivalent to (by expanding the density formula for a Gaussian distribution)
\begin{equation}
\label{eq:lambda-mu-x-cond}
(x-\mu)^{\top}(\Sigma_0 - \lambda \mu\mu^{\top})^{-1}(x-\mu) - x^{\top}\Sigma_0^{-1}x + \log\det(I - \lambda \mu\mu^{\top}\Sigma_0^{-1}) \stackrel{?}{\geq} 2\log(\goodfrac) \text{ for all } x.
\end{equation}
Let us simplify the above quadratic in $x$. Letting $s = x^{\top}\Sigma_0^{-1}\mu$ and $t = \mu^{\top}\Sigma_0^{-1}\mu$, we have
\begin{align}
\lefteqn{\hskip -0.25in (x-\mu)^{\top}(\Sigma_0 - \lambda\mu\mu^{\top})^{-1}(x-\mu) - x^{\top}\Sigma_0^{-1}x} \\
 &\stackrel{(i)}{=} (x-\mu)^{\top}\p{\Sigma_0^{-1} + \lambda\frac{\Sigma_0^{-1}\mu\mu^{\top}\Sigma_0^{-1}}{1-\lambda\mu^{\top}\Sigma_0^{-1}\mu}}(x-\mu) - x^{\top}\Sigma_0^{-1}x \\
 &\stackrel{(ii)}{=} \lambda \frac{(x^{\top}\Sigma_0^{-1}\mu)^2}{1-\lambda \mu^{\top}\Sigma_0^{-1}\mu} - 2\p{x^{\top}\Sigma_0^{-1}\mu + \lambda \frac{(x^{\top}\Sigma_0^{-1}\mu)(\mu^{\top}\Sigma_0^{-1}\mu)}{1-\lambda \mu^{\top}\Sigma_0^{-1}\mu}} + \p{\mu^{\top}\Sigma_0^{-1}\mu + \lambda \frac{(\mu^{\top}\Sigma_0^{-1}\mu)^2}{1 - \lambda \mu^{\top}\Sigma_0^{-1}\mu}} \\
 &= \lambda \frac{s^2}{1-\lambda t} - 2\p{s + \frac{\lambda st}{1-\lambda t}} + \p{t + \lambda \frac{t^2}{1-\lambda t}} \\
 &= \frac{1}{1-\lambda t}\p{\lambda s^2 - 2s + t},
\end{align}
where (i) is the Woodbury matrix inversion lemma, and (ii) is expanding the quadratic.
The minimizing value of $s$ is $\frac{1}{\lambda}$, so that we obtain a lower bound (over all $x$) of 
$$(x-\mu)^{\top}\p{\Sigma_0 - \lambda \mu\mu^{\top}}^{-1}(x-\mu) - x^{\top}\Sigma_0^{-1}x \geq \frac{1}{1-\lambda t}\p{-\frac{1}{\lambda} + t} = -\frac{1}{\lambda}.$$
Plugging back into \eqref{eq:lambda-mu-x-cond}, we obtain the condition
\begin{equation}
\label{eq:lambda-mu-cond}
-\frac{1}{\lambda} + \log\det(I - \lambda \mu\mu^{\top}\Sigma_0^{-1}) \stackrel{?}{\geq} 2\log(\goodfrac).
\end{equation}
Simplifying further and again letting $t = \mu^{\top}\Sigma_0^{-1}\mu$, we have
\begin{align}
-\frac{1}{\lambda} + \log\det(I - \lambda \mu\mu^{\top}\Sigma_0^{-1})
 &= -\frac{1}{\lambda} + \log(1 - \lambda \mu^{\top}\Sigma_0^{-1}\mu) \\
 &= -\frac{1}{\lambda} + \log(1 - \lambda t) \\
 &\geq -\frac{1}{\lambda} - \frac{\lambda t}{1 - \lambda t},
\end{align}
so it suffices to have $\frac{1}{\lambda} + \frac{\lambda t}{1 - \lambda t} \leq 2\log(\frac{1}{\goodfrac})$.
Now, we will take $\lambda = \frac{1}{\log(\frac{1}{\goodfrac})}$, in which case we need 
$\frac{t}{\log(\frac{1}{\goodfrac}) - t} \leq \log(\frac{1}{\goodfrac})$, or equivalently 
$t \leq \frac{\log^2(\frac{1}{\goodfrac})}{1 + \log(\frac{1}{\goodfrac})}$, as claimed.

\subsection{Bounding the square of a sub-Gaussian}

\begin{lemma}
\label{lem:sg-se}
Suppose that $Z$ is a random variable
satisfying $\bE[\exp\p{\lambda X}] \leq \exp\p{\frac{\sg^2}{2}\lambda^2}$. 
Then, $Y = Z^2 - \bE[Z^2]$ satisfies
\begin{equation}
\bE[\exp\p{\lambda Y}] \leq \exp\p{16\sg^4\lambda^2}
\end{equation}
for $|\lambda| \leq \frac{1}{4\sg^2}$.
\end{lemma}
\begin{proof}
We begin by bounding the moments of 
$Z$. We have, for any $p \geq 0$ and $u$, 
\begin{align}
\bE\left[|Z|^p\right] 
 &= \bE\left[\exp\p{p\log|Z|}\right] \\
 &\stackrel{(i)}{\leq} \bE\left[\exp\p{p\p{\log a + \frac{1}{a}|Z| - 1}}\right] \\
 &\stackrel{(ii)}{\leq} \bE\left[\exp\p{p\p{\log a + \frac{1}{a}Z - 1}} + \exp\p{p\p{\log a - \frac{1}{a}Z - 1}}\right] \\
 &= \bE\left[\exp\p{p\p{\log a -1}}\p{\exp\p{\frac{p}{a}Z} + \exp\p{-\frac{p}{a}Z}}\right] \\
 &\stackrel{(iii)}{\leq} 2\exp\p{p\p{\log a - 1}} \exp\p{\frac{p^2\sg^2}{2a^2}}.
\end{align}
Here (i) uses the inequality $\log(z) - \log(a) \leq \frac{1}{a}\p{z-a}$ (by concavity of $\log$), 
while (iii) simply invokes the sub-Gaussian assumption at 
$\lambda = \frac{p}{a}$. The inequality (ii) is used to move from $|Z|$ to $Z$.

We will take the particular value $a = \sqrt{p}\sg$, which yields
\begin{align}
\bE\left[|Z|^p\right] 
 &\leq 2\exp\p{p\p{\frac{1}{2}\log p + \log(\sg) - \frac{1}{2}}} \\
 &= 2\sg^p \exp\p{\frac{1}{2}p\p{\log(p)-1}}.
\end{align}
Now, letting $u = \bE[Z^2]$, we have
\begin{align}
\bE[|Z^2-u|^p] 
 &\leq 2^{p-1}\p{\bE[|Z|^{2p}] + u^p} \\
 &\leq 2^{p-1}\p{2\sg^{2p}\exp\p{p(\log(2p)-1)} + \sg^{2p}} \\
\label{eq:sg-se-1}
 &\leq 2^{p+1}\sg^{2p}\exp\p{p(\log(2p)-1)}.
\end{align}
Hence,
\begin{align}
\bE[\exp\p{\lambda Y}] 
 &= \bE[\exp\p{\lambda (Z^2 - u)}] \\
 &\leq 1 + \sum_{p=2}^{\infty} \bE[|Z^2-u|^p] \frac{|\lambda|^p}{p!} \\
 &\stackrel{\eqref{eq:sg-se-1}}{\leq} 1 + 2 \sum_{p=2}^{\infty} \frac{(2|\lambda| \sg^2)^p}{p!} \exp\p{p(\log(2p)-1)} \\
 &\leq 1 + 2 \sum_{p=2}^{\infty} (2|\lambda| \sg^2)^p \exp\p{p(\log(2p)-1) - p\log(p)} \\
 &= 1 + 2 \sum_{p=2}^{\infty} (2|\lambda| \sg^2)^p \exp\p{p(\log(2)-1)} \\
 &\leq 1 + 2 \sum_{p=2}^{\infty} (2|\lambda| \sg^2)^p \\
 &= 1 + \frac{8\lambda^2 \sg^4}{1 - 2|\lambda| \sg^2}.
\end{align}
Clearly, when $|\lambda| \leq \frac{1}{4\sg^2}$, the above 
is bounded by $1 + 16\lambda^2 \sg^4 \leq \exp\p{16\sigma^4\lambda^2}$,
as claimed.
\end{proof}

\bibliographystyle{plainnat}
\bibliography{refdb/all}

\end{document}